\RequirePackage{snapshot}
\documentclass[10pt,twocolumn,letterpaper]{article}

\usepackage[pagenumbers]{cvpr} %

\usepackage[dvipsnames]{xcolor}

\definecolor{cvprblue}{rgb}{0.21,0.49,0.74}
\usepackage[pagebackref,breaklinks,colorlinks,citecolor=cvprblue]{hyperref}

\usepackage{amsmath}
\usepackage{amssymb}
\usepackage{bm,mathtools}

\newcommand{\cskip}{c_\mathrm{skip}}
\newcommand{\cout}{c_\mathrm{out}}
\newcommand{\cin}{c_\mathrm{in}}
\newcommand{\cnoise}{c_\mathrm{noise}}

\def\eqref#1{equation~\ref{#1}}

\def\1{\bm{1}}

\def\rvc{{\mathbf{c}}}

\def\rvn{{\mathbf{n}}}

\def\rvx{{\mathbf{x}}}

\def\vtheta{{\bm{\theta}}}

\def\vs{{\bm{s}}}

\DeclareMathAlphabet{\mathsfit}{\encodingdefault}{\sfdefault}{m}{sl}
\SetMathAlphabet{\mathsfit}{bold}{\encodingdefault}{\sfdefault}{bx}{n}

\def\gN{{\mathcal{N}}}

\newcommand{\pdata}{p_{\rm{data}}}

\newcommand{\E}{\mathbb{E}}

\newcommand{\R}{\mathbb{R}}

\DeclareMathOperator*{\argmin}{arg\,min}

\usepackage{url}
\usepackage{graphicx}
\usepackage{adjustbox,booktabs,multirow}
\usepackage{caption}
\usepackage{subcaption}
\usepackage{pifont}
\usepackage{xspace}
\usepackage{tabularx}

\definecolor{codegreen}{rgb}{0,0.6,0}
\definecolor{codegray}{rgb}{0.5,0.5,0.5}
\definecolor{codepurple}{rgb}{0.58,0,0.82}
\definecolor{backcolour}{rgb}{0.95,0.95,0.92}
\definecolor{codered}{rgb}{0.89,0.4,.45}

\usepackage{listings}
\lstdefinestyle{mystyle}{
    backgroundcolor=\color{backcolour},   
    commentstyle=\color{codegreen},
    keywordstyle=\color{codered},
    numberstyle=\tiny\color{codegray},
    stringstyle=\color{codepurple},
    basicstyle=\ttfamily\footnotesize,
    breakatwhitespace=false,         
    breaklines=true,                 
    captionpos=b,                    
    keepspaces=true,                 
    numbers=left,                    
    numbersep=5pt,                  
    showspaces=false,                
    showstringspaces=false,
    showtabs=false,                  
    tabsize=2
}

\usepackage{graphicx,multirow}

\def\paperID{12936} %
\def\confName{CVPR}
\def\confYear{2024}

\title{Stable Video Diffusion: Scaling Latent Video Diffusion Models to Large Datasets\vspace{-22pt}}

\newcommand{\authortable}{
    \begingroup
    \renewcommand{\arraystretch}{1.2}
    \setlength{\tabcolsep}{4pt}
    \begin{tabular}{cccc} 
        Andreas Blattmann\textsuperscript{*} \quad\quad Tim Dockhorn\textsuperscript{*} \quad\quad Sumith Kulal\textsuperscript{*} \quad\quad Daniel Mendelevitch \quad\quad
        \\ Maciej Kilian \quad\quad Dominik Lorenz \quad\quad Yam Levi \quad\quad Zion English \quad\quad
        Vikram Voleti \\ Adam Letts \quad\quad Varun Jampani \quad\quad Robin Rombach
        \vspace{2pt} \\
    \end{tabular}
    \endgroup\\
    \vspace{6pt}
    Stability AI
}

\author{\authortable}

\newcommand{\cmark}{\ding{51}}%
\newcommand{\xmark}{\ding{55}}%
\newcommand{\dataset}{\emph{LVD}\xspace}%
\newcommand{\datasetfiltered}{\emph{LVD-F}\xspace}%
\newcommand{\datasetsmall}{\emph{LVD-10M}\xspace}%
\newcommand{\datasetsmallfiltered}{\emph{LVD-10M-F}\xspace}%

\newcommand{\datastats}{
\begin{table}
\begin{center}
\caption{\label{tab:subset_stats} Comparison of our dataset before and after fitering with publicly available research datasets.}
\vspace{-1em}
\begin{adjustbox}{max width=.48\textwidth}
\begin{tabular}{lc c c c | c c}
   \toprule
 & \dataset & \datasetfiltered & \datasetsmall& \datasetsmallfiltered &  \emph{WebVid}& \emph{InternVid}\\ 
   \midrule
 \#Clips  &  577M & 152M & 9.8M & 2.3M & 10.7M & 234M  \\[1pt]  
 Clip Duration (s)  & 11.58  & 10.53 & 12.11 & 10.99 & 18.0 & 11.7 \\[1pt]
 Total Duration (y) & 212.09 & 50.64& 3.76 & 0.78 & 5.94 & 86.80\\[1pt]
 Mean \#Frames  & 325 & 301& 335 & 320 & - & - \\[1pt]
 Mean Clips/Video & 11.09 & 4.76&1.2 & 1.1 & 1.0 & 32.96 \\
 Motion Annotations? &  \cmark & \cmark& \cmark& \cmark& \xmark & \xmark \\
 \bottomrule
\end{tabular}
\end{adjustbox}
\end{center}
\vspace{-2em}
\end{table}
}

\newcommand{\ucfzeroshotandsotavid}{
\begin{figure}[ht]
    \centering
    \hspace{-6pt}
        \begin{minipage}{0.23\textwidth}
            \captionof{table}{\small UCF-101 zero-shot text-to-video generation. Comparing our base model to baselines (numbers from literature).}
            \label{tab:ucf}
            \centering
            \resizebox*{!}{.7\columnwidth}{
            \begin{tabular}{l c}
                \toprule
                \textbf{Method} & FVD ($\downarrow$)  \\
                \midrule
                CogVideo (ZH)~\citep{hong2022cogvideo} & 751.34 \\
                CogVideo (EN)~\citep{hong2022cogvideo} & 701.59 \\
                Make-A-Video~\citep{singer2022make} & 367.23 \\
                Video LDM~\citep{blattmann2023align} & 550.61 \\
                MagicVideo~\citep{zhou2022magicvideo} & 655.00 \\
                PYOCO~\citep{ge2023preserve} & 355.20 \\
                \midrule
                SVD (\emph{ours}) & \textbf{242.02} \\
                \bottomrule
            \end{tabular}
            }
        \end{minipage}
        \begin{minipage}{0.23\textwidth}
            \centering
            \includegraphics[width=\linewidth]{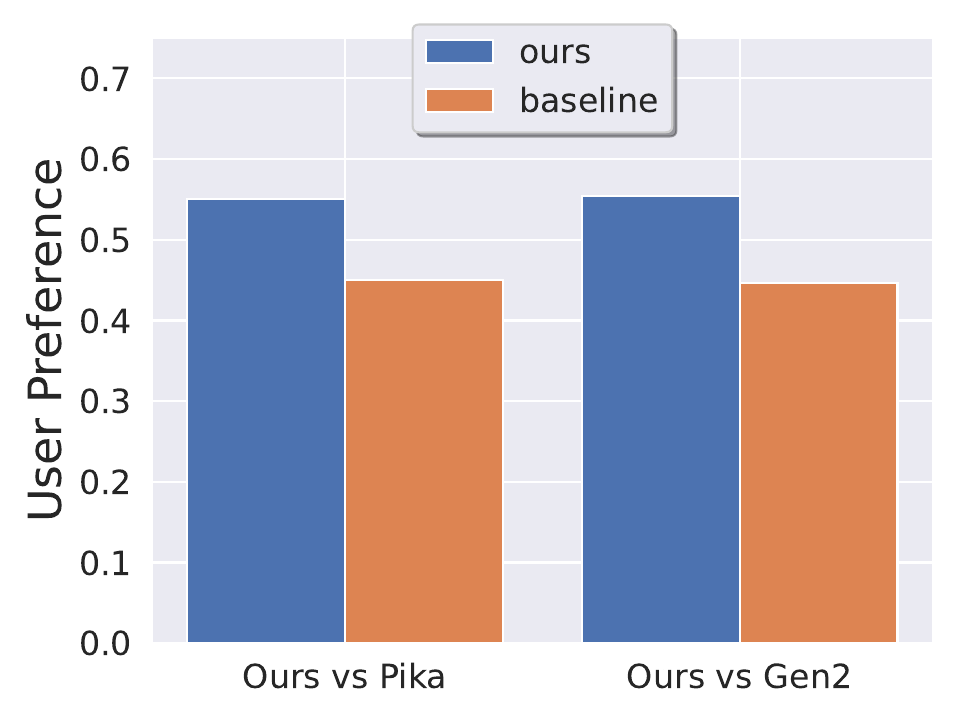}
            \vspace{-18pt}
            \caption{\label{fig:sotaimg2vid}\small Our 25 frame Image-to-Video model is preferred by human voters over GEN-2~\cite{gen2} and PikaLabs~\cite{pika}.}
        \end{minipage}
\end{figure}
}

\providecommand{\impath}[1]{}
\providecommand{\imwidth}{}

\newcommand{\clippingflow}{
\renewcommand{\imwidth}{0.5\textwidth}
\begin{figure}[h]
    \centering
    \hspace{-10pt}
    \includegraphics[width=0.49\textwidth]{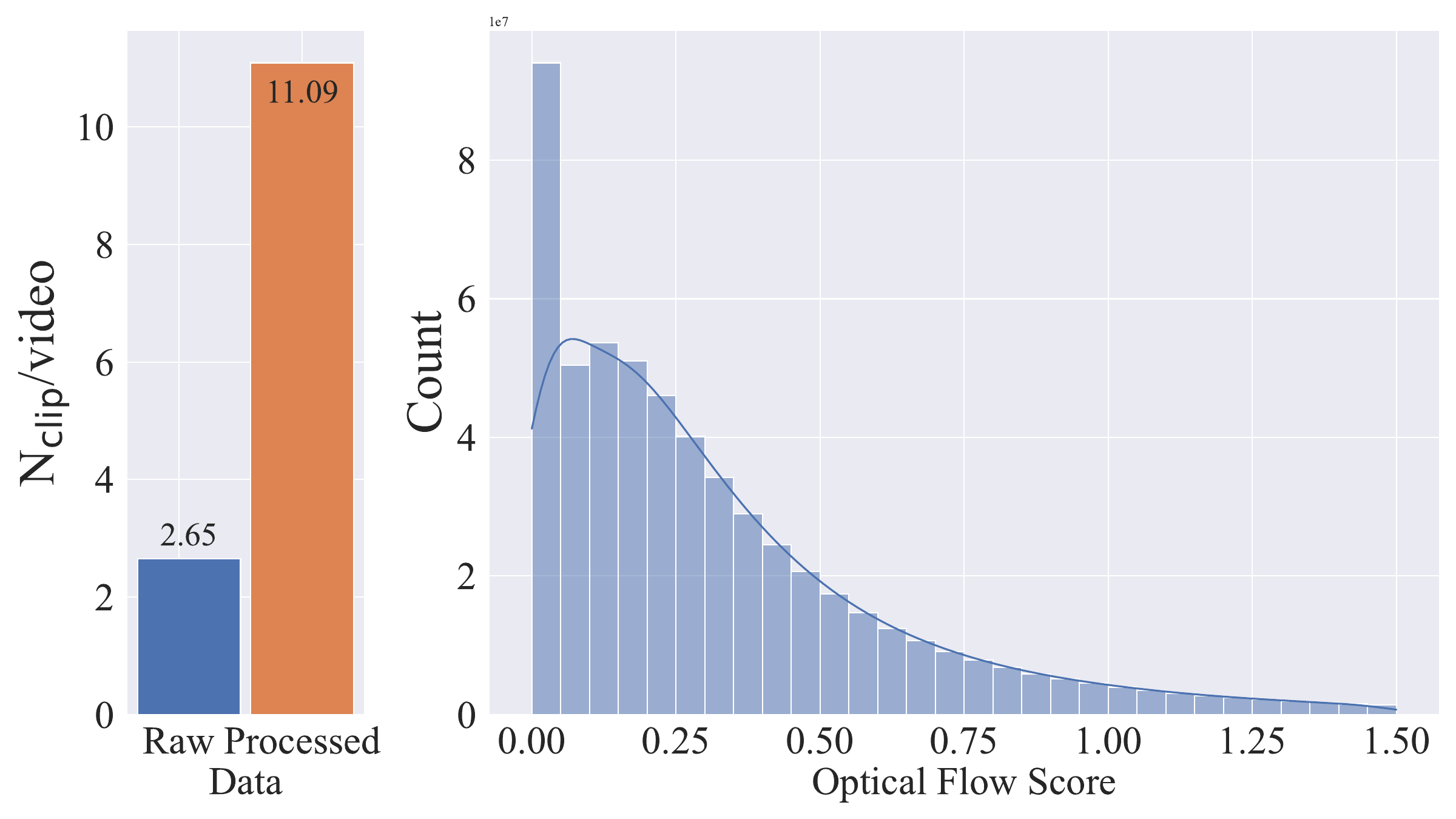}
    \caption{\label{fig:cuts_and_motion}Our initial dataset contains many static scenes and  cuts which hurts training of generative video models. \emph{Left}: Average number of clips per video before and after our processing, revealing that our pipeline detects lots of additional cuts. \emph{Right}: We show the distribution of average optical flow score for one of these subsets before our processing, which contains many static clips. }
\end{figure}
}

\newcommand{\bigfatplot}{
\begin{figure*}[htbp]
\renewcommand{\imwidth}{0.19\textwidth}
\begin{center}
    \begin{subfigure}[t]{\imwidth}
        \includegraphics[width=\textwidth]{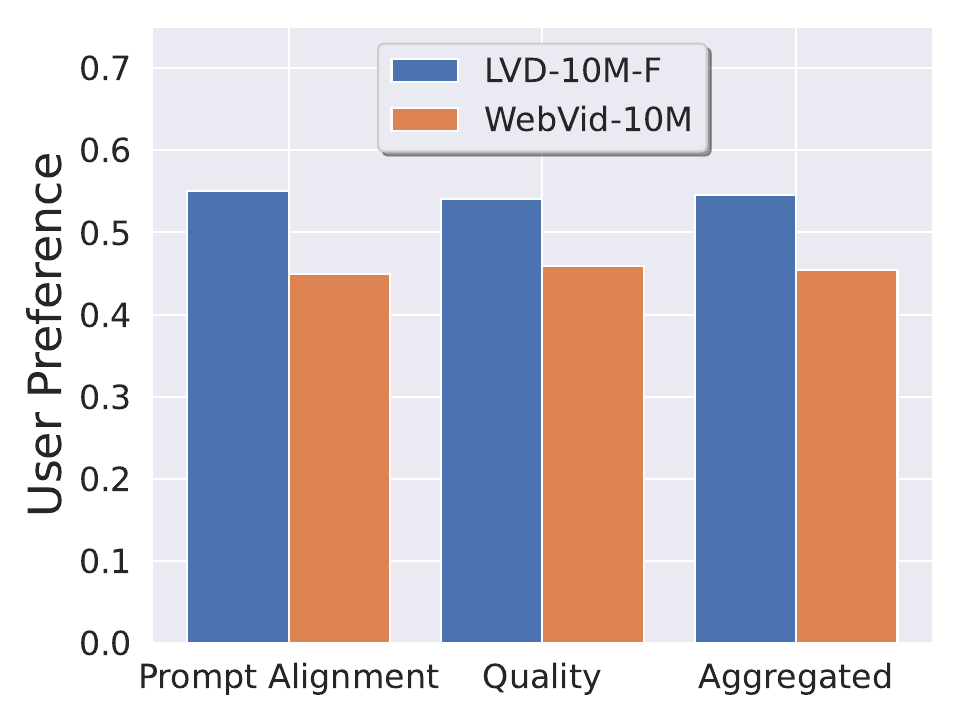}
        
        \caption{\label{fig:webivd_comp} User preference for \datasetsmallfiltered and WebVid~\cite{bain2022frozen}. }
    \end{subfigure}
    \hfill
    \begin{subfigure}[t]{\imwidth}
        \includegraphics[width=\textwidth]{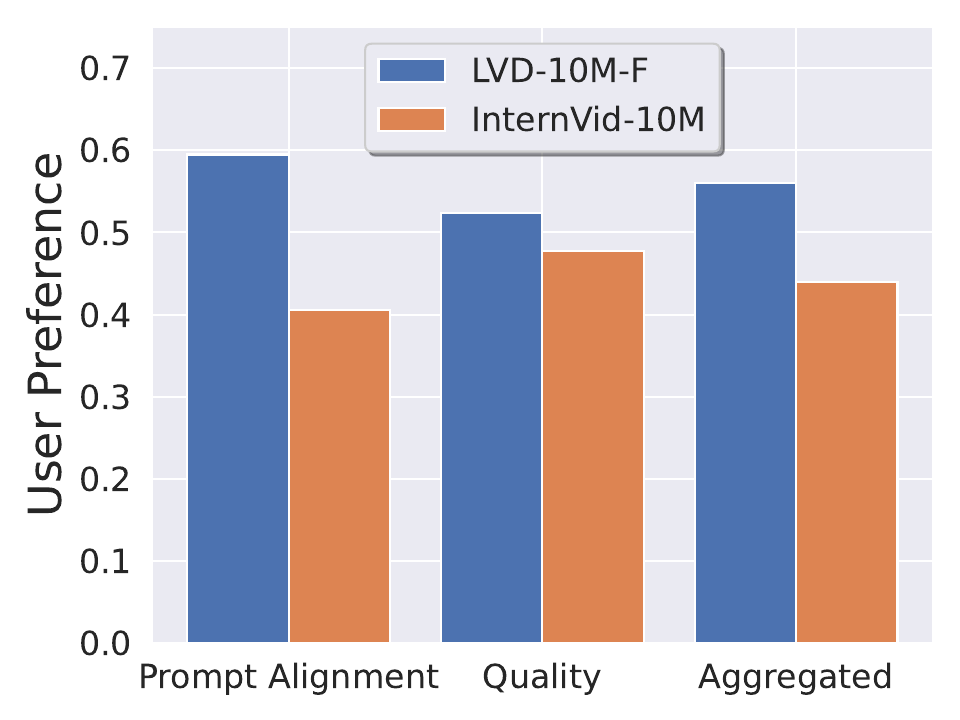}
        \caption{\label{fig:internvid_comp} User preference for \datasetsmallfiltered and InternVid~\cite{wang2023internvid}.}
    \end{subfigure}%
    \hfill
    \begin{subfigure}[t]{\imwidth}
        \includegraphics[width=\textwidth]{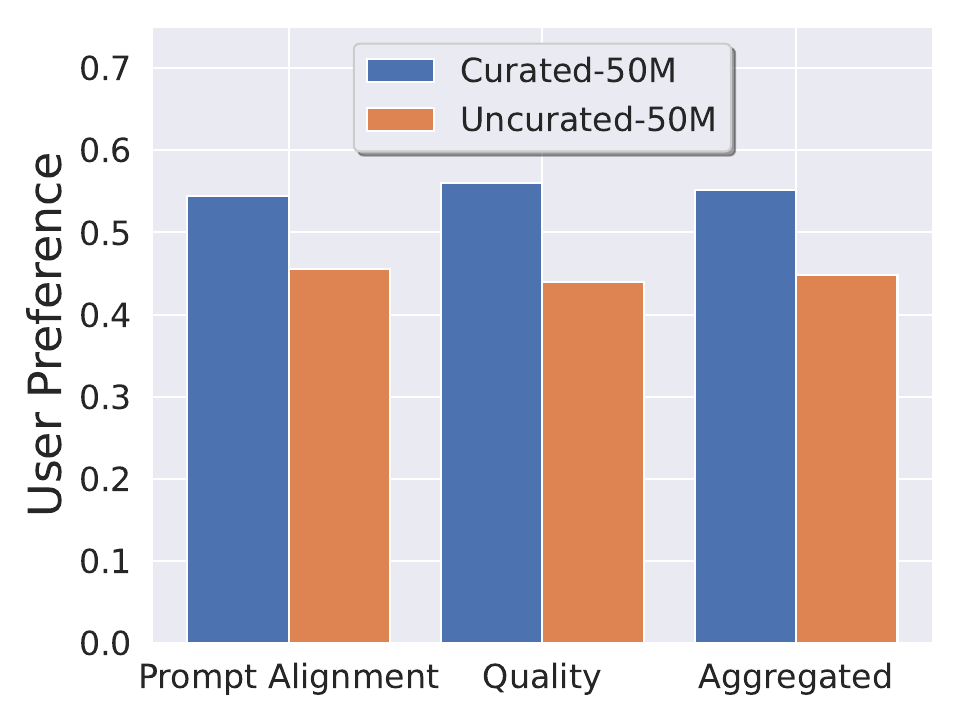}
        \caption{\label{fig:more_data} User preference at 50M samples scales. }
    \end{subfigure}
    \hfill
    \begin{subfigure}[t]{\imwidth}
        \includegraphics[width=\textwidth]{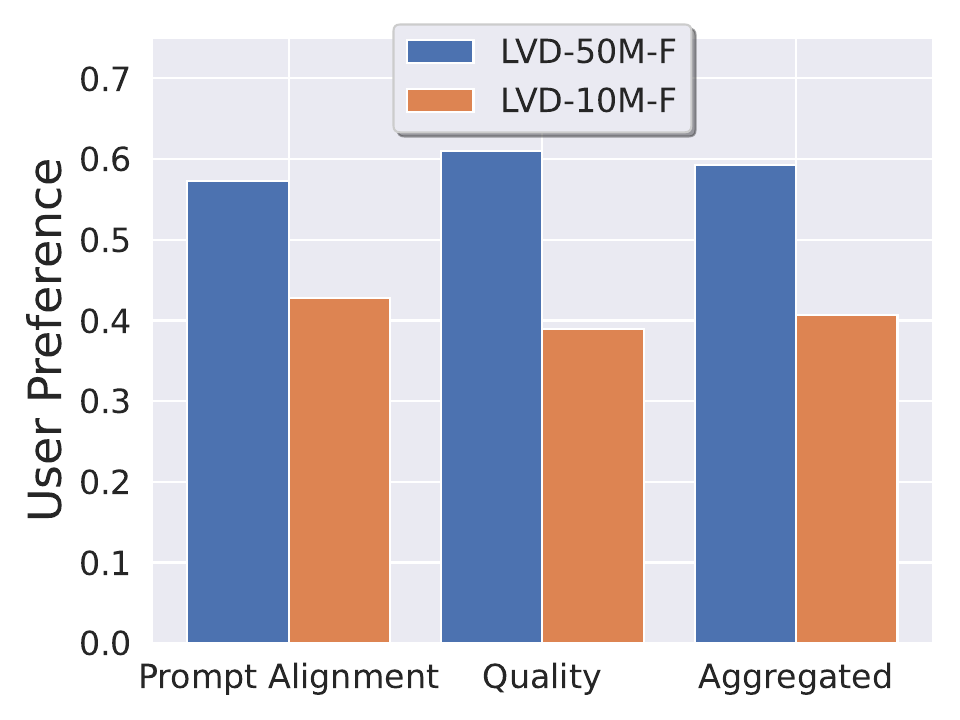}
        \caption{\label{fig:scaled_mid} User preference on scaling datasets.}
    \end{subfigure}%
    \hfill
    \begin{subfigure}[t]{\imwidth}
        \includegraphics[width=\textwidth]{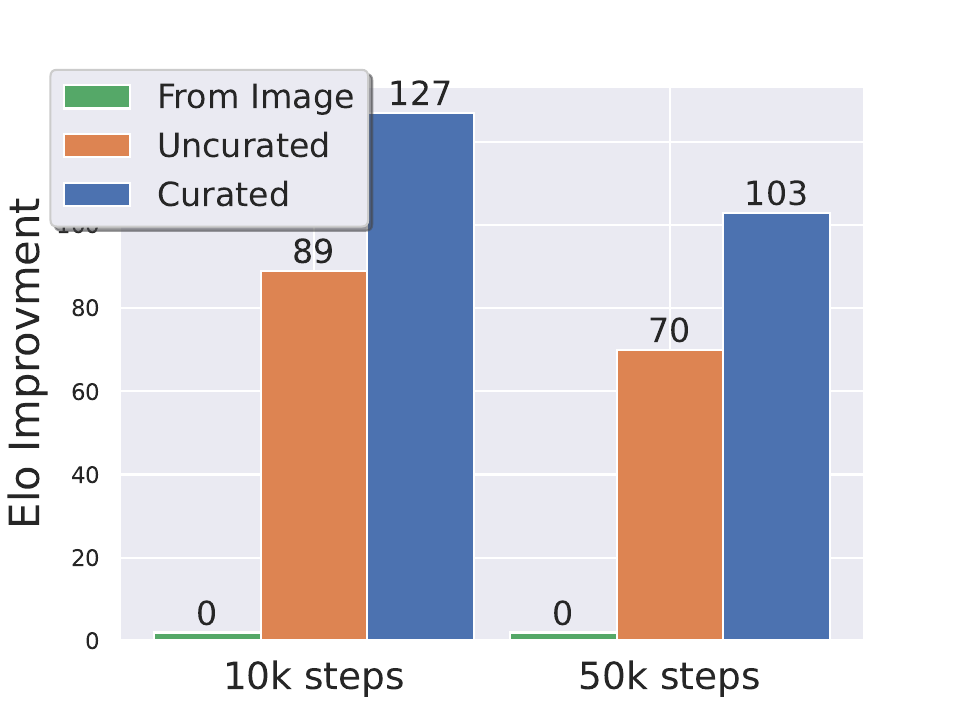}
        \caption{\label{fig:finetune_elos} Relative ELO progression over time during Stage III.}
    \end{subfigure}%
    \vspace{-2em}
    \end{center}
    \caption{\label{fig:scaling_mid}  \emph{Summarized findings of \Cref{subsec:data_curation,subsec:stage3}}: Pretraining on curated datasets consistently boosts performance of generative video models during \emph{video pretraining} at small (\Cref{fig:webivd_comp,fig:internvid_comp}) and larger scales (\Cref{fig:more_data,fig:scaled_mid}). Remarkably, this performance improvement persists even after 50k steps of \emph{video finetuning} on high quality data (\Cref{fig:finetune_elos}).}
\end{figure*}
    \vspace{-0.5em}
}

\newcommand{\humanevalserver}{
\begin{figure*}[htbp]
\renewcommand{\imwidth}{0.48\textwidth}
\begin{center}
    \begin{subfigure}[t]{\imwidth}
        \includegraphics[width=\textwidth]{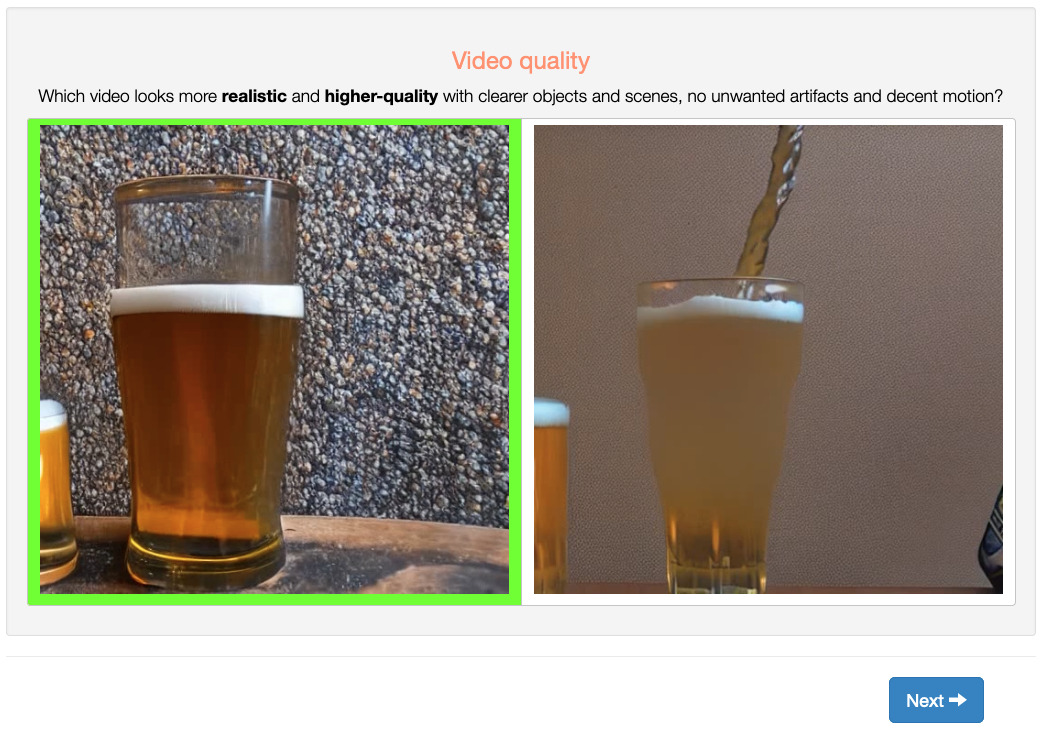}
        \caption{Sample instructions for evaluating visual quality of videos.}
    \end{subfigure}
    \hfill
    \begin{subfigure}[t]{\imwidth}
        \includegraphics[width=\textwidth]{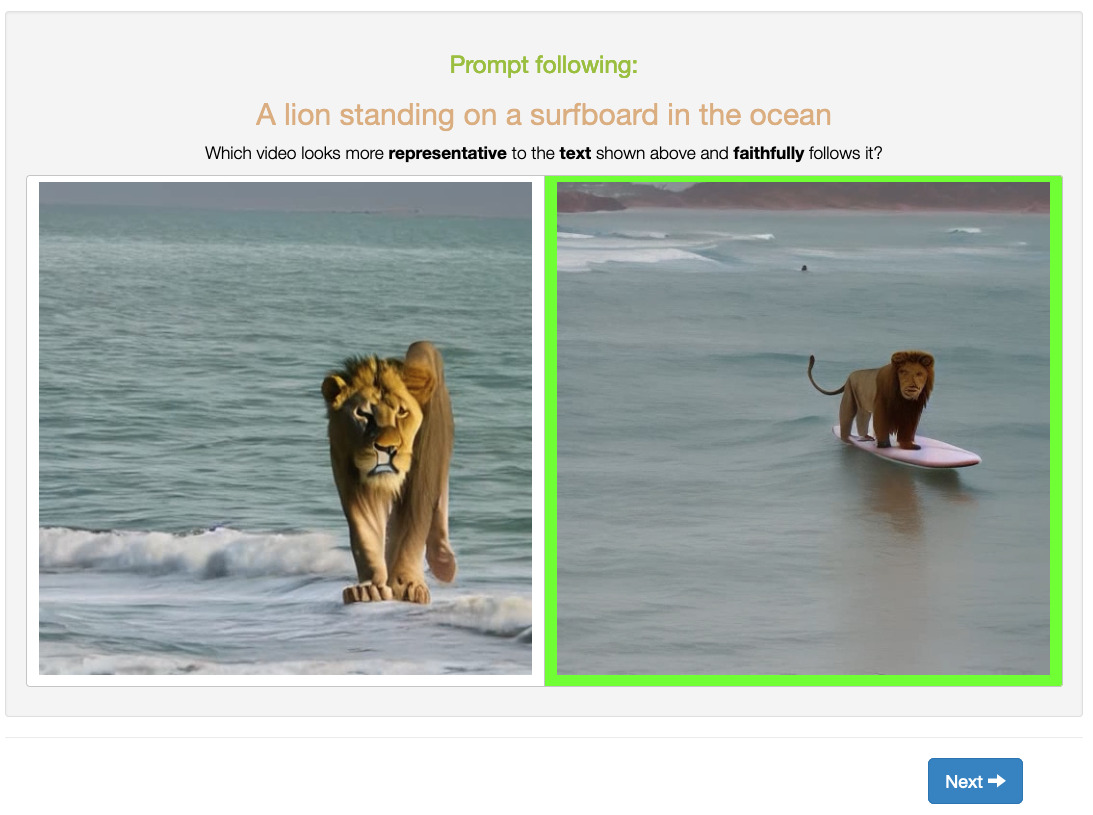}
        \caption{Sample instructions for evaluating the prompt following of videos.}
    \end{subfigure}%
\end{center}
    \caption{\label{fig:humanevalserver} Our human evaluation framework, as seen by the annotators. The prompt \& task order and model choices are fully randomized.}

\end{figure*}
}

\newcommand{\imageonlyandfiltering}{
\begin{figure}[htbp]
\renewcommand{\imwidth}{0.23\textwidth}
\begin{center}
    \begin{subfigure}[t]{\imwidth}
        \begin{minipage}[b]{\textwidth}
        
        \includegraphics[width=\textwidth]{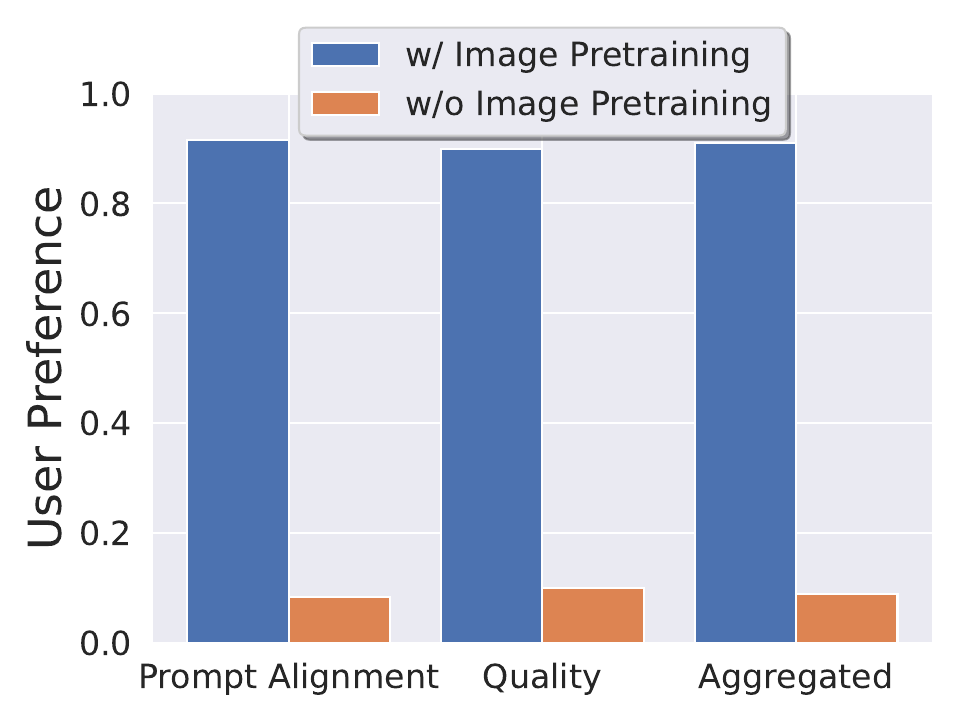}
        
        \end{minipage}
        \caption{\label{fig:imageonly_comp} Initializing spatial layers from pretrained images models greatly improves performance.}
    \end{subfigure}
    \hfill
    \begin{subfigure}[t]{\imwidth}
    \begin{minipage}[t]{\textwidth}
        \includegraphics[width=\textwidth]{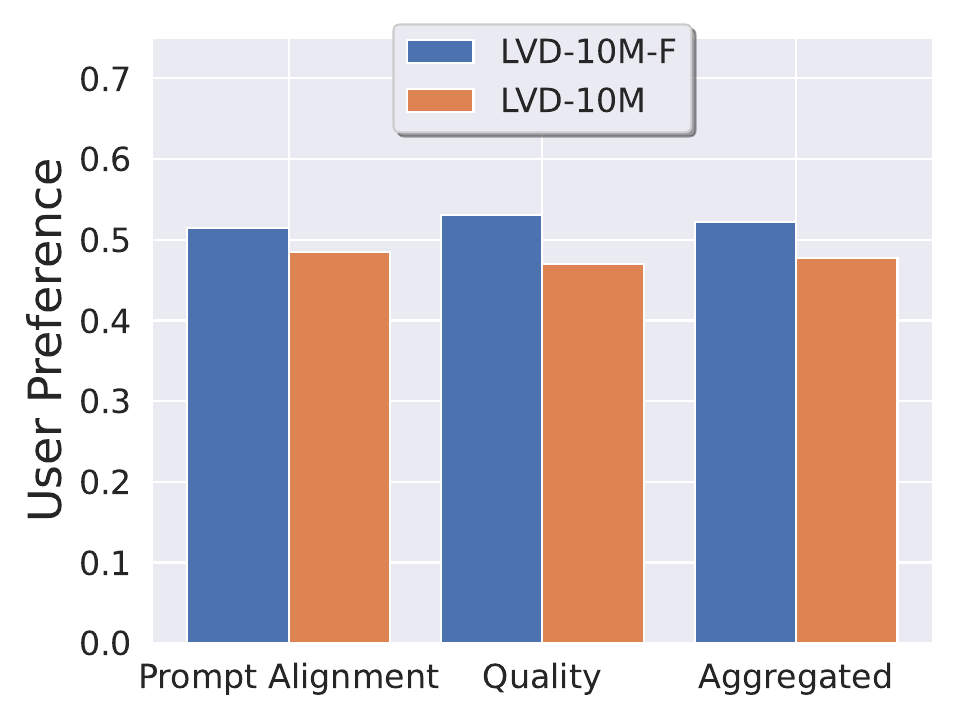}
    \end{minipage}
        \caption{\label{fig:filtering_effects} Video data curation boosts performance after video pretraining. }
    \end{subfigure}%
    \end{center}
    \vspace{-1.5em}
    \caption{\label{fig:imageonlyandfiltering}  \emph{Effects of image-only pretraining and data curation on video-pretraining on \datasetsmall:} A video model with spatial layers initialized from a pretrained image model clearly outperforms a similar one with randomly initialized spatial weights as shown in \Cref{fig:imageonly_comp}. \Cref{fig:filtering_effects} emphasizes the importance of data curation for pretraining, since training on a curated subset of \datasetsmall with the filtering threshold proposed in \Cref{subsec:data_curation} improves upon training on the entire uncurated \datasetsmall.}
    \vspace{-.5em}
\end{figure}
}

\newcommand{\linearpredictioncode}{
\begin{figure}[htbp]
\centering
\resizebox{.81\textwidth}{!}{%
\lstinputlisting[language=Python]{code/linear_prediction_guidance.py}
}
\caption{\label{fig:linear_pred_guidance} PyTorch code for our novel \emph{linearly increasing guidance} technique.}
\end{figure}
}

\newcommand{\teaserfigure}{
\twocolumn[{
    \maketitle
    \begin{center}
        \vspace{-30pt}
        \includegraphics[width=\linewidth]{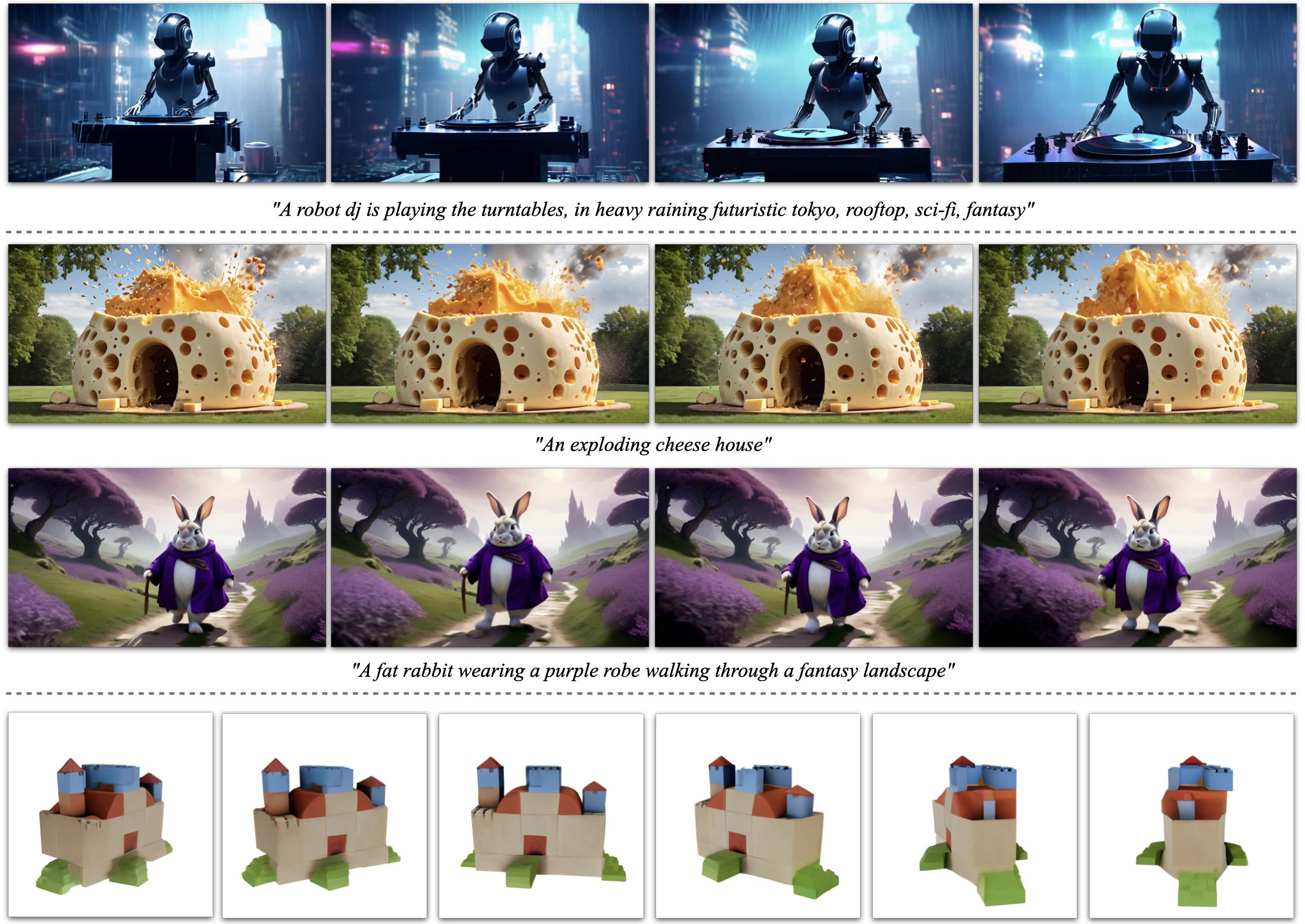}
    \vspace{-0.7cm}
    \captionof{figure}{
        \small \textbf{Stable Video Diffusion samples}. \emph{Top:} Text-to-Video generation. \emph{Middle:} (Text-to-)Image-to-Video generation. \emph{Bottom:} Multi-view synthesis via Image-to-Video finetuning. 
    }
    \vspace{-0.2cm}
    \label{fig:teaser}
    \end{center}
}]
}

\newcommand{\imgtwovid}{
\begin{figure*}
    \centering
    \includegraphics[width=\linewidth]{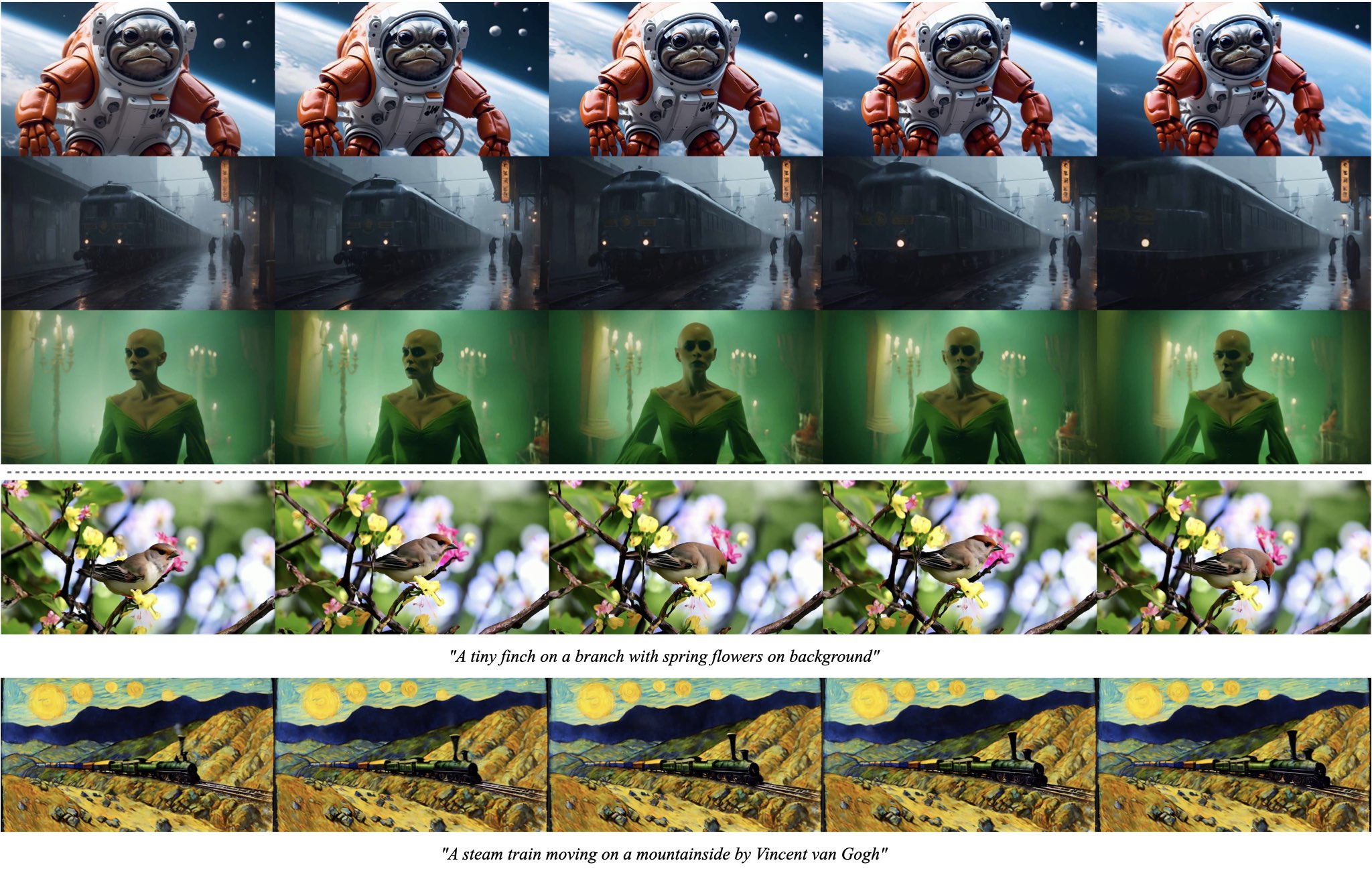}
    \vspace{-0.7cm}
    \caption{
        \small Samples at $576\times1024$. \emph{Top:} Image-to-video samples (conditioned on leftmost frame). \emph{Bottom:} Text-to-video samples.
    }
\label{fig:teaser2}
\end{figure*}
}

\newcommand{\MVIfig}{
\begin{figure}
    \centering
    \includegraphics[width=\linewidth]{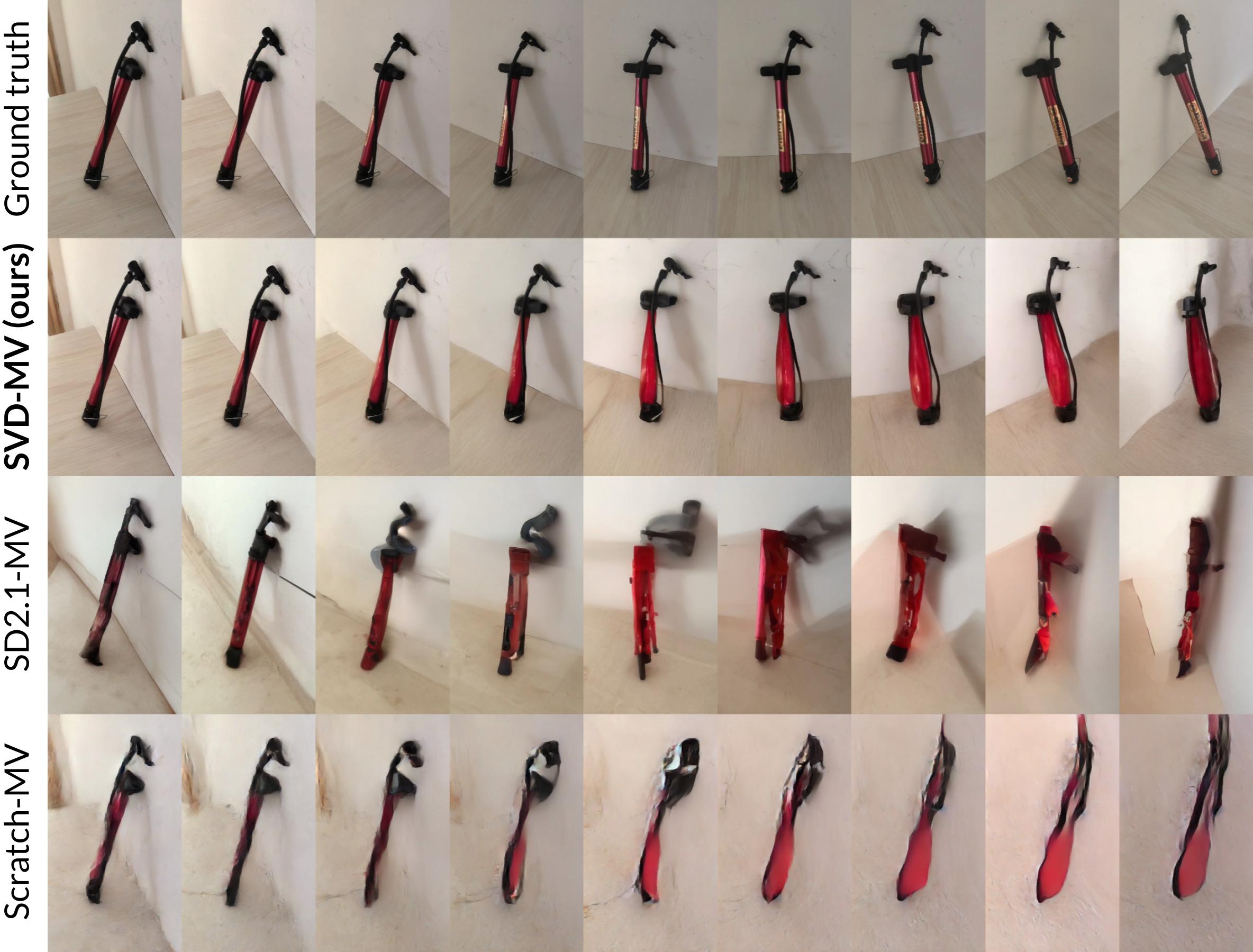}
    \vspace{-2em}
    \caption{
        \small Generated novel multi-view frames for MVImgNet dataset using our SVD-MV model, SD2.1-MV~\cite{rombach2021highresolution}, Scratch-MV.
    }
\label{fig:MVIfig}
\vspace{-1em}
\end{figure}
}

\newcommand{\vidtothreedfig}{
\begin{figure}
    \centering
    \includegraphics[width=\linewidth]{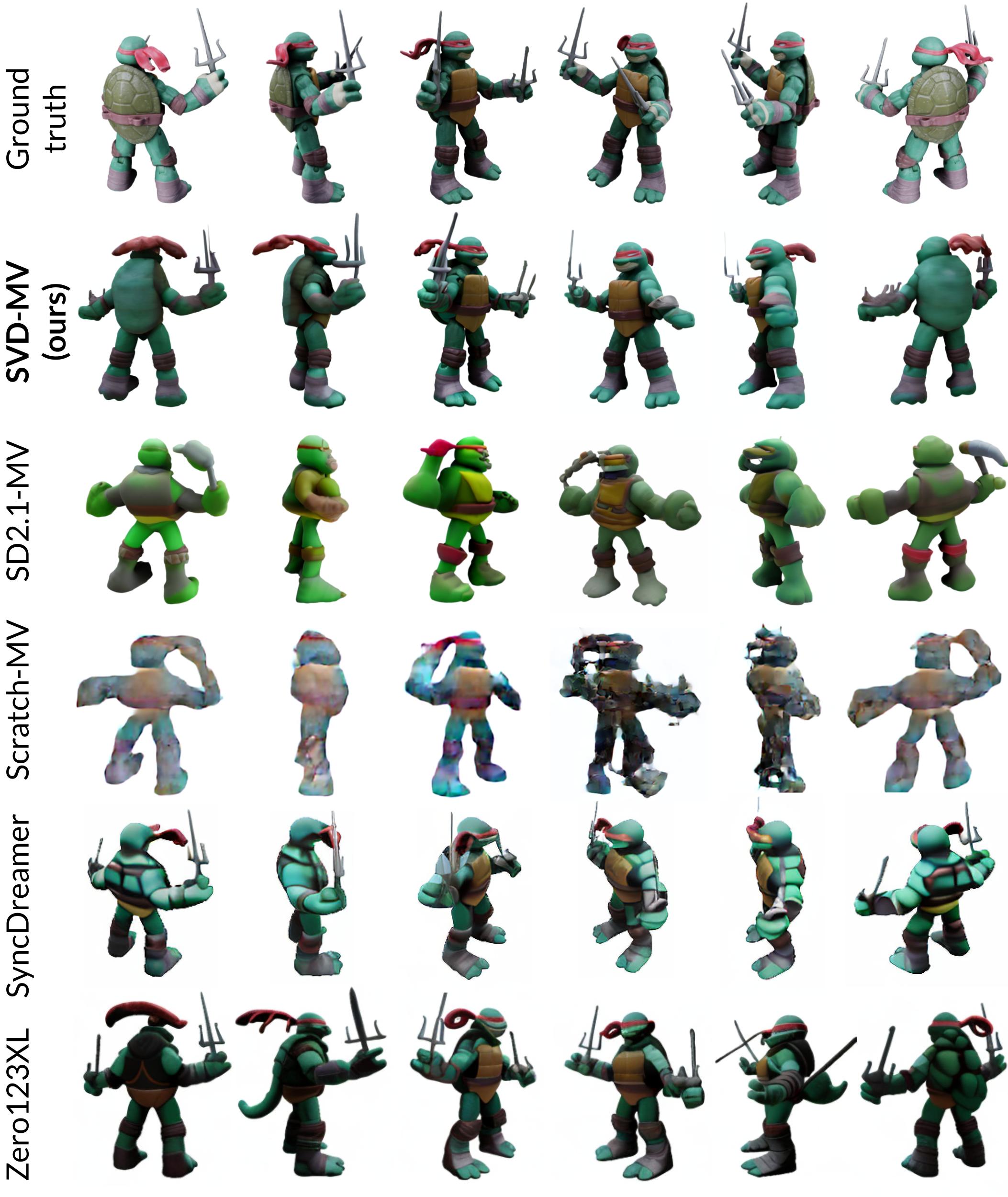}
    \vspace{-0.7cm}
    \caption{
        \small Generated multi-view frames of a GSO test object using our SVD-MV model (i.e. SVD finetuned for Multi-View generation), SD2.1-MV~\cite{rombach2021highresolution}, Scratch-MV, SyncDreamer~\cite{liu2023syncdreamer}, and Zero123XL~\cite{deitke2023objaversexl}.
    }
\vspace{-12pt}
\label{fig:vidto3dfig}
\end{figure}
}

\newcommand{\motionlora}{
\begin{figure}
    \centering
    \includegraphics[width=\linewidth]{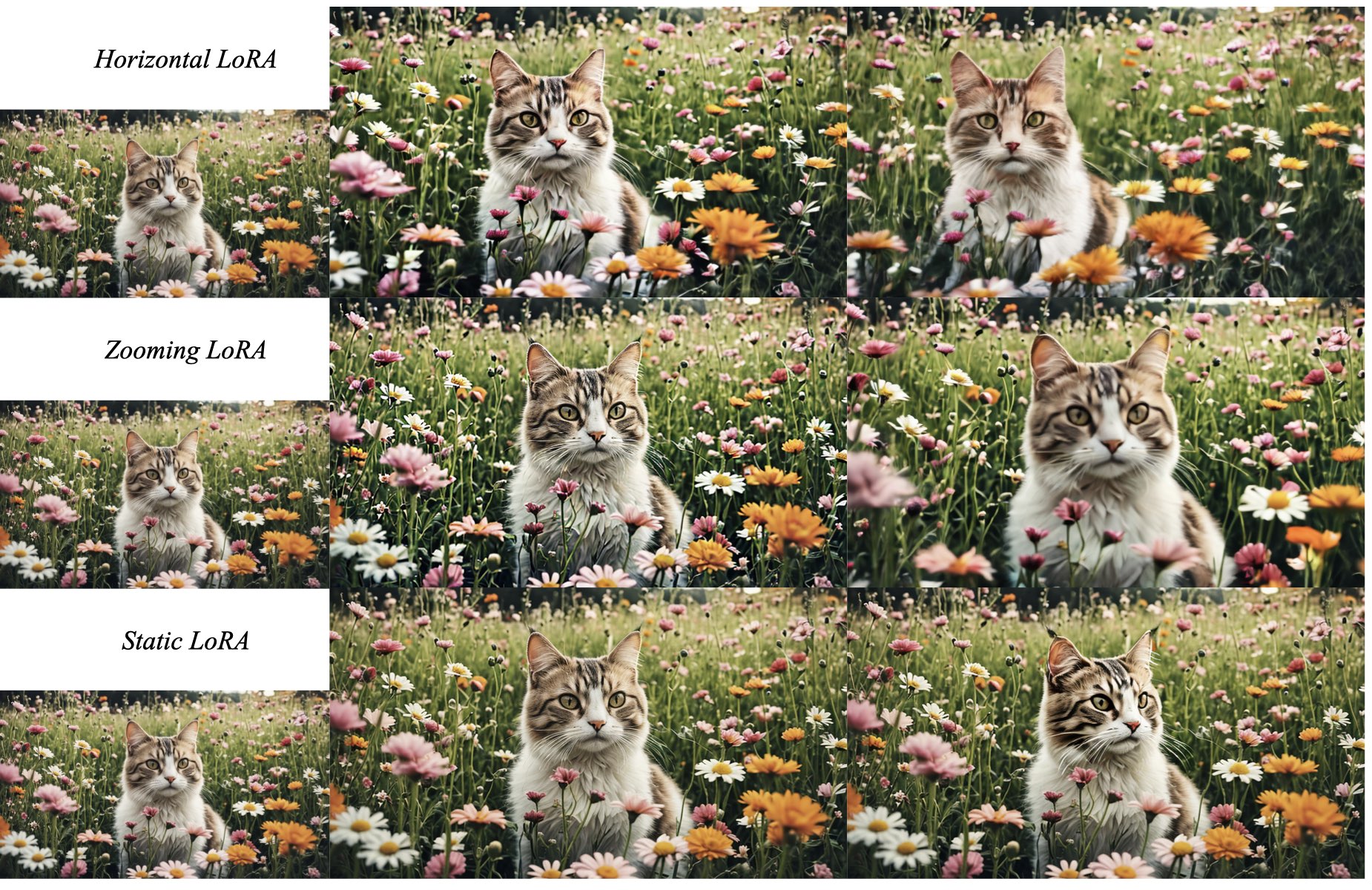}
    \caption{\small Applying three camera motion LoRAs (\emph{horizontal}, \emph{zooming}, \emph{static}) to the same conditioning frame (on the left).}
    \label{fig:motion_lora}
    \vspace{-10pt}
\end{figure}
}

\newcommand{\gsoresults}{
\begin{figure}
\begin{minipage}[b]{0.55\linewidth}
\centering
\hspace{-0.9em}
\resizebox*{!}{.58\columnwidth}{
\setlength{\tabcolsep}{1pt}
\begin{tabular}{lccc}
    \toprule
    \textbf{Method} & LPIPS$\downarrow$ & PSNR$\uparrow$ & CLIP-S$\uparrow$ \\ \midrule 
    SyncDreamer~\cite{liu2023syncdreamer} & 0.18 & 15.29 & 0.88 \\
    Zero123~\cite{liu2023zero1to3} & 0.18 & 14.87 & 0.87 \\ 
    Zero123XL~\cite{deitke2023objaversexl} & 0.20 & 14.51 & 0.87 \\
    \midrule
    Scratch-MV & 0.22 & 14.20 &  0.76 \\
    SD2.1-MV~\cite{rombach2021highresolution} & 0.18 & 15.06 & 0.83 \\
    \textbf{SVD-MV (\emph{ours})} & \textbf{0.14} & \textbf{16.83} & \textbf{0.89}\\
    \bottomrule
    \end{tabular}
    }
    \caption*{\small (a)}
    \label{tab:gso_multiview}
\end{minipage}
\hfill
\begin{minipage}[b]{0.42\linewidth}
\centering
\includegraphics[width=\linewidth]{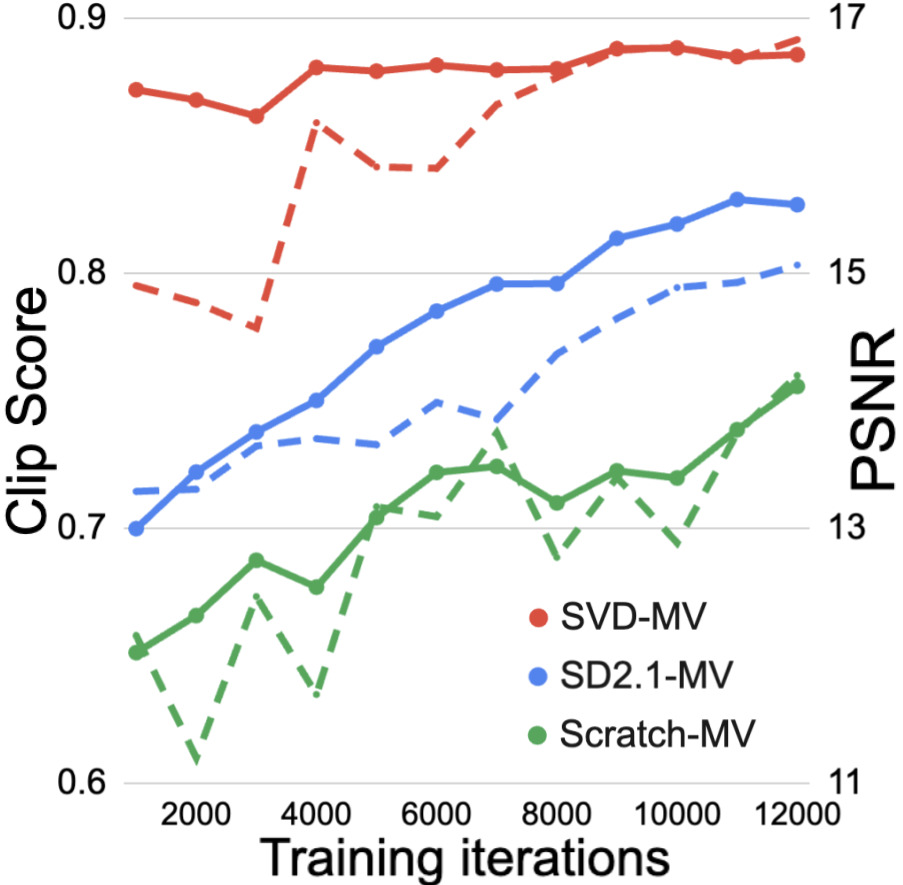}
\vspace{-2em}
\captionof{figure}*{\small (b)}
\label{fig:sotaimg2vid}
\end{minipage}
\vspace{-0.2em}
\caption{\small (a) Multi-view generation metrics on Google Scanned Objects (GSO) test dataset.
SVD-MV outperforms image-prior (SD2.1-MV) and no-prior (Scratch-MV) variants, as well other state-of-the-art techniques.
(b) Training progress of multi-view generation models with CLIP-S (solid, left axis) and PSNR (dotted, right axis) computed on GSO test dataset. 
SVD-MV shows better metrics consistently from the start of finetuning. 
}
\label{tab:gso_res}
\vspace{-0.2em}
\end{figure}
}

\newcommand{\additionalimagetwovideo}{
    \begin{figure*}
    \includegraphics[width=\linewidth]{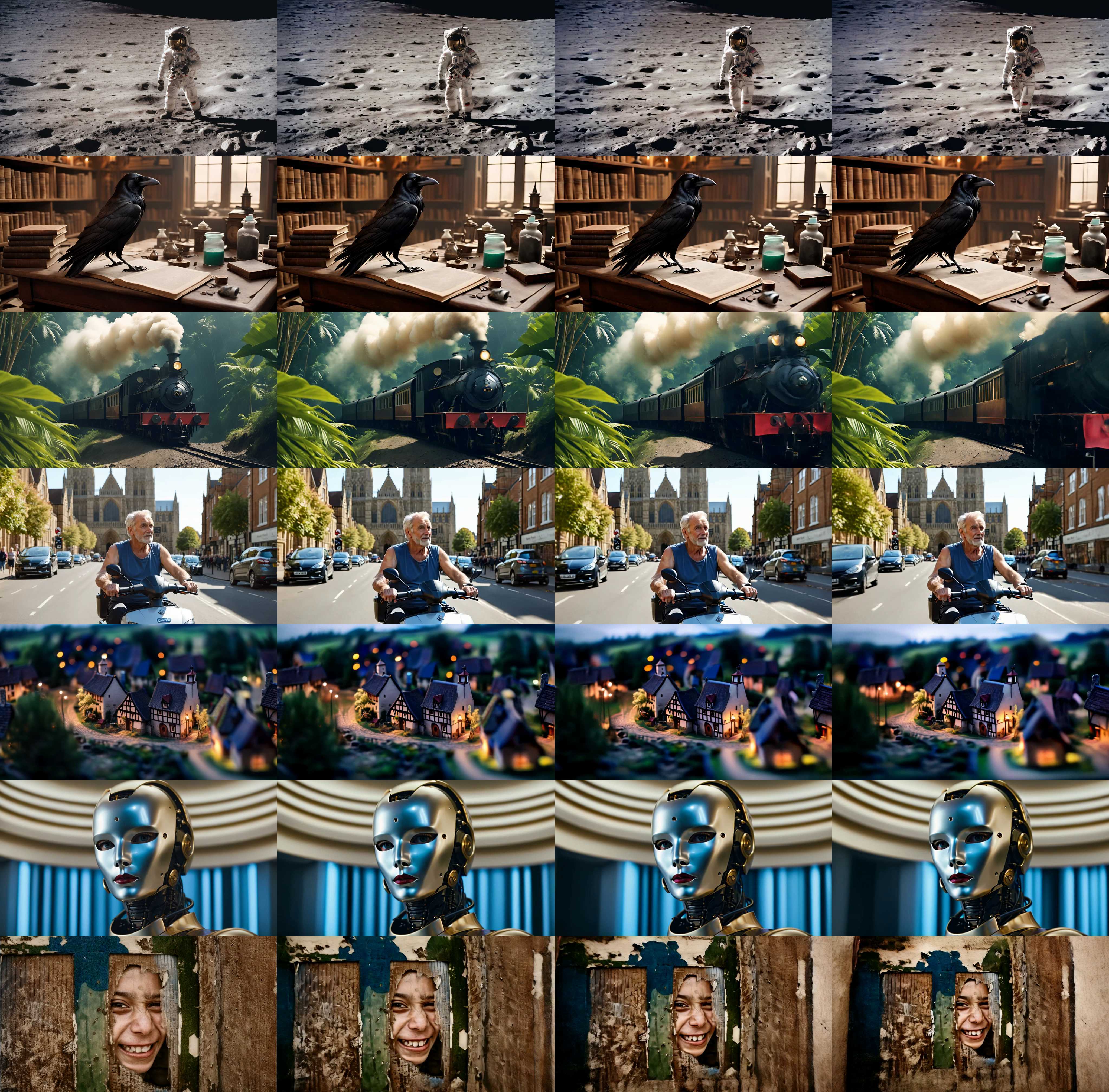}
    \caption{
    \label{fig:additional_img2vid}
        \small Additional Image-to-Video samples. Leftmost frame is use for conditioning.
    }
    \end{figure*}
}

\newcommand{\additionalmotionlora}{
    \begin{figure*}
    \includegraphics[width=\linewidth]{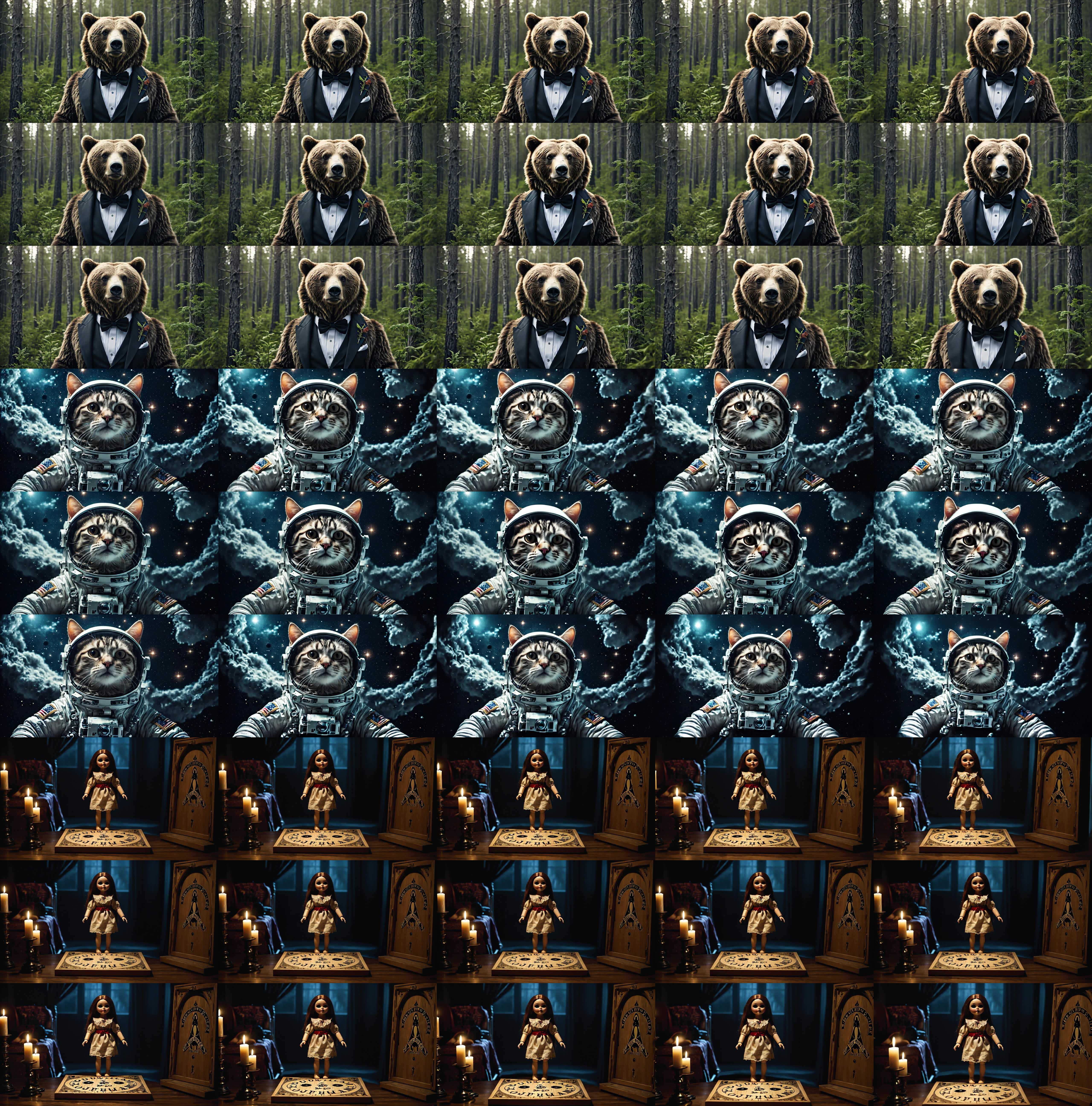}
    \caption{
        \small Additional Image-to-Video samples with camera motion LoRAs (conditioned on leftmost frame). The first, second, and thirs rows correspond to \emph{horizontal}, \emph{static}, \emph{zooming}, respectively.
    }
    \label{fig:additional_motion_lora}
    \end{figure*}
}

\newcommand{\additionaltemporalattention}{
    \begin{figure*}
    \includegraphics[width=\linewidth]{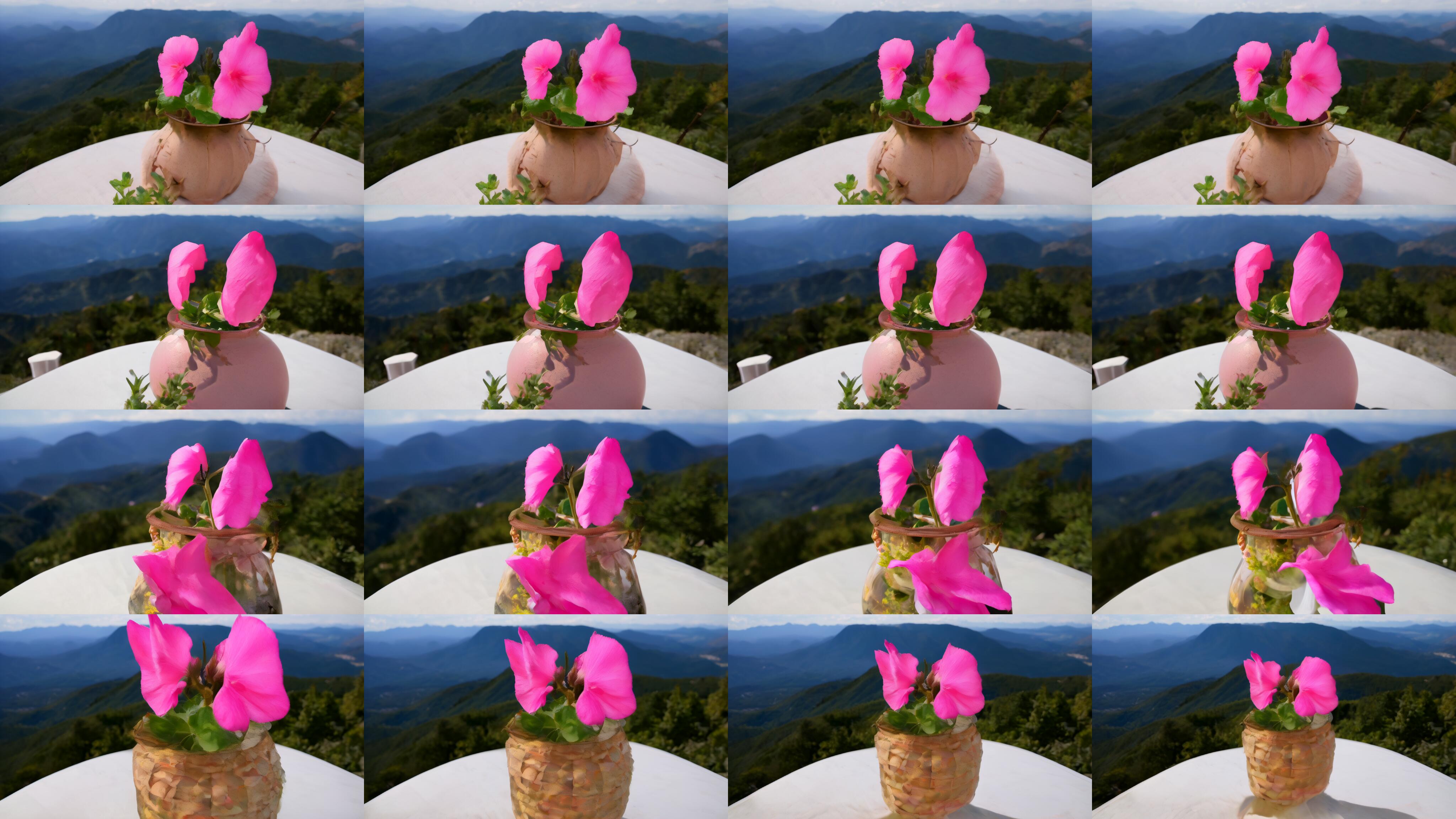}
    \caption{
    \label{fig:additional_temporal_attention}
        \small Text-to-video samples using the prompt ``Flowers in a pot in front of a mountainside'' (for spatial cross-attention). We adjust the camera control by replacing the prompt in the temporal attention using ``'', ``panning'', ``rotating'', and ``zooming'' (from top to bottom). While not being trained for this inference task, the model performs surprisingly well.
    }
    \end{figure*}
}

\newcommand{\additionaltexttwovideo}{
    \begin{figure*}
    \includegraphics[width=\linewidth]{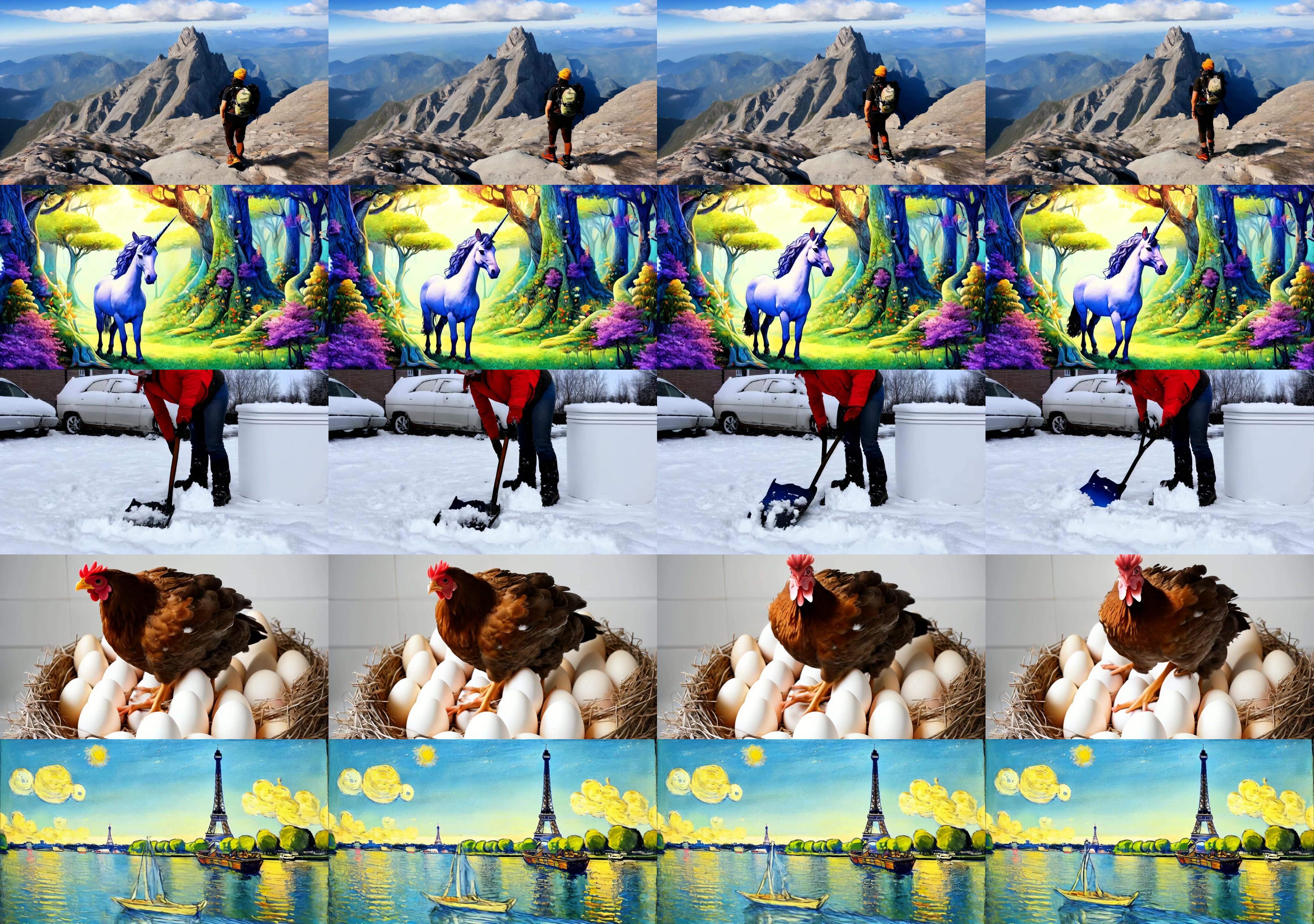}
    \caption{
    \label{fig:additional_txt2vid}
        \small Additional Text-to-Video samples. Captions from top to bottom: ``A hiker is reaching the summit of a mountain, taking in the breathtaking panoramic view of nature.'', ``A unicorn in a magical grove, extremely detailed.'', ``Shoveling snow'', ``A beautiful fluffy domestic hen sitting on white eggs in a brown nest, eggs are under the hen.'', and ``A boat sailing leisurely along the Seine River with the Eiffel Tower in background by Vincent van Gogh''.
    }
    \end{figure*}
}

\begin{document}
\teaserfigure
\begin{NoHyper}
  \let\thefootnote\relax\footnotetext{* Equal contributions.}
\end{NoHyper}
\begin{abstract}
\vspace{-0.3cm}

We present Stable Video Diffusion --- a latent video diffusion model for high-resolution, state-of-the-art text-to-video and image-to-video generation. Recently, latent diffusion models trained for 2D image synthesis have been turned into generative video models by inserting temporal layers and finetuning them on small, high-quality video datasets. However, training methods in the literature vary widely, and the field has yet to agree on a unified strategy for curating video data.
In this paper, we identify and evaluate three different stages for successful training of video LDMs: text-to-image pretraining, video pretraining, and high-quality video finetuning. Furthermore, we demonstrate the necessity of a well-curated pretraining dataset for generating high-quality videos and present a systematic curation process to train a strong base model, including captioning and filtering strategies. 
We then explore the impact of finetuning our base model on high-quality data and train a text-to-video model that is competitive with closed-source video generation. We also show that our base model provides a powerful motion representation for downstream tasks such as image-to-video generation and adaptability to camera motion-specific LoRA modules.
Finally, we demonstrate that our model provides a strong multi-view 3D-prior and can serve as a base to finetune a multi-view diffusion model that jointly generates multiple views of objects in a feedforward fashion, outperforming image-based methods at a fraction of their compute budget. %
We release code and model weights at \url{https://github.com/Stability-AI/generative-models}.

\end{abstract}

\section{Introduction}
\label{sec:intro}

Driven by advances in generative image modeling with diffusion models~\citep{ho2020ddpm, rameshdalle2, saharia2022imagen, rombach2021high}, there has been significant recent progress on generative video models both in research~\citep{ho2022video, singer2022make, voleti2022mcvd, blattmann2023align} and real-world applications~\citep{gen2,pika} 
Broadly, these models are either trained from scratch~\citep{ho2022imagenvideo} or finetuned (partially or fully) from pretrained image models with additional temporal layers inserted~\citep{blattmann2023align,guo2023animatediff,hong2022cogvideo,singer2022make}. Training is often carried out on a mix of image and video datasets~\citep{ho2022imagenvideo}. 

While research around improvements in video modeling has primarily focused on the exact arrangement of the spatial and temporal layers~\cite{hong2022cogvideo,singer2022make,ho2022imagenvideo,blattmann2023align}, none of the aforementioned works investigate the influence of data selection. This is surprising, especially since the significant impact of the training data distribution on generative models is undisputed~\citep{dai2023emu,xu2023demystifying}. Moreover, for generative image modeling, it is known that pretraining on a large and diverse dataset 
and finetuning on a smaller but higher quality dataset significantly improves the performance~\cite{rombach2021high,dai2023emu}. Since many previous approaches to video modeling have successfully drawn on techniques from the image domain~\cite{blattmann2023align,ho2022video,hong2022cogvideo}, it is noteworthy that the effect of data and training strategies, i.e., the separation of video pretraining at lower resolutions and high-quality finetuning, has yet to be studied. 
This work directly addresses these previously uncharted territories.
We believe that the significant contribution of \emph{data selection} is heavily underrepresented in today's video research landscape despite being well-recognized among practitioners when training video models at scale. Thus, in contrast to previous works, we draw on simple latent video diffusion baselines~\citep{blattmann2023align} for which we fix architecture and training scheme and assess the effect of \emph{data curation}. 
To this end, we first identify three different video training stages that we find crucial for good performance: text-to-image pretraining, \emph{video pretraining} on a large dataset at low resolution, and high-resolution \emph{video finetuning} on a much smaller dataset with higher-quality videos. Borrowing from large-scale image model training~\citep{radford2021learning,podell2023sdxl,dai2023emu}, we 
introduce a systematic approach to curate video data at scale and present an empirical study on the effect of data curation during video pretraining. Our main findings imply that pretraining on well-curated datasets leads to significant performance improvements that persist after high-quality finetuning.

\paragraph{A general motion and multi-view prior}
Drawing on these findings, we apply our proposed curation scheme to a large video dataset comprising roughly 600 million samples and train a strong pretrained text-to-video base model, which provides a general motion representation. 
We exploit this and finetune the base model on a smaller,  high-quality dataset for high-resolution downstream tasks such as text-to-video (see \Cref{fig:teaser}, top row) and image-to-video, where we predict a sequence of frames from a single conditioning image (see \Cref{fig:teaser}, mid rows). Human preference studies reveal that the resulting model outperforms state-of-the-art image-to-video models.

Furthermore, we also demonstrate that our model provides a strong multi-view prior and can serve as a base to finetune a multi-view diffusion model that generates multiple consistent views of an object in a feedforward manner and outperforms specialized novel view synthesis methods such as Zero123XL~\citep{liu2023zero1to3,deitke2023objaversexl} and SyncDreamer~\citep{liu2023syncdreamer}. %
Finally, we demonstrate that our model allows for explicit motion control by specifically prompting the temporal layers with motion cues and also via training LoRA-modules~\cite{hu2021lora,guo2023animatediff} on datasets resembling specific motions only, which can be efficiently plugged into the model. 
\newline 
To summarize, our core contributions are threefold:
    (i) We present a systematic data curation workflow to turn a large uncurated video collection into a quality dataset for generative video modeling.
    Using this workflow, we (ii) train state-of-the-art text-to-video and image-to-video models, outperforming all prior models.
    Finally, we (iii) probe the strong prior of motion and 3D understanding in our models by conducting domain-specific experiments. Specifically, we provide evidence that pretrained video diffusion models can be turned into strong multi-view generators, which may help overcome the data scarcity typically observed in the 3D domain~\cite{deitke2023objaversexl}.
\section{Background}
\label{sec:background}
Most recent works on video generation rely on diffusion models~\citep{sohl2015deep,ho2020ddpm,song2020score} to jointly synthesize multiple 
consistent frames from text- or image-conditioning. 
Diffusion models implement an iterative refinement process by learning to gradually denoise a sample from a normal distribution and have been successfully applied to
high-resolution text-to-image~\cite{rameshdalle2,saharia2021image,rombach2021high,podell2023sdxl,dai2023emu} and video synthesis~\cite{ho2022imagenvideo,singer2022make,voleti2022mcvd,blattmann2023align,ge2023preserve}. 

In this work, we follow this paradigm and train a latent~\citep{rombach2021high,vahdat2021score} video diffusion model~\citep{blattmann2023align,esser2023structure} on our video dataset. 
We provide a brief overview of related works which utilize latent video diffusion models (Video-LDMs) in the following paragraph; a full discussion that includes approaches using GANs~\citep{goodfellow2014generative,brooks2022generating} and
autoregressive models~\citep{hong2022cogvideo} can be found in~\Cref{supsec:related_work}. 
An introduction to diffusion models can be found in~\Cref{supsec:model_and_implementation_details}.

\paragraph{Latent Video Diffusion Models}
Video-LDMs~\citep{he2023latent,blattmann2023align,gu2023reuse,guo2023animatediff,wang2023modelscope} train the main generative model in a latent space of reduced computational complexity~\citep{esser2020taming,rombach2021high}.
Most related works make use of a pretrained text-to-image model and insert temporal mixing layers of various forms~\citep{an2023latent,ge2023preserve,blattmann2023align,gu2023reuse,guo2023animatediff} into the pretrained architecture.
\citet{ge2023preserve} additionally relies on temporally correlated noise to increase temporal consistency and ease the learning task.
In this work, we follow the architecture proposed in~\citet{blattmann2023align} and insert temporal convolution and attention layers after every spatial convolution and attention layer.
In contrast to works that only train temporal layers~\citep{guo2023animatediff,blattmann2023align} or are completely training-free~\citep{khachatryan2023text2videozero,zhang2023controlvideo}, we finetune the full model. 
For text-to-video synthesis in particular, most works directly condition the model on a text prompt~\citep{blattmann2023align,wang2023modelscope} or make use of an additional text-to-image prior~\cite{esser2023structure,singer2022make}.

In our work, we follow the former approach and show that the resulting model is a strong general motion prior, which can easily be finetuned into an image-to-video or multi-view synthesis model. 
Additionally, we introduce micro-conditioning~\citep{podell2023sdxl} on frame rate. 
We also employ the EDM-framework~\citep{karras2022elucidating} and significantly shift the noise schedule towards higher noise values, which we find to be essential for high-resolution finetuning. See \Cref{sec:sota} for a detailed discussion of the latter.

\paragraph{Data Curation}
Pretraining on %
large-scale datasets~\cite{schuhmann2022laion} is an essential ingredient for powerful models in several tasks such as discriminative text-image~\citep{xu2023demystifying,radford2021learning} and language~\citep{pile, refinedweb, 2019t5} modeling. 
By leveraging efficient language-image representations such as CLIP~\citep{ilharco2021open,radford2021learning,xu2023demystifying}, data curation has similarly been successfully applied for generative image modeling~\citep{schuhmann2022laion,dai2023emu,podell2023sdxl}. However, discussions on such data curation strategies have largely been missing in the video generation literature~\citep{ho2022imagenvideo, villegas2022phenaki, hong2022cogvideo, singer2022make}, and processing and filtering strategies have been introduced in an ad-hoc manner.
Among the publicly accessible video datasets, WebVid-10M~\citep{bain2022frozen} dataset has been a popular choice~\citep{singer2022make,blattmann2023align,zhou2022magicvideo} despite being watermarked and suboptimal in size. 
Additionally, WebVid-10M is often used in combination with image data~\cite{schuhmann2022laion}, to enable joint image-video training. However, this amplifies the difficulty of separating the effects of image and video data on the final model. 
To address these shortcomings, this work presents a systematic study of methods for video data curation and further introduces a general three-stage training strategy for generative video models, producing a state-of-the-art model.

\section{Curating Data for HQ Video Synthesis}
\label{sec:approach}
\noindent In this section, we introduce a general strategy to train a state-of-the-art video diffusion model on large datasets of videos. 
To this end, we (i) introduce data processing and curation methods, for which we systematically analyze the impact on the quality of the final model in \Cref{subsec:data_curation} and \Cref{subsec:stage3}, and (ii), identify three different training regimes for generative video modeling. %
In particular, these regimes consist of 
\begin{itemize}
    \item Stage I: \emph{image pretraining}, \ie a 2D text-to-image diffusion model~\cite{rombach2021high,podell2023sdxl,dai2023emu}. %
    \item Stage II: \emph{video pretraining}, which trains on large amounts of videos.
    \item Stage III: \emph{video finetuning}, which refines the model on a small subset of high-quality videos at higher resolution.
\end{itemize}
We study the importance of each regime separately in \Cref{subsec:image_pretrain,subsec:data_curation,subsec:stage3}.

\subsection{Data Processing and Annotation}
\label{subsec:data_proc}
\clippingflow
\noindent  We collect an initial dataset of long videos 
which forms the base data for our \emph{video pretraining} stage.
To avoid cuts and fades leaking into synthesized videos, we apply a cut detection pipeline\footnote{\url{https://github.com/Breakthrough/PySceneDetect}} in a cascaded manner at three different FPS levels. 
\Cref{fig:cuts_and_motion}, left, provides evidence for the need for cut detection: After applying our cut-detection pipeline, we obtain a significantly higher number ($\sim\hspace{-0.3em}4\times$) of clips, indicating that many video clips in the unprocessed dataset contain cuts beyond those obtained from metadata.

Next, we annotate each clip with three different synthetic captioning methods: First, we use the image captioner CoCa~\cite{yu2022coca} to annotate the mid-frame of each clip and use V-BLIP~\citep{videoblip} to obtain a video-based caption. Finally, we generate a third description of the clip via an LLM-based summarization of the first two captions. 

The resulting initial dataset, which we dub \emph{Large Video Dataset} (\dataset), consists of 580M annotated video clip pairs, forming \emph{212 years} of content.  
\datastats

However, further investigation reveals that the resulting dataset contains examples that can be expected to degrade the performance of our final video model, 
such as clips with less motion, excessive text presence, or generally low aesthetic value. 
We therefore additionally annotate our dataset with dense optical flow~\citep{farneback2003flow,itseez2015opencv}, which we calculate at 2 FPS and with which we filter out static scenes by removing any videos whose average optical flow magnitude is below a certain threshold.
Indeed, when considering the motion distribution of \dataset (see \Cref{fig:cuts_and_motion}, right) via optical flow scores, we identify a subset of close-to-static clips therein.

Moreover, we apply optical character recognition~\cite{baek2019character} to weed out clips containing large amounts of written text. 
Lastly, we annotate the first, middle, and last frames of each clip with CLIP~\citep{radford2021learning} embeddings 
from which we calculate aesthetics scores~\citep{schuhmann2022laion} as well as text-image similarities. %
Statistics of our dataset, including the total size and average duration of clips, are provided in~\Cref{tab:subset_stats}.

\subsection{Stage I: Image Pretraining}
\label{subsec:image_pretrain}
We consider image pretraining as the first stage in our training pipeline.
Thus, in line with concurrent work on %
video models~\citep{singer2022make,ho2022imagenvideo,blattmann2023align}, we ground our initial model on a pretrained image diffusion model - namely \emph{Stable Diffusion 2.1}~\citep{rombach2021high}  - to equip it with a strong visual representation. 

To analyze the effects of image pretraining,  we train and compare two identical video models as detailed in~\Cref{supsec:model_and_implementation_details} on a 10M subset of \dataset; one with and one without pretrained spatial weights. 
We compare these models using a human preference study (see~\Cref{supsec:experiment_details} for details) in \Cref{fig:imageonly_comp}, which clearly shows that the image-pretrained model is preferred in both quality and prompt-following.
\imageonlyandfiltering
\subsection{Stage II: Curating a Video Pretraining Dataset}
\label{subsec:data_curation}
\textbf{A systematic approach to video data curation.} 
For multimodal image modeling, data curation is a key element of many powerful discriminative~\citep{radford2021learning,xu2023demystifying} and generative~\citep{dai2023emu,ho2022imagen,ramesh2022dalle2} models. 
However, since there are no equally powerful off-the-shelf representations available in the video domain to filter out unwanted examples, we rely on human preferences as a signal to create a suitable pretraining dataset.  
Specifically, we curate subsets of \dataset using different methods described below and then consider the human-preference-based ranking of latent video diffusion models trained on these datasets.
\bigfatplot
\vspace{1em}

More specifically, for each type of annotation introduced in \Cref{subsec:data_proc} (\ie, CLIP scores, aesthetic scores, OCR detection rates, synthetic captions, optical flow scores), we start from an unfiltered, randomly sampled 9.8M-sized subset of \dataset, \datasetsmall, and systematically remove the bottom 12.5, 25 and 50\% of examples. Note that for the synthetic captions, we cannot filter in this sense. Instead, we assess Elo rankings~\cite{elo1978rating} for the different captioning methods from \Cref{subsec:data_proc}. 
To keep the number of total subsets tractable, we apply this scheme separately to each type of annotation. %
We train models with the same training hyperparameters on each of these filtered subsets and compare the results of all models within the same class of annotation with an Elo ranking~\cite{elo1978rating} for human preference votes. 
Based on these votes, we consequently select the best-performing filtering threshold for each annotation type. The details of this study are presented and discussed in \Cref{supsec:experiment_details}. 
Applying this filtering approach to \dataset results in a final pretraining dataset of 152M training examples, which we refer to as \datasetfiltered, \cf \Cref{tab:subset_stats}.

\textbf{Curated training data improves performance.}
In this section, we demonstrate that the data curation approach described above improves the training of our video diffusion models.
To show this, we apply the filtering strategy described above to \datasetsmall and obtain a four times smaller subset, \datasetsmallfiltered. Next, 
we
use it to train a baseline model that follows our standard architecture and training schedule and evaluate the preference scores for visual quality and prompt-video alignment compared to a model trained on uncurated \datasetsmall. 

We visualize the results in~\Cref{fig:filtering_effects}, where we can see the benefits of filtering: In both categories, the model trained on the much smaller \datasetsmallfiltered is preferred. To further show the efficacy of our curation approach, we compare the model trained on \datasetsmallfiltered with similar video models trained on WebVid-10M~\citep{bain2022frozen}, which is the most recognized research licensed dataset, and InternVid-10M~\citep{wang2023internvid}, which is specifically filtered for high aesthetics. Although \datasetsmallfiltered is also four times smaller than these datasets, the corresponding model is preferred by human evaluators in both spatiotemporal quality and prompt alignment as shown in \Cref{fig:internvid_comp,fig:internvid_comp}.

\textbf{Data curation helps at scale.}
To verify that our data curation strategy from above also works on larger, more practically relevant datasets, 
we repeat the experiment above and train a video diffusion model on a filtered subset with 50M examples and a non-curated one of the same size. 
We conduct a human preference study and summarize the results of this study in \Cref{fig:more_data}, 
where we can see that the advantages of data curation also come into play with larger amounts of data. 
Finally, we show that dataset size is also a crucial factor when training on curated data in \Cref{fig:scaled_mid}, 
where a model trained on 50M curated samples is superior to a model trained on \datasetsmallfiltered for the same number of steps.      

\subsection{Stage III: High-Quality Finetuning}
\label{subsec:stage3}
In the previous section, we demonstrated the beneficial effects of systematic data curation for \emph{video pretraining}. However, since we are primarily interested in optimizing the performance after \emph{video finetuning}, we now investigate how these differences after Stage II translate to the final performance after Stage III. Here, we draw on training techniques from latent image diffusion modeling~\citep{podell2023sdxl, dai2023emu} and increase the resolution of the training examples. Moreover, we use a small finetuning dataset comprising 250K pre-captioned video clips of high visual fidelity.

To analyze the influence of \emph{video pretraining} on this last stage, we finetune three identical models, which only differ in their initialization. We initialize the weights of the first with a pretrained image model and skip \emph{video pretraining}, a common choice among many recent video modeling approaches~\citep{blattmann2023align,singer2022make}. The remaining two models are initialized with the weights of the latent video models from the previous section, specifically, the ones trained on 50M curated and uncurated video clips. We finetune all models for 50K steps and assess human preference rankings early during finetuning (10K steps) and at the end to measure how performance differences progress in the course of finetuning. We show the obtained results in \Cref{fig:finetune_elos}, where we plot the Elo improvements of user preference relative to the model ranked last, which is the one initialized from an image model. Moreover, the finetuning resumed from curated pretrained weights ranks consistently higher than the one initialized from video weights after uncurated training. 

Given these results, we conclude that i) the separation of video model training in \emph{video pretraining} and \emph{video finetuning} is beneficial for the final model performance after finetuning and that ii) \emph{video pretraining} should ideally occur on a large scale, curated dataset, since performance differences after pretraining persist after finetuning.

\imgtwovid
\section{Training Video Models at Scale}
\label{sec:sota}

In this section, we borrow takeaways from~\Cref{sec:approach} and present results of training state-of-the-art video models at scale. We first use the optimal data strategy inferred from ablations to train a powerful base model at $320\times576$ in~\Cref{sec:base-model}. We then perform finetuning to yield several strong state-of-the-art models for different tasks such as text-to-video in~\Cref{sec:txt2vid}, image-to-video in~\Cref{sec:img2vid} and frame interpolation in~\Cref{sec:interpolation}. 
Finally, we demonstrate that our video-pretraining can serve as a strong implicit 3D prior, by tuning our image-to-video models on multi-view generation in \Cref{sec:multiview} and outperform concurrent work, in particular Zero123XL~\cite{deitke2023objaversexl, liu2023zero1to3} and SyncDreamer~\cite{liu2023syncdreamer} in terms of multi-view consistency.

\subsection{Pretrained Base Model}
\label{sec:base-model}
\ucfzeroshotandsotavid
As discussed in~\Cref{subsec:image_pretrain}, our video model is based on \emph{Stable Diffusion 2.1}~\citep{rombach2021high} (SD 2.1). 
Recent works~\citep{hoogeboom2023simple} show that it is crucial to adopt the noise schedule when training image diffusion models, shifting towards more noise for higher-resolution images. 
As a first step, we finetune the fixed discrete noise schedule from our image model towards continuous noise~\citep{song2020score} using the network preconditioning proposed in~\citet{karras2022elucidating} for images of size $256\times384$. After inserting temporal layers, we then train the model on \datasetfiltered on 14 frames at resolution $256\times384$. We use the standard EDM noise schedule~\citep{karras2022elucidating} for 150k iterations and batch size 1536. Next, we finetune the model to generate 14 $320\times576$ frames for 100k iterations using batch size 768. We find that it is important to shift the noise schedule towards more noise for this training stage, confirming results by~\citet{hoogeboom2023simple} for image models. For further training details, see~\Cref{supsec:model_and_implementation_details}. 
We refer to this model as our \emph{base model} which can be easily finetuned for a variety of tasks as we show in the following sections. 
The base model has learned a powerful motion representation, for example, it significantly outperforms all baselines for zero-shot text-to-video generation on UCF-101~\citep{soomro2012ucf101} (\Cref{tab:ucf}). 
Evaluation details can be found in~\Cref{supsec:experiment_details}.
\subsection{High-Resolution Text-to-Video Model}
\label{sec:txt2vid}
We finetune the base text-to-video model on a high-quality video dataset of $\sim$ 1M samples. Samples in the dataset generally contain lots of object motion, steady camera motion, and well-aligned captions, and are of high visual quality altogether. We finetune our base model for 50k iterations at resolution $576\times1024$ (again shifting the noise schedule towards more noise) using batch size 768. Samples in~\Cref{fig:teaser2}, more can be found in~\Cref{supsec:experiment_details}.
\subsection{High Resolution Image-to-Video Model}
\label{sec:img2vid}
Besides text-to-video, we finetune our base model for image-to-video generation, where the video model receives a still input image as a conditioning. 
Accordingly, we replace text embeddings that are fed into the base model with the CLIP image embedding of the conditioning.
Additionally, we concatenate a noise-augmented~\citep{ho2021cascaded} version of the conditioning frame channel-wise to the input of the UNet~\cite{ronneberger2015u}. We do not use any masking techniques and simply copy the frame across the time axis.
We finetune two models, one predicting 14 frames and another one predicting 25 frames; implementation and training details can be found in~\Cref{supsec:model_and_implementation_details}. 
We occasionally found that standard vanilla classifier-free guidance~\citep{ho2021classifierfree} can lead to artifacts: too little guidance may result in inconsistency with the conditioning frame while too much guidance can result in oversaturation. 
Instead of using a constant guidance scale, we found it helpful to linearly increase the guidance scale across the frame axis (from small to high). Details can be found in~\Cref{supsec:model_and_implementation_details}. Samples in~\Cref{fig:teaser2}, more can be found in~\Cref{supsec:experiment_details}.

In~\Cref{fig:sotaimg2vid} we compare our model with state-of-the-art, closed-source video generative models, in particular GEN-2~\citep{esser2023structure,gen2} and PikaLabs~\citep{pika}, and show that our model is preferred in terms of visual quality by human voters. 
Details on the experiment, as well as many more image-to-video samples, can be found in~\Cref{supsec:experiment_details}.
\vspace{-10pt}
\subsubsection{Camera Motion LoRA}
\label{subsec:motion_lora}
\motionlora
To facilitate controlled camera motion in image-to-video generation, we train a variety of \emph{camera motion LoRAs} within the temporal attention blocks of our model~\citep{guo2023animatediff}; see~\Cref{supsec:model_and_implementation_details} for exact implementation details. We train these additional parameters on a small dataset with rich camera-motion metadata. In particular, we use three subsets of the data for which the camera motion is categorized as ``horizontally moving'', ``zooming'', and ``static''. In~\Cref{fig:motion_lora} we show samples of the three models for identical conditioning frames; more samples can be found in~\Cref{supsec:experiment_details}.
\subsection{Frame Interpolation} \label{sec:interpolation}
To obtain smooth videos at high frame rates, we finetune our high-resolution text-to-video model into a frame interpolation model. We follow~\citet{blattmann2023align} and concatenate the left and right frames to the input of the UNet via~\emph{masking}. The model learns to predict three frames within the two conditioning frames, effectively increasing the frame rate by four. Surprisingly, we found that a very small number of iterations ($\approx 10k$) suffices to get a good model. Details and samples can be found in~\Cref{supsec:model_and_implementation_details} and~\Cref{supsec:experiment_details}, respectively.
\subsection{Multi-View Generation}
\label{sec:multiview}
\vidtothreedfig
To obtain multiple novel views of an object simultaneously, we finetune our image-to-video SVD model on multi-view datasets~\cite{deitke2023objaverse,deitke2023objaversexl,yu2023mvimgnet}.

\vspace{2pt}

\noindent \textbf{Datasets.}
We finetuned our SVD model on two datasets, where the SVD model takes a single image and outputs a sequence of multi-view images: (i) A subset of Objaverse~\cite{deitke2023objaverse} consisting of 150K curated and CC-licensed synthetic 3D objects from the original dataset~\cite{deitke2023objaverse}. For each object, we rendered $360^\circ$ orbital videos of 21 frames with randomly sampled HDRI environment map and elevation angles between $[-5^\circ,30^\circ]$. We evaluate the resulting models on an unseen test dataset consisting of 50 sampled objects from  Google Scanned Objects (GSO) dataset~\cite{downs2022google}.
and (ii) MVImgNet~\cite{yu2023mvimgnet} consisting of casually captured multi-view videos of general household objects. We split the videos into $\sim$200K train and 900 test videos. We rotate the frames captured in portrait mode to landscape orientation.

The Objaverse-trained model is additionally conditioned on the elevation angle of the input image, and outputs orbital videos at that elevation angle. The MVImgNet-trained models are not conditioned on pose and can choose an arbitrary camera path in their generations. For details on the pose conditioning mechanism, see \Cref{supsec:experiment_details}.

\vspace{2pt}

\noindent \textbf{Models}.
We refer to our finetuned Multi-View model as SVD-MV. We perform an ablation study on the importance of the video prior of SVD for multi-view generation. To this effect, we compare the results from SVD-MV i.e. from a video prior to those finetuned from an image prior i.e. the text-to-image model SD2.1 (SD2.1-MV), and that trained without a prior i.e. from random initialization (Scratch-MV). In addition, we compare with the current state-of-the-art multiview generation models of Zero123~\cite{liu2023zero1to3}, Zero123XL~\cite{deitke2023objaversexl}, and SyncDreamer~\cite{liu2023syncdreamer}.

\vspace{2pt}

\noindent \textbf{Metrics}.
We use the standard metrics of Peak Signal-to-Noise Ratio (PSNR), LPIPS~\cite{zhang2018unreasonable}, and CLIP~\cite{radford2021learning} Similarity scores (CLIP-S) between the corresponding pairs of ground truth and generated frames on 50 GSO test objects.

\vspace{2pt}

\noindent \textbf{Training}. We train all our models for 12k steps ($\sim$16 hours) with 8 80GB A100 GPUs using a total batch size of 16, with a learning rate of 1e-5.

\vspace{2pt}

\noindent \textbf{Results.}
\Cref{tab:gso_res}(a) shows the average metrics on the GSO test dataset. The higher performance of SVD-MV compared to SD2.1-MV and Scratch-MV clearly demonstrates the advantage of the learned video prior in the SVD model for multi-view generation. In addition, as in the case of other models finetuned from SVD, we found that a very small number of iterations ($\approx 12k$) suffices to get a good model. Moreover, SVD-MV is competitive w.r.t state-of-the-art techniques with lesser training time (12$k$ iterations in 16 hours), whereas existing models are typically trained for much longer (for example, SyncDreamer was trained for four days specifically on Objaverse). \Cref{tab:gso_res}(b) shows convergence of different finetuned models. After only 1k iterations, SVD-MV has much better CLIP-S and PSNR scores than its image-prior and no-prior counterparts.

\Cref{fig:vidto3dfig} shows a qualitative comparison of multi-view generation results on a GSO test object and \Cref{fig:MVIfig} on an MVImgNet test object. As can be seen, our generated frames are multi-view consistent and realistic.
More details on the experiments, as well as more multi-view generation samples, can be found in~\Cref{supsec:experiment_details}.

\gsoresults
\MVIfig

\section{Conclusion}

We present \emph{Stable Video Diffusion} (SVD), a latent video diffusion model for high-resolution, state-of-the-art text-to-video and image-to-video synthesis. To construct its pretraining dataset, we conduct a systematic data selection and scaling study, and propose a method to curate vast amounts of video data and turn large and noisy video collection into suitable datasets for generative video models. Furthermore, we introduce three distinct stages of video model training which we separately analyze to assess their impact on the final model performance. \emph{Stable Video Diffusion} provides a powerful video representation from which we finetune video models for state-of-the-art image-to-video synthesis and other highly relevant applications such as LoRAs for camera control. Finally we provide a pioneering study on multi-view finetuning of video diffusion models and show that SVD constitutes a strong 3D prior, which obtains state-of-the-art results in multi-view synthesis while using only a fraction of the compute of previous methods. 

We hope these findings will be broadly useful in the generative video modeling literature. A discussion on our work's broader impact and limitations can be found in~\Cref{supsec:broader_impact_and_limitations}.
\section*{Acknowledgements}
Special thanks to Emad Mostaque for his excellent support on this project. Many thanks go to our colleagues Jonas M\"uller, Axel Sauer, Dustin Podell and Rahim Entezari for fruitful discussions and comments. Finally, we thank Harry Saini and the one and only Richard Vencu for maintaining and optimizing our data and computing infrastructure.

{
    \small
    \bibliographystyle{ieeenat_fullname}
    \bibliography{arxiv,non_arxiv,postings,old}

\begin{thebibliography}{116}
\providecommand{\natexlab}[1]{#1}
\providecommand{\url}[1]{\texttt{#1}}
\expandafter\ifx\csname urlstyle\endcsname\relax
  \providecommand{\doi}[1]{doi: #1}\else
  \providecommand{\doi}{doi: \begingroup \urlstyle{rm}\Url}\fi

\bibitem[An et~al.(2023)An, Zhang, Yang, Gupta, Huang, Luo, and
  Yin]{an2023latent}
Jie An, Songyang Zhang, Harry Yang, Sonal Gupta, Jia-Bin Huang, Jiebo Luo, and
  Xi Yin.
\newblock Latent-shift: Latent diffusion with temporal shift for efficient
  text-to-video generation.
\newblock \emph{arXiv preprint arXiv:2304.08477}, 2023.

\bibitem[Anciukevi{\v{c}}ius et~al.(2023)Anciukevi{\v{c}}ius, Xu, Fisher,
  Henderson, Bilen, Mitra, and Guerrero]{anciukevivcius2023renderdiffusion}
Titas Anciukevi{\v{c}}ius, Zexiang Xu, Matthew Fisher, Paul Henderson, Hakan
  Bilen, Niloy~J Mitra, and Paul Guerrero.
\newblock Renderdiffusion: Image diffusion for 3d reconstruction, inpainting
  and generation.
\newblock In \emph{Proceedings of the IEEE/CVF Conference on Computer Vision
  and Pattern Recognition}, pages 12608--12618, 2023.

\bibitem[Askell et~al.(2021)Askell, Bai, Chen, Drain, Ganguli, Henighan, Jones,
  Joseph, Mann, DasSarma, Elhage, Hatfield-Dodds, Hernandez, Kernion, Ndousse,
  Olsson, Amodei, Brown, Clark, McCandlish, Olah, and
  Kaplan]{askell2021general}
Amanda Askell, Yuntao Bai, Anna Chen, Dawn Drain, Deep Ganguli, Tom Henighan,
  Andy Jones, Nicholas Joseph, Ben Mann, Nova DasSarma, Nelson Elhage, Zac
  Hatfield-Dodds, Danny Hernandez, Jackson Kernion, Kamal Ndousse, Catherine
  Olsson, Dario Amodei, Tom Brown, Jack Clark, Sam McCandlish, Chris Olah, and
  Jared Kaplan.
\newblock A general language assistant as a laboratory for alignment, 2021.

\bibitem[Babaeizadeh et~al.(2018)Babaeizadeh, Finn, Erhan, Campbell, and
  Levine]{babaeizadeh2018stochastic}
Mohammad Babaeizadeh, Chelsea Finn, Dumitru Erhan, Roy~H. Campbell, and Sergey
  Levine.
\newblock Stochastic variational video prediction.
\newblock In \emph{International Conference on Learning Representations}, 2018.

\bibitem[Baek et~al.(2019)Baek, Lee, Han, Yun, and Lee]{baek2019character}
Youngmin Baek, Bado Lee, Dongyoon Han, Sangdoo Yun, and Hwalsuk Lee.
\newblock Character region awareness for text detection.
\newblock In \emph{Proceedings of the IEEE/CVF conference on computer vision
  and pattern recognition}, pages 9365--9374, 2019.

\bibitem[Bai et~al.(2022)Bai, Jones, Ndousse, Askell, Chen, DasSarma, Drain,
  Fort, Ganguli, Henighan, Joseph, Kadavath, Kernion, Conerly, El-Showk,
  Elhage, Hatfield-Dodds, Hernandez, Hume, Johnston, Kravec, Lovitt, Nanda,
  Olsson, Amodei, Brown, Clark, McCandlish, Olah, Mann, and
  Kaplan]{bai2022training}
Yuntao Bai, Andy Jones, Kamal Ndousse, Amanda Askell, Anna Chen, Nova DasSarma,
  Dawn Drain, Stanislav Fort, Deep Ganguli, Tom Henighan, Nicholas Joseph,
  Saurav Kadavath, Jackson Kernion, Tom Conerly, Sheer El-Showk, Nelson Elhage,
  Zac Hatfield-Dodds, Danny Hernandez, Tristan Hume, Scott Johnston, Shauna
  Kravec, Liane Lovitt, Neel Nanda, Catherine Olsson, Dario Amodei, Tom Brown,
  Jack Clark, Sam McCandlish, Chris Olah, Ben Mann, and Jared Kaplan.
\newblock Training a helpful and harmless assistant with reinforcement learning
  from human feedback, 2022.

\bibitem[Bain et~al.(2022)Bain, Nagrani, Varol, and Zisserman]{bain2022frozen}
Max Bain, Arsha Nagrani, Gül Varol, and Andrew Zisserman.
\newblock Frozen in time: A joint video and image encoder for end-to-end
  retrieval, 2022.

\bibitem[Blattmann et~al.(2021)Blattmann, Milbich, Dorkenwald, and
  Ommer]{ipoke}
Andreas Blattmann, Timo Milbich, Michael Dorkenwald, and Bj{\"{o}}rn Ommer.
\newblock ipoke: Poking a still image for controlled stochastic video
  synthesis.
\newblock In \emph{2021 {IEEE/CVF} International Conference on Computer Vision,
  {ICCV} 2021, Montreal, QC, Canada, October 10-17, 2021}, 2021.

\bibitem[Blattmann et~al.(2023)Blattmann, Rombach, Ling, Dockhorn, Kim, Fidler,
  and Kreis]{blattmann2023align}
Andreas Blattmann, Robin Rombach, Huan Ling, Tim Dockhorn, Seung~Wook Kim,
  Sanja Fidler, and Karsten Kreis.
\newblock {Align your Latents: High-Resolution Video Synthesis with Latent
  Diffusion Models}.
\newblock \emph{arXiv:2304.08818}, 2023.

\bibitem[Brooks et~al.(2022)Brooks, Hellsten, Aittala, Wang, Aila, Lehtinen,
  Liu, Efros, and Karras]{brooks2022generating}
Tim Brooks, Janne Hellsten, Miika Aittala, Ting-Chun Wang, Timo Aila, Jaakko
  Lehtinen, Ming-Yu Liu, Alexei~A Efros, and Tero Karras.
\newblock Generating long videos of dynamic scenes.
\newblock In \emph{NeurIPS}, 2022.

\bibitem[Carreira and Zisserman(2017)]{carreira2017quo}
Joao Carreira and Andrew Zisserman.
\newblock Quo vadis, action recognition? a new model and the kinetics dataset.
\newblock In \emph{proceedings of the IEEE Conference on Computer Vision and
  Pattern Recognition}, pages 6299--6308, 2017.

\bibitem[Castrejon et~al.(2019)Castrejon, Ballas, and Courville]{hvrnn}
Lluis Castrejon, Nicolas Ballas, and Aaron Courville.
\newblock Improved conditional vrnns for video prediction.
\newblock In \emph{The IEEE International Conference on Computer Vision
  (ICCV)}, 2019.

\bibitem[Dai et~al.(2023)Dai, Hou, Ma, Tsai, Wang, Wang, Zhang, Vandenhende,
  Wang, Dubey, Yu, Kadian, Radenovic, Mahajan, Li, Zhao, Petrovic, Singh,
  Motwani, Wen, Song, Sumbaly, Ramanathan, He, Vajda, and Parikh]{dai2023emu}
Xiaoliang Dai, Ji Hou, Chih-Yao Ma, Sam Tsai, Jialiang Wang, Rui Wang, Peizhao
  Zhang, Simon Vandenhende, Xiaofang Wang, Abhimanyu Dubey, Matthew Yu,
  Abhishek Kadian, Filip Radenovic, Dhruv Mahajan, Kunpeng Li, Yue Zhao, Vladan
  Petrovic, Mitesh~Kumar Singh, Simran Motwani, Yi Wen, Yiwen Song, Roshan
  Sumbaly, Vignesh Ramanathan, Zijian He, Peter Vajda, and Devi Parikh.
\newblock Emu: Enhancing image generation models using photogenic needles in a
  haystack, 2023.

\bibitem[Deitke et~al.(2023{\natexlab{a}})Deitke, Liu, Wallingford, Ngo,
  Michel, Kusupati, Fan, Laforte, Voleti, Gadre, et~al.]{deitke2023objaversexl}
Matt Deitke, Ruoshi Liu, Matthew Wallingford, Huong Ngo, Oscar Michel, Aditya
  Kusupati, Alan Fan, Christian Laforte, Vikram Voleti, Samir~Yitzhak Gadre,
  et~al.
\newblock Objaverse-{XL}: A universe of 10m+ 3d objects.
\newblock \emph{arXiv preprint arXiv:2307.05663}, 2023{\natexlab{a}}.

\bibitem[Deitke et~al.(2023{\natexlab{b}})Deitke, Schwenk, Salvador, Weihs,
  Michel, VanderBilt, Schmidt, Ehsani, Kembhavi, and
  Farhadi]{deitke2023objaverse}
Matt Deitke, Dustin Schwenk, Jordi Salvador, Luca Weihs, Oscar Michel, Eli
  VanderBilt, Ludwig Schmidt, Kiana Ehsani, Aniruddha Kembhavi, and Ali
  Farhadi.
\newblock Objaverse: A universe of annotated 3d objects.
\newblock In \emph{Proceedings of the IEEE/CVF Conference on Computer Vision
  and Pattern Recognition}, pages 13142--13153, 2023{\natexlab{b}}.

\bibitem[Deng et~al.(2023)Deng, Jiang, Qi, Yan, Zhou, Guibas, Anguelov,
  et~al.]{deng2023nerdi}
Congyue Deng, Chiyu Jiang, Charles~R Qi, Xinchen Yan, Yin Zhou, Leonidas
  Guibas, Dragomir Anguelov, et~al.
\newblock Nerdi: Single-view nerf synthesis with language-guided diffusion as
  general image priors.
\newblock In \emph{Proceedings of the IEEE/CVF Conference on Computer Vision
  and Pattern Recognition}, pages 20637--20647, 2023.

\bibitem[Denton and Fergus(2018)]{svg}
Emily Denton and Rob Fergus.
\newblock Stochastic video generation with a learned prior.
\newblock In \emph{Proceedings of the 35th International Conference on Machine
  Learning, {ICML} 2018, Stockholmsm{\"{a}}ssan, Stockholm, Sweden, July 10-15,
  2018}, 2018.

\bibitem[Dhariwal and Nichol(2021)]{dhariwal2021diffusion}
Prafulla Dhariwal and Alex Nichol.
\newblock {Diffusion Models Beat GANs on Image Synthesis}.
\newblock \emph{arXiv:2105.05233}, 2021.

\bibitem[Dorkenwald et~al.(2021)Dorkenwald, Milbich, Blattmann, Rombach,
  Derpanis, and Ommer]{si2v}
Michael Dorkenwald, Timo Milbich, Andreas Blattmann, Robin Rombach,
  Konstantinos~G. Derpanis, and Bj{\"{o}}rn Ommer.
\newblock Stochastic image-to-video synthesis using cinns.
\newblock In \emph{{IEEE} Conference on Computer Vision and Pattern
  Recognition, {CVPR} 2021, virtual, June 19-25, 2021}, 2021.

\bibitem[Downs et~al.(2022)Downs, Francis, Koenig, Kinman, Hickman, Reymann,
  McHugh, and Vanhoucke]{downs2022google}
Laura Downs, Anthony Francis, Nate Koenig, Brandon Kinman, Ryan Hickman, Krista
  Reymann, Thomas~B McHugh, and Vincent Vanhoucke.
\newblock Google scanned objects: A high-quality dataset of 3d scanned
  household items.
\newblock In \emph{2022 International Conference on Robotics and Automation
  (ICRA)}, pages 2553--2560. IEEE, 2022.

\bibitem[Elo(1978)]{elo1978rating}
Arpad~E. Elo.
\newblock \emph{The Rating of Chessplayers, Past and Present}.
\newblock Arco Pub., New York, 1978.

\bibitem[Esser et~al.(2020)Esser, Rombach, and Ommer]{esser2020taming}
Patrick Esser, Robin Rombach, and Björn Ommer.
\newblock Taming transformers for high-resolution image synthesis.
\newblock \emph{arXiv preprint arXiv:2012.09841}, 2020.

\bibitem[Esser et~al.(2023)Esser, Chiu, Atighehchian, Granskog, and
  Germanidis]{esser2023structure}
Patrick Esser, Johnathan Chiu, Parmida Atighehchian, Jonathan Granskog, and
  Anastasis Germanidis.
\newblock Structure and content-guided video synthesis with diffusion models,
  2023.

\bibitem[Farnebäck(2003)]{farneback2003flow}
Gunnar Farnebäck.
\newblock Two-frame motion estimation based on polynomial expansion.
\newblock pages 363--370, 2003.

\bibitem[Fox et~al.(2021)Fox, Tewari, Elgharib, and
  Theobalt]{fox2021stylevideogan}
Gereon Fox, Ayush Tewari, Mohamed Elgharib, and Christian Theobalt.
\newblock Stylevideogan: A temporal generative model using a pretrained
  stylegan.
\newblock In \emph{British Machine Vision Conference (BMVC)}, 2021.

\bibitem[Franceschi et~al.(2020)Franceschi, Delasalles, Chen, Lamprier, and
  Gallinari]{lsvg}
Jean-Yves Franceschi, Edouard Delasalles, Micka\"{e}l Chen, Sylvain Lamprier,
  and Patrick Gallinari.
\newblock Stochastic latent residual video prediction.
\newblock In \emph{Proceedings of the 37th International Conference on Machine
  Learning}, 2020.

\bibitem[Gao et~al.(2020)Gao, Biderman, Black, Golding, Hoppe, Foster, Phang,
  He, Thite, Nabeshima, Presser, and Leahy]{pile}
Leo Gao, Stella Biderman, Sid Black, Laurence Golding, Travis Hoppe, Charles
  Foster, Jason Phang, Horace He, Anish Thite, Noa Nabeshima, Shawn Presser,
  and Connor Leahy.
\newblock The {P}ile: An 800gb dataset of diverse text for language modeling.
\newblock \emph{arXiv preprint arXiv:2101.00027}, 2020.

\bibitem[Ge et~al.(2022)Ge, Hayes, Yang, Yin, Pang, Jacobs, Huang, and
  Parikh]{ge2022longvideo}
Songwei Ge, Thomas Hayes, Harry Yang, Xi Yin, Guan Pang, David Jacobs, Jia-Bin
  Huang, and Devi Parikh.
\newblock Long video generation with time-agnostic vqgan and time-sensitive
  transformer.
\newblock In \emph{Computer Vision -- ECCV 2022}, pages 102--118, Cham, 2022.
  Springer Nature Switzerland.

\bibitem[Ge et~al.(2023)Ge, Nah, Liu, Poon, Tao, Catanzaro, Jacobs, Huang, Liu,
  and Balaji]{ge2023preserve}
Songwei Ge, Seungjun Nah, Guilin Liu, Tyler Poon, Andrew Tao, Bryan Catanzaro,
  David Jacobs, Jia-Bin Huang, Ming-Yu Liu, and Yogesh Balaji.
\newblock Preserve your own correlation: A noise prior for video diffusion
  models.
\newblock In \emph{Proceedings of the IEEE/CVF International Conference on
  Computer Vision}, pages 22930--22941, 2023.

\bibitem[Goodfellow et~al.(2014)Goodfellow, Pouget-Abadie, Mirza, Xu,
  Warde-Farley, Ozair, Courville, and Bengio]{goodfellow2014generative}
Ian Goodfellow, Jean Pouget-Abadie, Mehdi Mirza, Bing Xu, David Warde-Farley,
  Sherjil Ozair, Aaron Courville, and Yoshua Bengio.
\newblock Generative adversarial nets.
\newblock \emph{Advances in neural information processing systems}, 27, 2014.

\bibitem[Gu et~al.(2023)Gu, Wang, Zhao, Lu, Zhang, Wu, Xu, Zhang, Jiang, and
  Xu]{gu2023reuse}
Jiaxi Gu, Shicong Wang, Haoyu Zhao, Tianyi Lu, Xing Zhang, Zuxuan Wu, Songcen
  Xu, Wei Zhang, Yu-Gang Jiang, and Hang Xu.
\newblock Reuse and diffuse: Iterative denoising for text-to-video generation.
\newblock \emph{arXiv preprint arXiv:2309.03549}, 2023.

\bibitem[Guo et~al.(2023)Guo, Yang, Rao, Wang, Qiao, Lin, and
  Dai]{guo2023animatediff}
Yuwei Guo, Ceyuan Yang, Anyi Rao, Yaohui Wang, Yu Qiao, Dahua Lin, and Bo Dai.
\newblock Animatediff: Animate your personalized text-to-image diffusion models
  without specific tuning.
\newblock \emph{arXiv preprint arXiv:2307.04725}, 2023.

\bibitem[Gupta et~al.(2022)Gupta, Keshari, and Das]{Gupta_2022_CVPR}
Sonam Gupta, Arti Keshari, and Sukhendu Das.
\newblock Rv-gan: Recurrent gan for unconditional video generation.
\newblock In \emph{Proceedings of the IEEE/CVF Conference on Computer Vision
  and Pattern Recognition (CVPR) Workshops}, pages 2024--2033, 2022.

\bibitem[Guttenberg and CrossLabs(2023)]{guttenberg2023diffusion}
Nicholas Guttenberg and CrossLabs.
\newblock Diffusion with offset noise, 2023.

\bibitem[He et~al.(2023)He, Yang, Zhang, Shan, and Chen]{he2023latent}
Yingqing He, Tianyu Yang, Yong Zhang, Ying Shan, and Qifeng Chen.
\newblock Latent video diffusion models for high-fidelity long video
  generation, 2023.

\bibitem[Ho and Salimans(2021)]{ho2021classifierfree}
Jonathan Ho and Tim Salimans.
\newblock Classifier-free diffusion guidance.
\newblock In \emph{NeurIPS 2021 Workshop on Deep Generative Models and
  Downstream Applications}, 2021.

\bibitem[Ho and Salimans(2022)]{ho2022classifier}
Jonathan Ho and Tim Salimans.
\newblock {Classifier-Free Diffusion Guidance}.
\newblock \emph{arXiv:2207.12598}, 2022.

\bibitem[Ho et~al.(2020)Ho, Jain, and Abbeel]{ho2020ddpm}
Jonathan Ho, Ajay Jain, and Pieter Abbeel.
\newblock Denoising diffusion probabilistic models.
\newblock In \emph{Advances in Neural Information Processing Systems}, 2020.

\bibitem[Ho et~al.(2021)Ho, Saharia, Chan, Fleet, Norouzi, and
  Salimans]{ho2021cascaded}
Jonathan Ho, Chitwan Saharia, William Chan, David~J Fleet, Mohammad Norouzi,
  and Tim Salimans.
\newblock Cascaded diffusion models for high fidelity image generation.
\newblock \emph{arXiv preprint arXiv:2106.15282}, 2021.

\bibitem[Ho et~al.(2022{\natexlab{a}})Ho, Chan, Saharia, Whang, Gao, Gritsenko,
  Kingma, Poole, Norouzi, Fleet, and Salimans]{ho2022imagen}
Jonathan Ho, William Chan, Chitwan Saharia, Jay Whang, Ruiqi Gao, Alexey
  Gritsenko, Diederik~P Kingma, Ben Poole, Mohammad Norouzi, David~J Fleet, and
  Tim Salimans.
\newblock {Imagen Video: High Definition Video Generation with Diffusion
  Models}.
\newblock \emph{arXiv:2210.02303}, 2022{\natexlab{a}}.

\bibitem[Ho et~al.(2022{\natexlab{b}})Ho, Chan, Saharia, Whang, Gao, Gritsenko,
  Kingma, Poole, Norouzi, Fleet, and Salimans]{ho2022imagenvideo}
Jonathan Ho, William Chan, Chitwan Saharia, Jay Whang, Ruiqi Gao, Alexey
  Gritsenko, Diederik~P. Kingma, Ben Poole, Mohammad Norouzi, David~J. Fleet,
  and Tim Salimans.
\newblock Imagen video: High definition video generation with diffusion models.
\newblock \emph{arXiv preprint arXiv:2210.02303}, 2022{\natexlab{b}}.

\bibitem[Ho et~al.(2022{\natexlab{c}})Ho, Salimans, Gritsenko, Chan, Norouzi,
  and Fleet]{ho2022video}
Jonathan Ho, Tim Salimans, Alexey Gritsenko, William Chan, Mohammad Norouzi,
  and David~J. Fleet.
\newblock Video diffusion models.
\newblock \emph{arXiv preprint arXiv:2204.03458}, 2022{\natexlab{c}}.

\bibitem[Hong et~al.(2022)Hong, Ding, Zheng, Liu, and Tang]{hong2022cogvideo}
Wenyi Hong, Ming Ding, Wendi Zheng, Xinghan Liu, and Jie Tang.
\newblock Cogvideo: Large-scale pretraining for text-to-video generation via
  transformers, 2022.

\bibitem[Hoogeboom et~al.(2023)Hoogeboom, Heek, and
  Salimans]{hoogeboom2023simple}
Emiel Hoogeboom, Jonathan Heek, and Tim Salimans.
\newblock {simple diffusion: End-to-end diffusion for high resolution images}.
\newblock \emph{arXiv preprint arXiv:2301.11093}, 2023.

\bibitem[Hu et~al.(2021)Hu, Shen, Wallis, Allen-Zhu, Li, Wang, Wang, and
  Chen]{hu2021lora}
Edward~J Hu, Yelong Shen, Phillip Wallis, Zeyuan Allen-Zhu, Yuanzhi Li, Shean
  Wang, Lu Wang, and Weizhu Chen.
\newblock Lora: Low-rank adaptation of large language models.
\newblock \emph{arXiv preprint arXiv:2106.09685}, 2021.

\bibitem[Hyv{\"a}rinen and Dayan(2005)]{hyvarinen2005estimation}
Aapo Hyv{\"a}rinen and Peter Dayan.
\newblock {Estimation of Non-Normalized Statistical Models by Score Matching}.
\newblock \emph{Journal of Machine Learning Research}, 6\penalty0 (4), 2005.

\bibitem[Ilharco et~al.(2021)Ilharco, Wortsman, Wightman, Gordon, Carlini,
  Taori, Dave, Shankar, Namkoong, Miller, Hajishirzi, Farhadi, and
  Schmidt]{ilharco2021open}
Gabriel Ilharco, Mitchell Wortsman, Ross Wightman, Cade Gordon, Nicholas
  Carlini, Rohan Taori, Achal Dave, Vaishaal Shankar, Hongseok Namkoong, John
  Miller, Hannaneh Hajishirzi, Ali Farhadi, and Ludwig Schmidt.
\newblock Openclip, 2021.

\bibitem[Itseez(2015)]{itseez2015opencv}
Itseez.
\newblock Open source computer vision library.
\newblock \url{https://github.com/itseez/opencv}, 2015.

\bibitem[Jun and Nichol(2023)]{jun2023shape}
Heewoo Jun and Alex Nichol.
\newblock Shap-e: Generating conditional 3d implicit functions, 2023.

\bibitem[Kahembwe and Ramamoorthy(2020)]{kahembwe2020lower}
Emmanuel Kahembwe and Subramanian Ramamoorthy.
\newblock Lower dimensional kernels for video discriminators.
\newblock \emph{Neural Networks}, 132:\penalty0 506--520, 2020.

\bibitem[Karras et~al.(2022)Karras, Aittala, Aila, and
  Laine]{karras2022elucidating}
Tero Karras, Miika Aittala, Timo Aila, and Samuli Laine.
\newblock {Elucidating the Design Space of Diffusion-Based Generative Models}.
\newblock \emph{arXiv:2206.00364}, 2022.

\bibitem[Khachatryan et~al.(2023)Khachatryan, Movsisyan, Tadevosyan, Henschel,
  Wang, Navasardyan, and Shi]{khachatryan2023text2videozero}
Levon Khachatryan, Andranik Movsisyan, Vahram Tadevosyan, Roberto Henschel,
  Zhangyang Wang, Shant Navasardyan, and Humphrey Shi.
\newblock Text2video-zero: Text-to-image diffusion models are zero-shot video
  generators, 2023.

\bibitem[Kingma et~al.(2021)Kingma, Salimans, Poole, and
  Ho]{kingma2021variational}
Diederik Kingma, Tim Salimans, Ben Poole, and Jonathan Ho.
\newblock Variational diffusion models.
\newblock \emph{Advances in neural information processing systems},
  34:\penalty0 21696--21707, 2021.

\bibitem[Labs(2023)]{pika}
Pika Labs.
\newblock Pika labs, \url{https://www.pika.art/}, 2023.

\bibitem[Lee et~al.(2018)Lee, Zhang, Ebert, Abbeel, Finn, and
  Levine]{lee2018savp}
Alex~X. Lee, Richard Zhang, Frederik Ebert, Pieter Abbeel, Chelsea Finn, and
  Sergey Levine.
\newblock Stochastic adversarial video prediction.
\newblock \emph{arXiv preprint arXiv:1804.01523}, 2018.

\bibitem[Lin et~al.(2023)Lin, Liu, Li, and Yang]{lin2023common}
Shanchuan Lin, Bingchen Liu, Jiashi Li, and Xiao Yang.
\newblock {Common Diffusion Noise Schedules and Sample Steps are Flawed}.
\newblock \emph{arXiv:2305.08891}, 2023.

\bibitem[Liu et~al.(2023{\natexlab{a}})Liu, Wu, Hoorick, Tokmakov, Zakharov,
  and Vondrick]{liu2023zero1to3}
Ruoshi Liu, Rundi Wu, Basile~Van Hoorick, Pavel Tokmakov, Sergey Zakharov, and
  Carl Vondrick.
\newblock Zero-1-to-3: Zero-shot one image to 3d object, 2023{\natexlab{a}}.

\bibitem[Liu et~al.(2023{\natexlab{b}})Liu, Lin, Zeng, Long, Liu, Komura, and
  Wang]{liu2023syncdreamer}
Yuan Liu, Cheng Lin, Zijiao Zeng, Xiaoxiao Long, Lingjie Liu, Taku Komura, and
  Wenping Wang.
\newblock Syncdreamer: Generating multiview-consistent images from a
  single-view image.
\newblock \emph{arXiv preprint arXiv:2309.03453}, 2023{\natexlab{b}}.

\bibitem[Loshchilov and Hutter(2017)]{loshchilov2017decoupled}
Ilya Loshchilov and Frank Hutter.
\newblock Decoupled weight decay regularization.
\newblock \emph{arXiv preprint arXiv:1711.05101}, 2017.

\bibitem[Luc et~al.(2020)Luc, Clark, Dieleman, de~Las~Casas, Doron, Cassirer,
  and Simonyan]{Luc2020TransformationbasedAV}
Pauline Luc, Aidan Clark, Sander Dieleman, Diego de Las~Casas, Yotam Doron,
  Albin Cassirer, and Karen Simonyan.
\newblock Transformation-based adversarial video prediction on large-scale
  data.
\newblock \emph{ArXiv}, 2020.

\bibitem[Meng et~al.(2023)Meng, Rombach, Gao, Kingma, Ermon, Ho, and
  Salimans]{meng2023distillation}
Chenlin Meng, Robin Rombach, Ruiqi Gao, Diederik~P. Kingma, Stefano Ermon,
  Jonathan Ho, and Tim Salimans.
\newblock On distillation of guided diffusion models, 2023.

\bibitem[Nichol et~al.(2022)Nichol, Jun, Dhariwal, Mishkin, and
  Chen]{nichol2022pointe}
Alex Nichol, Heewoo Jun, Prafulla Dhariwal, Pamela Mishkin, and Mark Chen.
\newblock Point-e: A system for generating 3d point clouds from complex
  prompts, 2022.

\bibitem[Penedo et~al.(2023)Penedo, Malartic, Hesslow, Cojocaru, Cappelli,
  Alobeidli, Pannier, Almazrouei, and Launay]{refinedweb}
Guilherme Penedo, Quentin Malartic, Daniel Hesslow, Ruxandra Cojocaru,
  Alessandro Cappelli, Hamza Alobeidli, Baptiste Pannier, Ebtesam Almazrouei,
  and Julien Launay.
\newblock The {R}efined{W}eb dataset for {F}alcon {LLM}: outperforming curated
  corpora with web data, and web data only.
\newblock \emph{arXiv preprint arXiv:2306.01116}, 2023.

\bibitem[Podell et~al.(2023)Podell, English, Lacey, Blattmann, Dockhorn,
  M{\"u}ller, Penna, and Rombach]{podell2023sdxl}
Dustin Podell, Zion English, Kyle Lacey, Andreas Blattmann, Tim Dockhorn, Jonas
  M{\"u}ller, Joe Penna, and Robin Rombach.
\newblock {SDXL: Improving Latent Diffusion Models for High-Resolution Image
  Synthesis}.
\newblock \emph{arXiv:2307.01952}, 2023.

\bibitem[Puccetti et~al.(2023)Puccetti, Kilian, and
  Beaumont]{pucetti2023training}
Giovanni Puccetti, Maciej Kilian, and Romain Beaumont.
\newblock Training contrastive captioners.
\newblock \emph{LAION blog}, 2023.

\bibitem[Radford et~al.(2021)Radford, Kim, Hallacy, Ramesh, Goh, Agarwal,
  Sastry, Askell, Mishkin, Clark, Krueger, and Sutskever]{radford2021learning}
Alec Radford, Jong~Wook Kim, Chris Hallacy, Aditya Ramesh, Gabriel Goh,
  Sandhini Agarwal, Girish Sastry, Amanda Askell, Pamela Mishkin, Jack Clark,
  Gretchen Krueger, and Ilya Sutskever.
\newblock {Learning Transferable Visual Models From Natural Language
  Supervision}.
\newblock \emph{arXiv:2103.00020}, 2021.

\bibitem[Raffel et~al.(2019)Raffel, Shazeer, Roberts, Lee, Narang, Matena,
  Zhou, Li, and Liu]{2019t5}
Colin Raffel, Noam Shazeer, Adam Roberts, Katherine Lee, Sharan Narang, Michael
  Matena, Yanqi Zhou, Wei Li, and Peter~J. Liu.
\newblock Exploring the limits of transfer learning with a unified text-to-text
  transformer.
\newblock \emph{arXiv e-prints}, 2019.

\bibitem[Ramesh(2022)]{rameshdalle2}
Aditya Ramesh.
\newblock How dall·e 2 works, 2022.

\bibitem[Ramesh et~al.(2022{\natexlab{a}})Ramesh, Dhariwal, Nichol, Chu, and
  Chen]{ramesh2022dalle2}
Aditya Ramesh, Prafulla Dhariwal, Alex Nichol, Casey Chu, and Mark Chen.
\newblock Hierarchical text-conditional image generation with clip latents.
\newblock \emph{arXiv preprint arXiv:2204.06125}, 2022{\natexlab{a}}.

\bibitem[Ramesh et~al.(2022{\natexlab{b}})Ramesh, Dhariwal, Nichol, Chu, and
  Chen]{ramesh2022hierarchical}
Aditya Ramesh, Prafulla Dhariwal, Alex Nichol, Casey Chu, and Mark Chen.
\newblock {Hierarchical Text-Conditional Image Generation with CLIP Latents}.
\newblock \emph{arXiv:2204.06125}, 2022{\natexlab{b}}.

\bibitem[Rombach et~al.(2021{\natexlab{a}})Rombach, Blattmann, Lorenz, Esser,
  and Ommer]{rombach2021high}
Robin Rombach, Andreas Blattmann, Dominik Lorenz, Patrick Esser, and Bj{\"o}rn
  Ommer.
\newblock {High-Resolution Image Synthesis with Latent Diffusion Models}.
\newblock \emph{arXiv preprint arXiv:2112.10752}, 2021{\natexlab{a}}.

\bibitem[Rombach et~al.(2021{\natexlab{b}})Rombach, Blattmann, Lorenz, Esser,
  and Ommer]{rombach2021highresolution}
Robin Rombach, Andreas Blattmann, Dominik Lorenz, Patrick Esser, and Bj{\"o}rn
  Ommer.
\newblock High-resolution image synthesis with latent diffusion models.
\newblock \emph{arXiv preprint arXiv:2112.10752}, 2021{\natexlab{b}}.

\bibitem[Ronneberger et~al.(2015)Ronneberger, Fischer, and
  Brox]{ronneberger2015u}
Olaf Ronneberger, Philipp Fischer, and Thomas Brox.
\newblock {U-Net: Convolutional Networks for Biomedical Image Segmentation}.
\newblock \emph{arXiv:1505.04597}, 2015.

\bibitem[RunwayML(2023)]{gen2}
RunwayML.
\newblock Gen-2 by runway, \url{https://research.runwayml.com/gen2}, 2023.

\bibitem[Saharia et~al.(2021)Saharia, Ho, Chan, Salimans, Fleet, and
  Norouzi]{saharia2021image}
Chitwan Saharia, Jonathan Ho, William Chan, Tim Salimans, David~J Fleet, and
  Mohammad Norouzi.
\newblock Image super-resolution via iterative refinement.
\newblock \emph{arXiv preprint arXiv:2104.07636}, 2021.

\bibitem[Saharia et~al.(2022)Saharia, Chan, Saxena, Li, Whang, Denton,
  Ghasemipour, Ayan, Mahdavi, Lopes, Salimans, Ho, Fleet, and
  Norouzi]{saharia2022imagen}
Chitwan Saharia, William Chan, Saurabh Saxena, Lala Li, Jay Whang, Emily
  Denton, Seyed Kamyar~Seyed Ghasemipour, Burcu~Karagol Ayan, S.~Sara Mahdavi,
  Rapha~Gontijo Lopes, Tim Salimans, Jonathan Ho, David~J Fleet, and Mohammad
  Norouzi.
\newblock Photorealistic text-to-image diffusion models with deep language
  understanding.
\newblock \emph{arXiv preprint arXiv:2205.11487}, 2022.

\bibitem[Saito et~al.(2017)Saito, Matsumoto, and Saito]{TGAN2017}
Masaki Saito, Eiichi Matsumoto, and Shunta Saito.
\newblock Temporal generative adversarial nets with singular value clipping.
\newblock In \emph{ICCV}, 2017.

\bibitem[Saito et~al.(2020)Saito, Saito, Koyama, and Kobayashi]{TGAN2020}
Masaki Saito, Shunta Saito, Masanori Koyama, and Sosuke Kobayashi.
\newblock Train sparsely, generate densely: Memory-efficient unsupervised
  training of high-resolution temporal gan.
\newblock \emph{International Journal of Computer Vision}, 2020.

\bibitem[Salimans and Ho(2022)]{salimans2022progressive}
Tim Salimans and Jonathan Ho.
\newblock {Progressive Distillation for Fast Sampling of Diffusion Models}.
\newblock \emph{arXiv preprint arXiv:2202.00512}, 2022.

\bibitem[Schuhmann et~al.(2022)Schuhmann, Beaumont, Vencu, Gordon, Wightman,
  Cherti, Coombes, Katta, Mullis, Wortsman, et~al.]{schuhmann2022laion}
Christoph Schuhmann, Romain Beaumont, Richard Vencu, Cade Gordon, Ross
  Wightman, Mehdi Cherti, Theo Coombes, Aarush Katta, Clayton Mullis, Mitchell
  Wortsman, et~al.
\newblock Laion-5b: An open large-scale dataset for training next generation
  image-text models.
\newblock \emph{Advances in Neural Information Processing Systems},
  35:\penalty0 25278--25294, 2022.

\bibitem[Shi et~al.(2023)Shi, Wang, Ye, Long, Li, and Yang]{shi2023mvdream}
Yichun Shi, Peng Wang, Jianglong Ye, Mai Long, Kejie Li, and Xiao Yang.
\newblock Mvdream: Multi-view diffusion for 3d generation.
\newblock \emph{arXiv preprint arXiv:2308.16512}, 2023.

\bibitem[Singer et~al.(2022)Singer, Polyak, Hayes, Yin, An, Zhang, Hu, Yang,
  Ashual, Gafni, Parikh, Gupta, and Taigman]{singer2022make}
Uriel Singer, Adam Polyak, Thomas Hayes, Xi Yin, Jie An, Songyang Zhang, Qiyuan
  Hu, Harry Yang, Oron Ashual, Oran Gafni, Devi Parikh, Sonal Gupta, and Yaniv
  Taigman.
\newblock {Make-A-Video: Text-to-Video Generation without Text-Video Data}.
\newblock \emph{arXiv:2209.14792}, 2022.

\bibitem[Skorokhodov et~al.(2022)Skorokhodov, Tulyakov, and
  Elhoseiny]{Skorokhodov_2022_CVPR}
Ivan Skorokhodov, Sergey Tulyakov, and Mohamed Elhoseiny.
\newblock Stylegan-v: A continuous video generator with the price, image
  quality and perks of stylegan2.
\newblock In \emph{Proceedings of the IEEE/CVF Conference on Computer Vision
  and Pattern Recognition (CVPR)}, pages 3626--3636, 2022.

\bibitem[Sohl-Dickstein et~al.(2015)Sohl-Dickstein, Weiss, Maheswaranathan, and
  Ganguli]{sohl2015deep}
Jascha Sohl-Dickstein, Eric~A Weiss, Niru Maheswaranathan, and Surya Ganguli.
\newblock {Deep Unsupervised Learning using Nonequilibrium Thermodynamics}.
\newblock \emph{arXiv:1503.03585}, 2015.

\bibitem[Somepalli et~al.(2023)Somepalli, Singla, Goldblum, Geiping, and
  Goldstein]{somepalli2023understanding}
Gowthami Somepalli, Vasu Singla, Micah Goldblum, Jonas Geiping, and Tom
  Goldstein.
\newblock Understanding and mitigating copying in diffusion models, 2023.

\bibitem[Song and Ermon(2020)]{song2020improved}
Yang Song and Stefano Ermon.
\newblock {Improved Techniques for Training Score-Based Generative Models}.
\newblock \emph{arXiv:2006.09011}, 2020.

\bibitem[Song et~al.(2020)Song, Sohl-Dickstein, Kingma, Kumar, Ermon, and
  Poole]{song2020score}
Yang Song, Jascha Sohl-Dickstein, Diederik~P Kingma, Abhishek Kumar, Stefano
  Ermon, and Ben Poole.
\newblock {Score-Based Generative Modeling through Stochastic Differential
  Equations}.
\newblock \emph{arXiv:2011.13456}, 2020.

\bibitem[Soomro et~al.(2012)Soomro, Zamir, and Shah]{soomro2012ucf101}
Khurram Soomro, Amir~Roshan Zamir, and Mubarak Shah.
\newblock Ucf101: A dataset of 101 human actions classes from videos in the
  wild.
\newblock \emph{arXiv preprint arXiv:1212.0402}, 2012.

\bibitem[Teed and Deng(2020)]{teed2020raft}
Zachary Teed and Jia Deng.
\newblock Raft: Recurrent all-pairs field transforms for optical flow.
\newblock In \emph{Computer Vision--ECCV 2020: 16th European Conference,
  Glasgow, UK, August 23--28, 2020, Proceedings, Part II 16}, pages 402--419.
  Springer, 2020.

\bibitem[Tian et~al.(2021)Tian, Ren, Chai, Olszewski, Peng, Metaxas, and
  Tulyakov]{tian2021a}
Yu Tian, Jian Ren, Menglei Chai, Kyle Olszewski, Xi Peng, Dimitris~N. Metaxas,
  and Sergey Tulyakov.
\newblock A good image generator is what you need for high-resolution video
  synthesis.
\newblock In \emph{International Conference on Learning Representations}, 2021.

\bibitem[Tomar(2006)]{tomar2006converting}
Suramya Tomar.
\newblock Converting video formats with ffmpeg.
\newblock \emph{Linux Journal}, 2006\penalty0 (146):\penalty0 10, 2006.

\bibitem[Vahdat et~al.(2021)Vahdat, Kreis, and Kautz]{vahdat2021score}
Arash Vahdat, Karsten Kreis, and Jan Kautz.
\newblock Score-based generative modeling in latent space.
\newblock In \emph{Advances in Neural Information Processing Systems}, 2021.

\bibitem[Villegas et~al.(2017)Villegas, Yang, Hong, Lin, and
  Lee]{villegas17mcnet}
Ruben Villegas, Jimei Yang, Seunghoon Hong, Xunyu Lin, and Honglak Lee.
\newblock Decomposing motion and content for natural video sequence prediction.
\newblock \emph{ICLR}, 2017.

\bibitem[Villegas et~al.(2022)Villegas, Babaeizadeh, Kindermans, Moraldo,
  Zhang, Saffar, Castro, Kunze, and Erhan]{villegas2022phenaki}
Ruben Villegas, Mohammad Babaeizadeh, Pieter-Jan Kindermans, Hernan Moraldo,
  Han Zhang, Mohammad~Taghi Saffar, Santiago Castro, Julius Kunze, and Dumitru
  Erhan.
\newblock Phenaki: Variable length video generation from open domain textual
  description.
\newblock \emph{arXiv:2210.02399}, 2022.

\bibitem[Voleti et~al.(2022)Voleti, Jolicoeur-Martineau, and
  Pal]{voleti2022mcvd}
Vikram Voleti, Alexia Jolicoeur-Martineau, and Christopher Pal.
\newblock Mcvd: Masked conditional video diffusion for prediction, generation,
  and interpolation.
\newblock In \emph{(NeurIPS) Advances in Neural Information Processing
  Systems}, 2022.

\bibitem[Vondrick et~al.(2016)Vondrick, Pirsiavash, and Torralba]{scene_dyn}
Carl Vondrick, Hamed Pirsiavash, and Antonio Torralba.
\newblock Generating videos with scene dynamics.
\newblock In \emph{Proceedings of the 30th International Conference on Neural
  Information Processing Systems}, 2016.

\bibitem[Wang et~al.(2023{\natexlab{a}})Wang, Yuan, Chen, Zhang, Wang, and
  Zhang]{wang2023modelscope}
Jiuniu Wang, Hangjie Yuan, Dayou Chen, Yingya Zhang, Xiang Wang, and Shiwei
  Zhang.
\newblock Modelscope text-to-video technical report.
\newblock \emph{arXiv preprint arXiv:2308.06571}, 2023{\natexlab{a}}.

\bibitem[Wang et~al.(2020)Wang, Bilinski, Bremond, and
  Dantcheva]{Wang_2020_CVPR}
Yaohui Wang, Piotr Bilinski, Francois Bremond, and Antitza Dantcheva.
\newblock G3an: Disentangling appearance and motion for video generation.
\newblock In \emph{IEEE/CVF Conference on Computer Vision and Pattern
  Recognition (CVPR)}, 2020.

\bibitem[Wang et~al.(2023{\natexlab{b}})Wang, Chen, Ma, Zhou, Huang, Wang,
  Yang, He, Yu, Yang, et~al.]{wang2023lavie}
Yaohui Wang, Xinyuan Chen, Xin Ma, Shangchen Zhou, Ziqi Huang, Yi Wang, Ceyuan
  Yang, Yinan He, Jiashuo Yu, Peiqing Yang, et~al.
\newblock Lavie: High-quality video generation with cascaded latent diffusion
  models.
\newblock \emph{arXiv preprint arXiv:2309.15103}, 2023{\natexlab{b}}.

\bibitem[Wang et~al.(2023{\natexlab{c}})Wang, He, Li, Li, Yu, Ma, Chen, Wang,
  Luo, Liu, Wang, Wang, and Qiao]{wang2023internvid}
Yi Wang, Yinan He, Yizhuo Li, Kunchang Li, Jiashuo Yu, Xin Ma, Xinyuan Chen,
  Yaohui Wang, Ping Luo, Ziwei Liu, Yali Wang, Limin Wang, and Yu Qiao.
\newblock Internvid: A large-scale video-text dataset for multimodal
  understanding and generation, 2023{\natexlab{c}}.

\bibitem[Watson et~al.(2022)Watson, Chan, Martin-Brualla, Ho, Tagliasacchi, and
  Norouzi]{watson2022novel}
Daniel Watson, William Chan, Ricardo Martin-Brualla, Jonathan Ho, Andrea
  Tagliasacchi, and Mohammad Norouzi.
\newblock Novel view synthesis with diffusion models, 2022.

\bibitem[Weissenborn et~al.(2020)Weissenborn, Täckström, and
  Uszkoreit]{Weissenborn2020Scaling}
Dirk Weissenborn, Oscar Täckström, and Jakob Uszkoreit.
\newblock Scaling autoregressive video models.
\newblock In \emph{International Conference on Learning Representations}, 2020.

\bibitem[Wu et~al.(2021)Wu, Huang, Zhang, Li, Ji, Yang, Sapiro, and
  Duan]{wu2021godiva}
Chenfei Wu, Lun Huang, Qianxi Zhang, Binyang Li, Lei Ji, Fan Yang, Guillermo
  Sapiro, and Nan Duan.
\newblock Godiva: Generating open-domain videos from natural descriptions.
\newblock \emph{arXiv:2104.14806}, 2021.

\bibitem[Wu et~al.(2022)Wu, Liang, Ji, Yang, Fang, Jiang, and Duan]{wu2022nuwa}
Chenfei Wu, Jian Liang, Lei Ji, Fan Yang, Yuejian Fang, Daxin Jiang, and Nan
  Duan.
\newblock N{\"u}wa: Visual synthesis pre-training for neural visual world
  creation.
\newblock In \emph{European Conference on Computer Vision}, pages 720--736.
  Springer, 2022.

\bibitem[Xu et~al.(2023)Xu, Xie, Tan, Huang, Howes, Sharma, Li, Ghosh,
  Zettlemoyer, and Feichtenhofer]{xu2023demystifying}
Hu Xu, Saining Xie, Xiaoqing~Ellen Tan, Po-Yao Huang, Russell Howes, Vasu
  Sharma, Shang-Wen Li, Gargi Ghosh, Luke Zettlemoyer, and Christoph
  Feichtenhofer.
\newblock Demystifying clip data, 2023.

\bibitem[Xu et~al.(2016)Xu, Mei, Yao, and Rui]{xu2016msr-vtt}
Jun Xu, Tao Mei, Ting Yao, and Yong Rui.
\newblock Msr-vtt: A large video description dataset for bridging video and
  language.
\newblock In \emph{International Conference on Computer Vision and Pattern
  Recognition (CVPR)}, 2016.

\bibitem[Yan et~al.(2021)Yan, Zhang, Abbeel, and Srinivas]{yan2021videogpt}
Wilson Yan, Yunzhi Zhang, Pieter Abbeel, and Aravind Srinivas.
\newblock Videogpt: Video generation using vq-vae and transformers, 2021.

\bibitem[Yu et~al.(2022{\natexlab{a}})Yu, Wang, Vasudevan, Yeung,
  Seyedhosseini, and Wu]{yu2022coca}
Jiahui Yu, Zirui Wang, Vijay Vasudevan, Legg Yeung, Mojtaba Seyedhosseini, and
  Yonghui Wu.
\newblock Coca: Contrastive captioners are image-text foundation models,
  2022{\natexlab{a}}.

\bibitem[Yu(2023)]{videoblip}
Keunwoo~Peter Yu.
\newblock Videoblip.
\newblock \url{https://github.com/yukw777/VideoBLIP}, 2023.
\newblock If you use VideoBLIP, please cite it as below.

\bibitem[Yu et~al.(2022{\natexlab{b}})Yu, Tack, Mo, Kim, Kim, Ha, and
  Shin]{yu2022generating}
Sihyun Yu, Jihoon Tack, Sangwoo Mo, Hyunsu Kim, Junho Kim, Jung-Woo Ha, and
  Jinwoo Shin.
\newblock Generating videos with dynamics-aware implicit generative adversarial
  networks.
\newblock In \emph{International Conference on Learning Representations},
  2022{\natexlab{b}}.

\bibitem[Yu et~al.(2023)Yu, Xu, Zhang, Liu, Ye, Wu, Yan, Zhu, Xiong, Liang,
  et~al.]{yu2023mvimgnet}
Xianggang Yu, Mutian Xu, Yidan Zhang, Haolin Liu, Chongjie Ye, Yushuang Wu,
  Zizheng Yan, Chenming Zhu, Zhangyang Xiong, Tianyou Liang, et~al.
\newblock Mvimgnet: A large-scale dataset of multi-view images.
\newblock In \emph{Proceedings of the IEEE/CVF Conference on Computer Vision
  and Pattern Recognition}, pages 9150--9161, 2023.

\bibitem[Zhang et~al.(2018)Zhang, Isola, Efros, Shechtman, and
  Wang]{zhang2018unreasonable}
Richard Zhang, Phillip Isola, Alexei~A. Efros, Eli Shechtman, and Oliver Wang.
\newblock The unreasonable effectiveness of deep features as a perceptual
  metric, 2018.

\bibitem[Zhang et~al.(2023{\natexlab{a}})Zhang, Wang, Zhang, Zhao, Yuan, Qin,
  Wang, Zhao, and Zhou]{zhang2023i2vgen}
Shiwei Zhang, Jiayu Wang, Yingya Zhang, Kang Zhao, Hangjie Yuan, Zhiwu Qin,
  Xiang Wang, Deli Zhao, and Jingren Zhou.
\newblock I2vgen-xl: High-quality image-to-video synthesis via cascaded
  diffusion models.
\newblock \emph{arXiv preprint arXiv:2311.04145}, 2023{\natexlab{a}}.

\bibitem[Zhang et~al.(2023{\natexlab{b}})Zhang, Wei, Jiang, Zhang, Zuo, and
  Tian]{zhang2023controlvideo}
Yabo Zhang, Yuxiang Wei, Dongsheng Jiang, Xiaopeng Zhang, Wangmeng Zuo, and Qi
  Tian.
\newblock Controlvideo: Training-free controllable text-to-video generation,
  2023{\natexlab{b}}.

\bibitem[Zhou et~al.(2022)Zhou, Wang, Yan, Lv, Zhu, and
  Feng]{zhou2022magicvideo}
Daquan Zhou, Weimin Wang, Hanshu Yan, Weiwei Lv, Yizhe Zhu, and Jiashi Feng.
\newblock Magicvideo: Efficient video generation with latent diffusion models.
\newblock \emph{arXiv preprint arXiv:2211.11018}, 2022.

\bibitem[Zhou and Tulsiani(2023)]{zhou2023sparsefusion}
Zhizhuo Zhou and Shubham Tulsiani.
\newblock Sparsefusion: Distilling view-conditioned diffusion for 3d
  reconstruction.
\newblock In \emph{Proceedings of the IEEE/CVF Conference on Computer Vision
  and Pattern Recognition}, pages 12588--12597, 2023.

\end{thebibliography}
}

\newpage
\onecolumn
\tableofcontents
\appendix
\section*{Appendix}

\newcommand{\staticvis}{
\begin{figure}[htbp]
\begin{center}
\renewcommand{\arraystretch}{1.5}  %
\begin{adjustbox}{max width=.98\textwidth}
\begin{tabular}{cc}
\midrule
 \multirow{2}{*}{\textbf{Source Video}\textsuperscript{\phantom{*}}} & \textbf{Optical Flow}  \\[-.5em]
&\textbf{Score}\\
\toprule 
 \raisebox{-0.5\height}{\includegraphics[width=\textwidth]{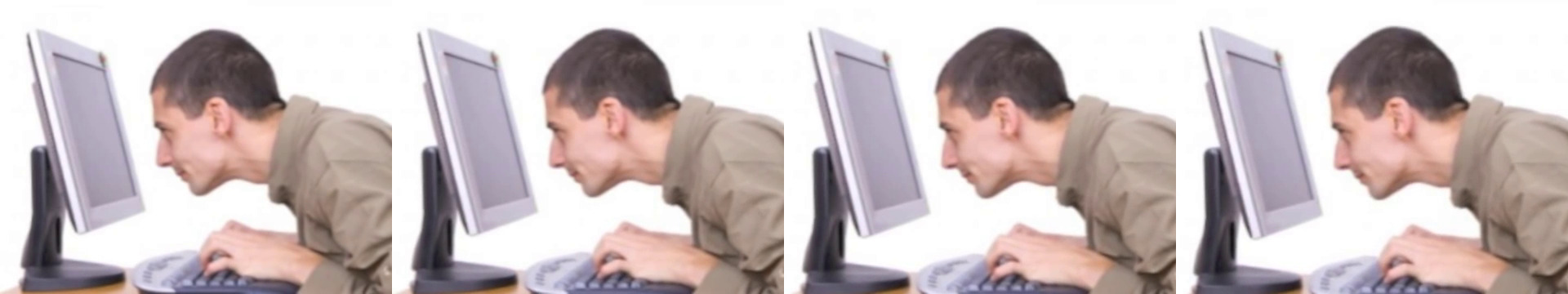}} & \large 0.043 \\
\bottomrule
\end{tabular}
\end{adjustbox}
\end{center}
\caption{\label{fig:static_scenes}
Examples for a static video. Since such static scenes can have a negative impact on generative video-text (pre-)training, we filter them out. 
}
\end{figure}
}

\newcommand{\smallscaleablations}{
\renewcommand{\imwidth}{0.19\textwidth}
\begin{figure*}
    \centering
    \includegraphics[width=\imwidth]{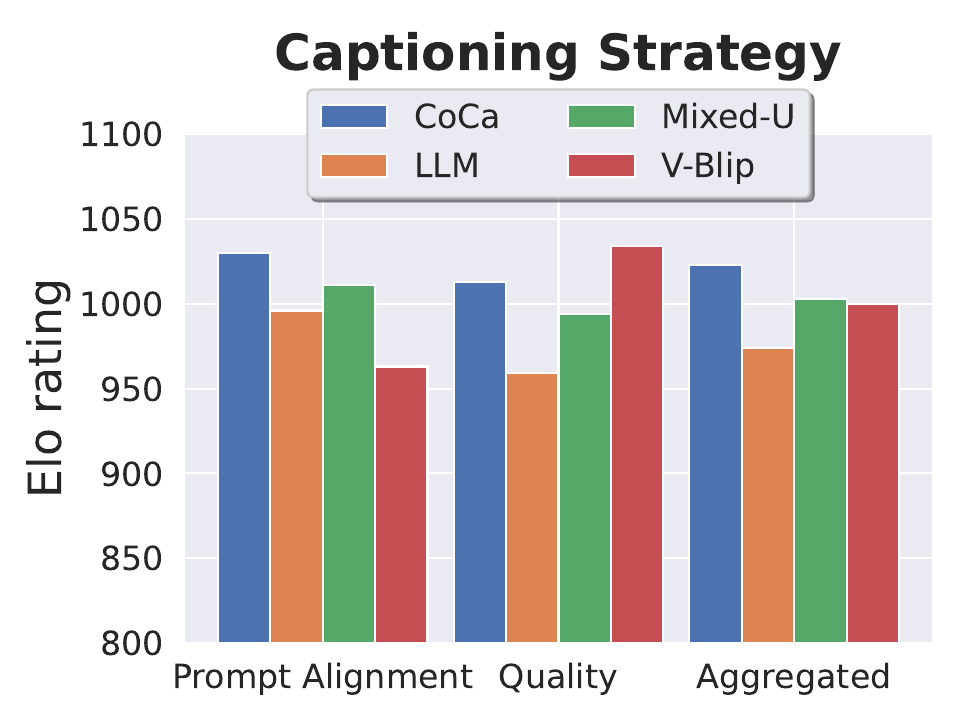}
    \hfill
    \includegraphics[width=\imwidth]{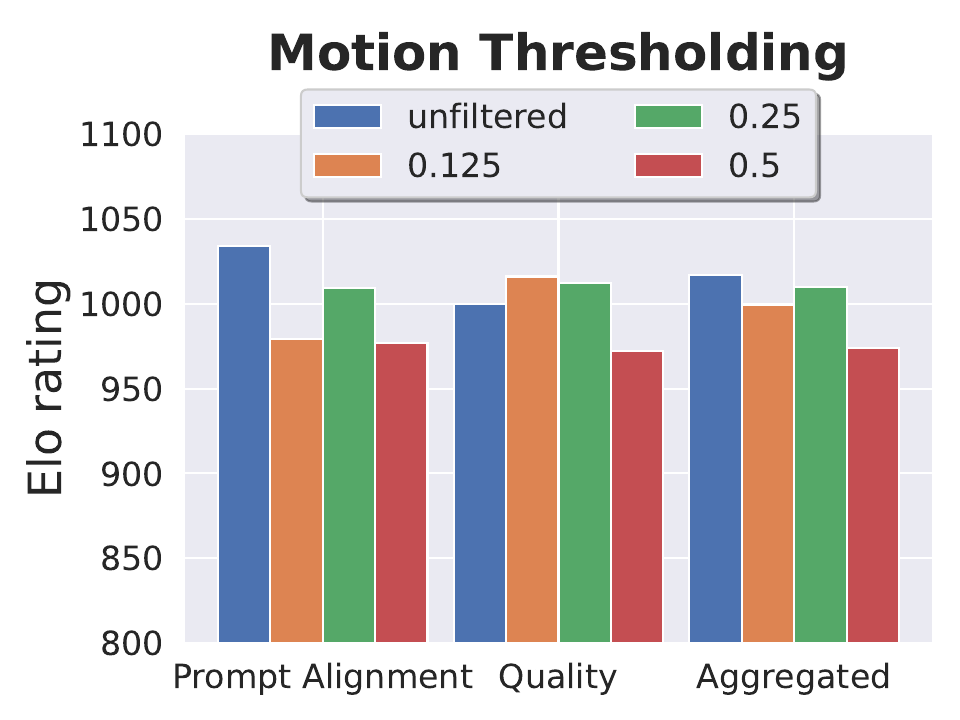}
    \hfill
    \includegraphics[width=\imwidth]{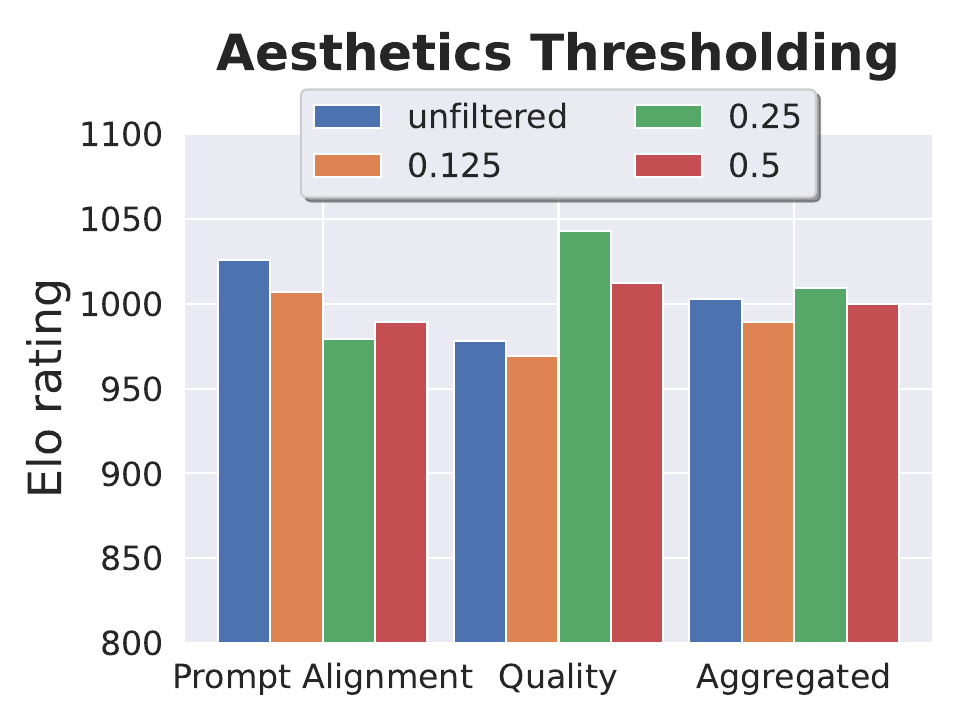}
    \hfill
    \includegraphics[width=\imwidth]{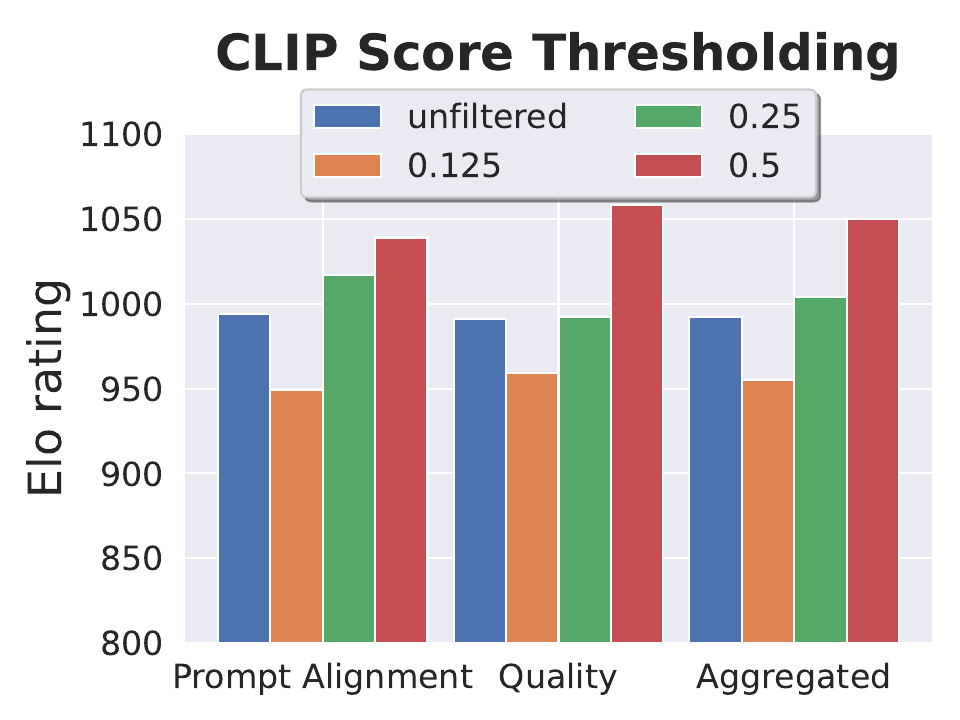}
        \hfill
    \includegraphics[width=\imwidth]{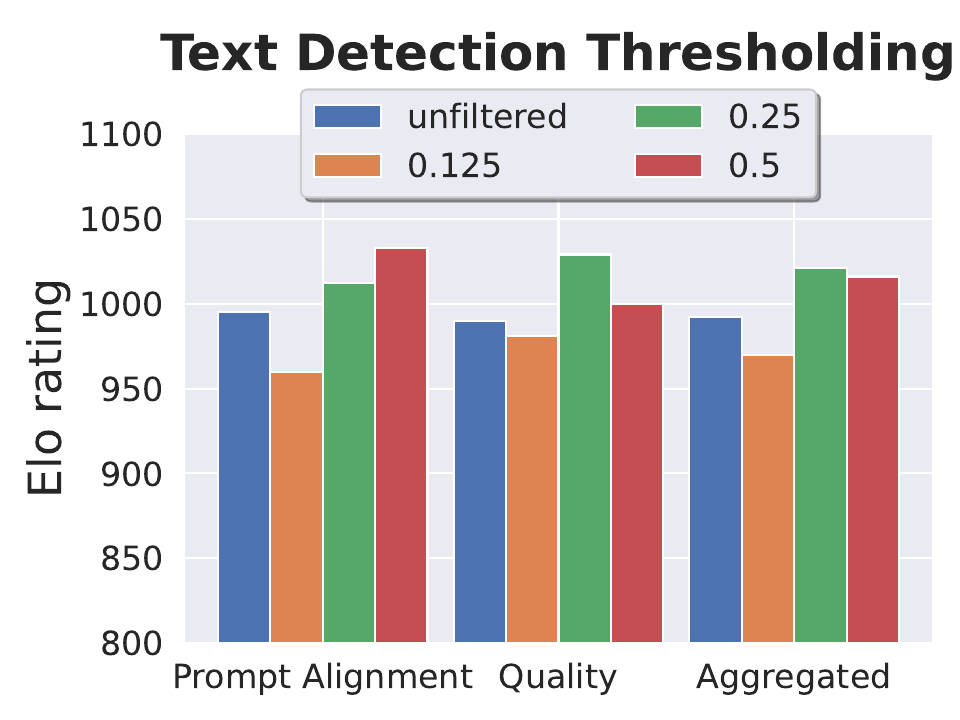}
    \caption{Results of the dedicated experiments conducted to identify most useful filtering thresholds for each ablation axis. For of these ablation studies we train four identical models using the architecture detailed in \Cref{supsubsec:filtering_ablations} on different subset of \datasetsmall, which we create by systematically increasing the thresholds which corresponds to filter out more and more examples.
    \label{fig:filtering_ablations}
    }
\end{figure*}
}

\newcommand{\ocrvis}{
\begin{figure}[htbp]
\begin{center}
\renewcommand{\arraystretch}{1.5}  %
\begin{adjustbox}{max width=.98\textwidth}
\begin{tabular}{cc}
\midrule
 \multirow{2}{*}{\textbf{Source Video}\textsuperscript{\phantom{*}}} & \multirow{2}{*}{\textbf{Text Area Ratio}}\\ \\
\toprule 
\raisebox{-0.5\height}{\includegraphics[width=\textwidth]{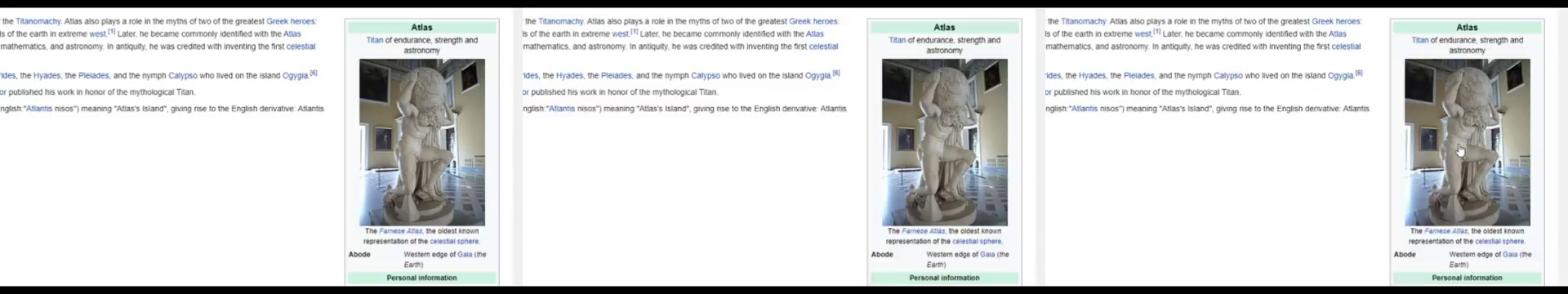}} & \large 0.102 \\
\bottomrule
\end{tabular}
\end{adjustbox}
\end{center}
\caption{\label{fig:ocr_vis}
An example of a video with lots of unwanted text. We apply text-detection and annotate bounding boxes around text, and then compute the ratio between the area of all the boxes and the size of the frame. 
}
\end{figure}
}

\newcommand{\cutsfadesvis}{
\begin{figure*}[htbp]
\begin{center}
\renewcommand{\arraystretch}{1.5}  %
\begin{adjustbox}{max width=.98\textwidth}
\begin{tabular}{ccc}
\midrule
\multirow{2}{*}{\textbf{Source Video}\textsuperscript{\phantom{*}}

} & \multicolumn{2}{c}{\textbf{Cut Detected?}} \\[-.5em]
& w/o cascade &  w/ cascade \textbf{(ours)}\\
\toprule 
\raisebox{-0.5\height}{\includegraphics[width=\textwidth]{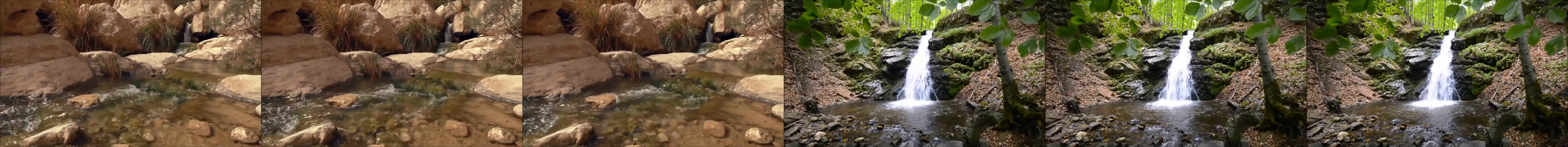}} &  \cmark & \cmark \\
\raisebox{-0.5\height}{\includegraphics[width=\textwidth]{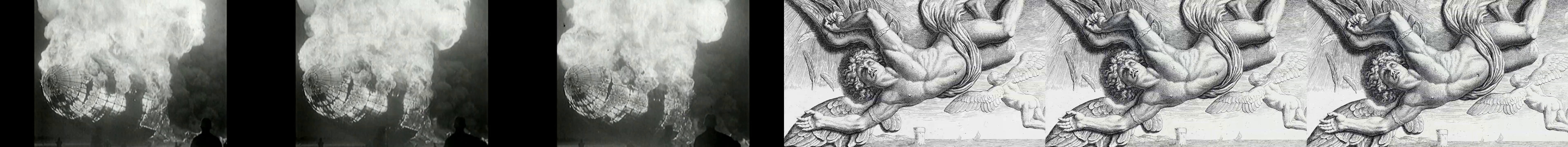}} &  \cmark & \cmark \\
\raisebox{-0.5\height}{\includegraphics[width=\textwidth]{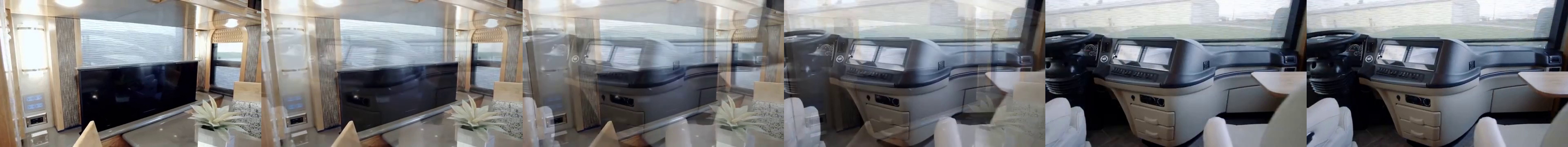}} & \xmark & \cmark \\
\raisebox{-0.5\height}{\includegraphics[width=\textwidth]{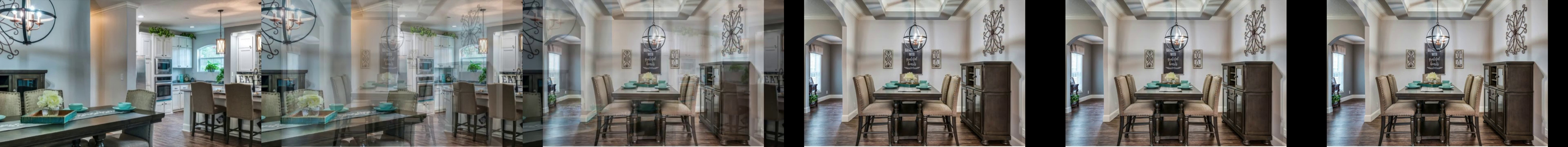}} &  \xmark & \cmark \\
\bottomrule
\end{tabular}
\end{adjustbox}
\end{center}
\caption{\label{fig:cuts_fades} 
Comparing a common cut detector with our cascaded approach, shows the benefits of our cascaded method: While normal single-fps cut detection 
can only detect sudden changes in scene, more continuous transitions tend to remain undetected, what is in contrast with our approach which reliably also detects the latter transitions. 
}
\end{figure*}
}

\newcommand{\captionexamples}{
\begin{figure*}[htbp]
\begin{center}
\renewcommand{\arraystretch}{1.5}  %
\begin{adjustbox}{max width=.98\textwidth}
\begin{tabular}{c >{\centering\arraybackslash}p{3cm} >{\centering\arraybackslash}p{3cm} >{\centering\arraybackslash}p{3cm}}
\midrule
\multirow{2}{*}{\textbf{Source Video
}\textsuperscript{\phantom{*}}} & \multicolumn{3}{c}{\textbf{Caption}} \\[-.5em]
& \multicolumn{1}{c}{CoCa} & \multicolumn{1}{c}{VBLIP} & \multicolumn{1}{c}{LLM} \\
\toprule 
\raisebox{-0.8\height}{\includegraphics[width=\textwidth]{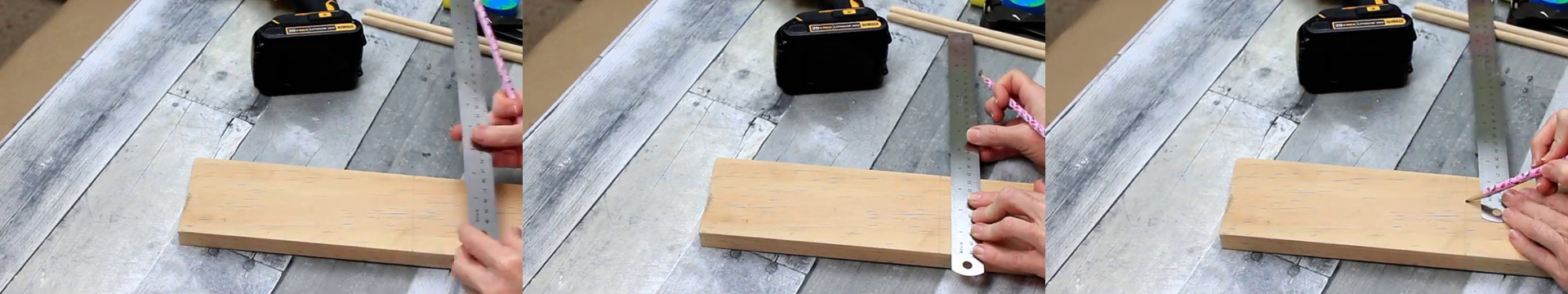}} & there is a piece of wood on the floor next to a tape measure . & a person is using a ruler to measure a piece of wood & A person is using a ruler to measure a piece of wood on the floor next to a tape measure. \\
\raisebox{-0.8\height}{\includegraphics[width=\textwidth]{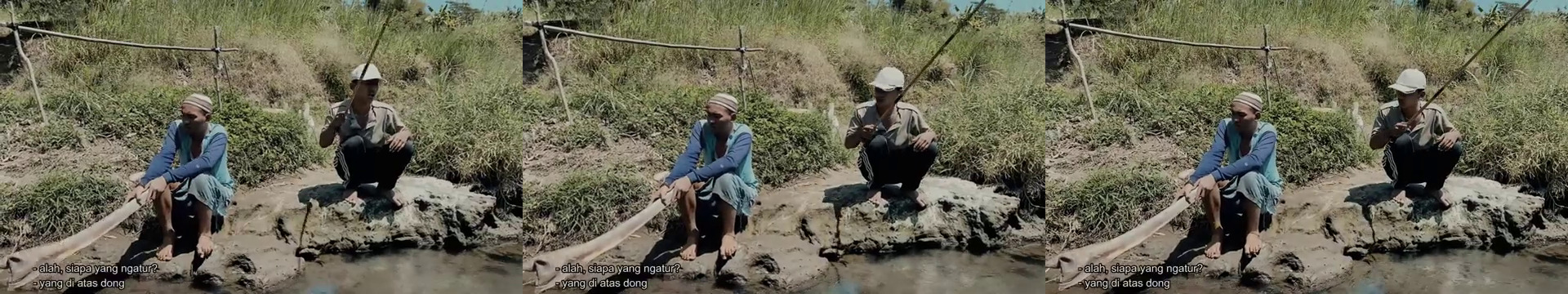}} & two men sitting on a rock near a river . one is holding a stick and the other is holding a pole . & two people are fishing in a river & Two men are fishing in a river. One is holding a stick and the other is holding a pole. \\
\bottomrule
\end{tabular}
\end{adjustbox}
\end{center}
\caption{\label{fig:captionextable} 
Comparison of various synthetic captioners. We observe that CoCa often captures good spatial details, whereas VBLIP tends to capture temporal details. We use an LLM to combine these two, and experiment with all three types of synthetic captions. }%
\end{figure*}
}

\newcommand{\MVIone}{
\begin{figure*}
    \centering
    \includegraphics[width=\linewidth]{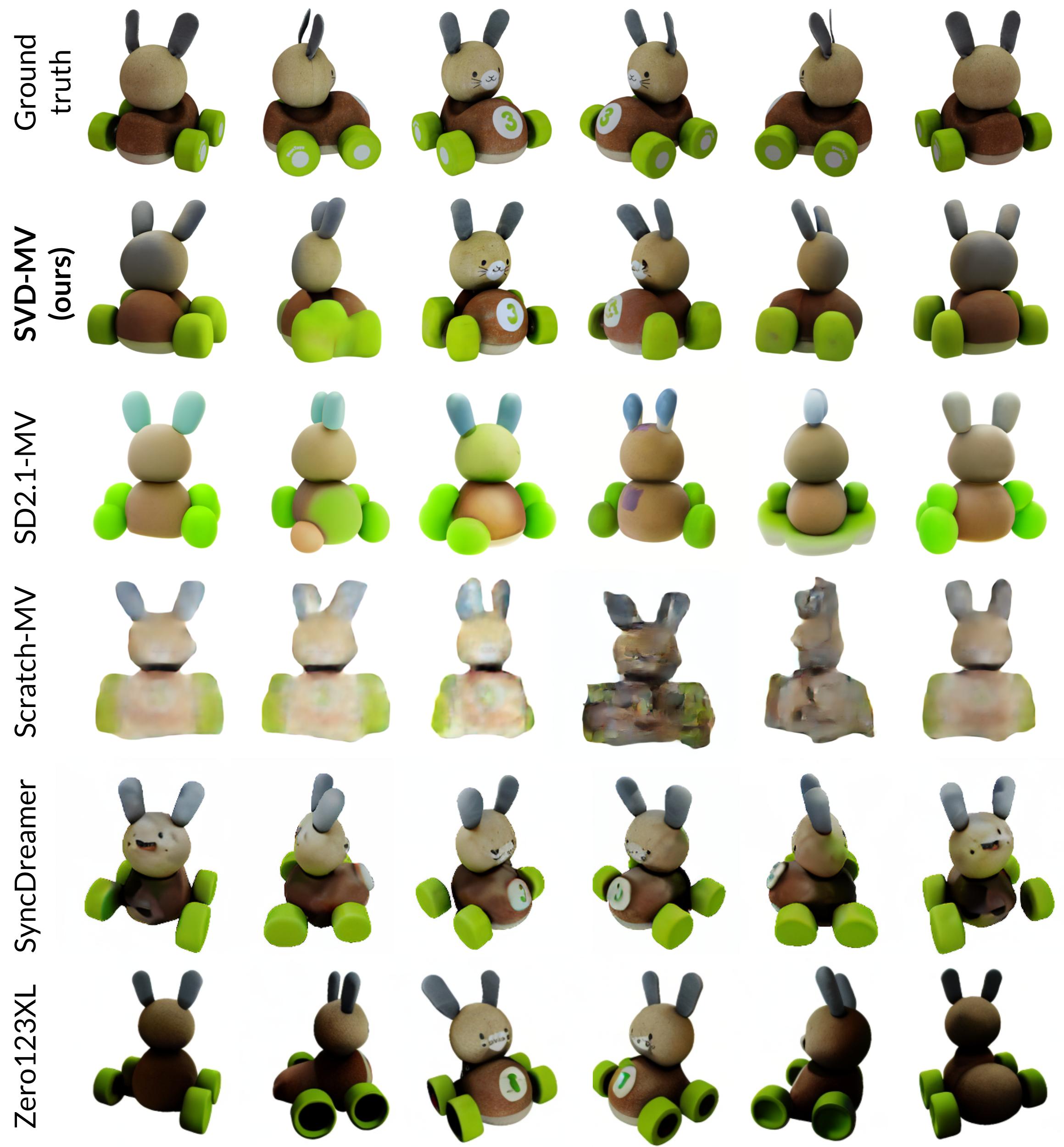}
    \caption{\label{fig:additional_mv1}\small Additional image-to-multi-view generation samples from GSO test dataset, using our SVD-MV model trained on Objaverse, and comparison with other methods.}
    \label{fig:MV1}
    \vspace{-10pt}
\end{figure*}
}

\newcommand{\MVItwo}{
\begin{figure}
    \centering
    \includegraphics[width=\linewidth]{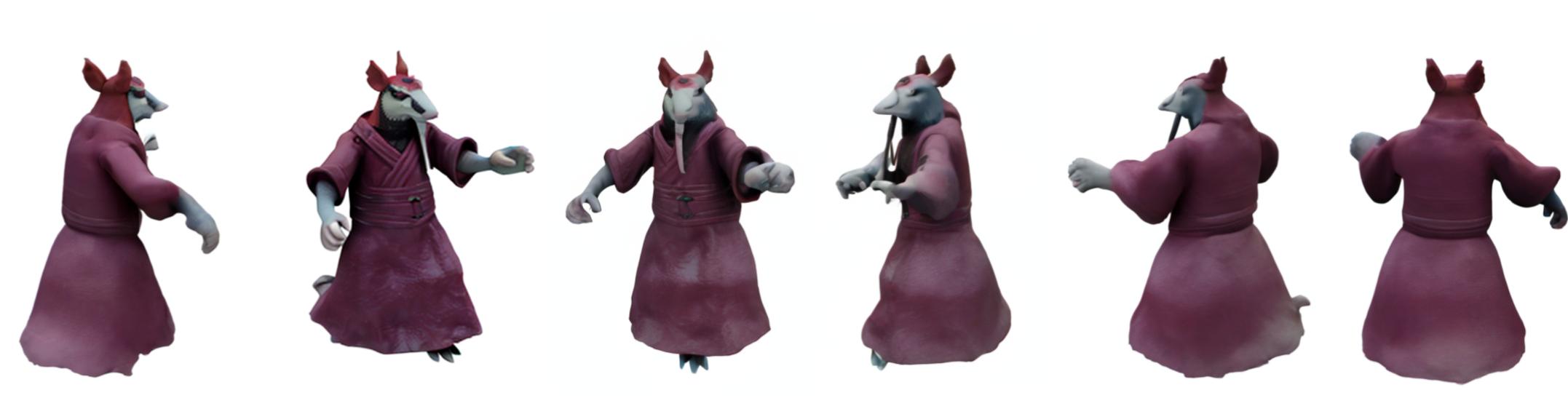}
    \caption{\label{fig:additional_mv2}\small Additional image-to-multi-view generation samples from GSO test dataset, using our SVD-MV model trained on Objaverse}
    \label{fig:MVI2}
    \vspace{-10pt}
\end{figure}
}

\newcommand{\MVIthree}{
\begin{figure}
    \centering
    \includegraphics[width=\linewidth]{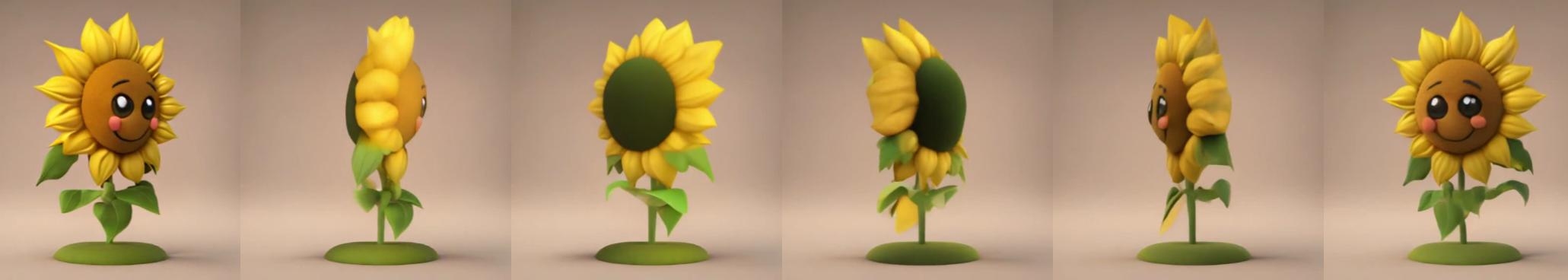}
    \caption{\label{fig:additional_mv3}\small Text-to-image-to-multi-view generation samples: text to image using SDXL with the prompt "Centered 3D model of a cute anthropomorphic sunflower figure (plain background, unreal engine render 4k)", and image-to-multi-view using our SVD-MV model trained on Objaverse}
    \label{fig:MVI2}
    \vspace{-10pt}
\end{figure}
}

\newcommand{\MVIfour}{
\begin{figure}
    \centering
    \includegraphics[width=\linewidth]{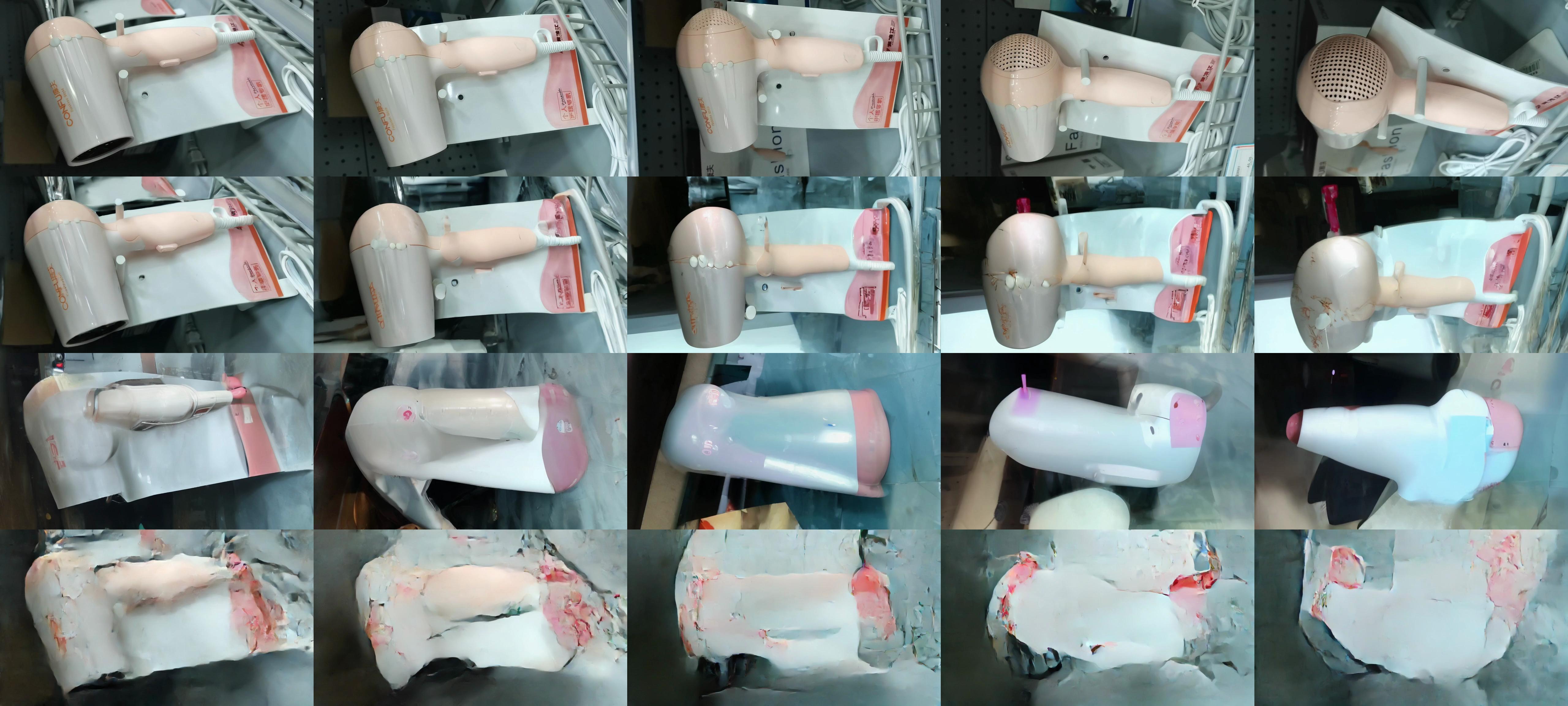}
    \caption{\label{fig:additional_mv4}\small Additional multi-view generation samples from MVI dataset, using our SVD-MV model trained on MVImgNet, and comparison with other methods. Top row is ground truth frames, second row is sample frames from SVD-MV (ours), third row is from SD2.1-MV, bottom row is from Scratch-MV}
    \label{fig:MVI3}
    \vspace{-10pt}
\end{figure}
}

\section{Broader Impact and Limitations} 
\label{supsec:broader_impact_and_limitations}
\textbf{Broader Impact:} 
Generative models for different modalities promise to revolutionize the landscape of media creation and use. While exploring their creative applications, 
reducing the potential to use them for creating misinformation and harm are crucial aspects before real-world deployment.
Furthermore, risk analyses need to highlight and evaluate the differences between the various existing model types, such as interpolation, text-to-video, animation, and long-form generation.
Before these models are used in practice, a thorough investigation of the models themselves, their intended uses, safety aspects, associated risks, and potential biases is essential.
 \\
\textbf{Limitations:} 
While our approach excels at short video generation, it comes with some fundamental shortcomings w.r.t. long video synthesis: 
Although a latent approach provides efficiency benefits, 
generating multiple keyframes at once is expensive both during training but also inference,  and future work on long video synthesis should either try a cascade of very coarse frame generation or build dedicated tokenizers for video generation.
Furthermore, videos generated with our approach sometimes suffer from too little generated motion. 
Lastly, video diffusion models are typically slow to sample and have high VRAM requirements, and our model is no exception. Diffusion distillation methods~\cite{meng2023distillation, salimans2022progressive, ho2022imagenvideo} are promising candidates for faster synthesis.

\section{Related Work}
\label{supsec:related_work}

\textbf{Video Synthesis.} 
Many approaches based on various models such as variational RNNs~\cite{babaeizadeh2018stochastic,svg,lee2018savp,hvrnn,lsvg}, normalizing flows~\citep{si2v,ipoke}, autoregressive transformers~\cite{Weissenborn2020Scaling,yan2021videogpt,hong2022cogvideo,wu2021godiva,wu2022nuwa,ge2022longvideo,Gupta_2022_CVPR}, and GANs~\citep{scene_dyn,yu2022generating,tian2021a, villegas17mcnet,Luc2020TransformationbasedAV,TGAN2020,brooks2022generating,Skorokhodov_2022_CVPR,kahembwe2020lower,TGAN2017,Wang_2020_CVPR,fox2021stylevideogan} have tackled video synthesis. Most of these works, however, have generated videos either on low-resolution~\citep{babaeizadeh2018stochastic,svg,lee2018savp,hvrnn,lsvg,si2v,ipoke,scene_dyn,yu2022generating,tian2021a, villegas17mcnet,Luc2020TransformationbasedAV} or on comparably small and noisy datasets~\cite{carreira2017quo,soomro2012ucf101,xu2016msr-vtt} which were originally proposed to train discriminative models. 

Driven by increasing amounts of available compute resources and datasets better suited for generative modeling such as WebVid-10M~\citep{bain2022frozen}, more competitive approaches have been proposed recently, mainly based on well-scalable, explicit likelihood-based approaches such as diffusion~\cite{ho2022video,singer2022make,ho2022imagenvideo} and autoregressive models~\citep{villegas2022phenaki}. Motivated by a lack of available clean video data, all these approaches are leveraging joint image-video training~\citep{singer2022make,ho2022imagenvideo,zhou2022magicvideo,blattmann2023align} and most methods are grounding their models on pretrained image models~\citep{singer2022make,zhou2022magicvideo,blattmann2023align}. Another commonality between these and most subsequent approaches to (text-to-)video synthesis~\cite{ge2023preserve,wang2023modelscope,wang2023lavie} is the usage of dedicated expert models to generate the actual visual content at a coarse frame rate and to temporally upscale this low-fps video to temporally smooth final outputs at 24-32 fps~\citep{singer2022make,ho2022imagenvideo,blattmann2023align}. Similar to the image domain, diffusion-based approaches can be mainly separated into cascaded approaches~\citep{ho2022imagenvideo} following~\citep{ho2021cascaded,ge2023preserve} and latent diffusion models~\citep{blattmann2023align,zhou2022magicvideo,zhang2023i2vgen} translating the approach of \citet{rombach2021high} to the video domain. While most of these works aim at learning general motion representation and are consequently trained on large and diverse datasets, another well-recognized branch of diffusion-based video synthesis tackles personalized video generation based on finetuning of pretrained text-to-image models on more narrow datasets tailored to a specific domain~\citep{guo2023animatediff} or application, partly including non-deep motion priors~\citep{zhang2023i2vgen}. Finally, many recent works tackle the task of image-to-video synthesis, where the start frame is already given, and the model has to generate the consecutive frames~\cite{zhang2023i2vgen,wang2023modelscope,guo2023animatediff}. Importantly, as shown in our work (see~\Cref{fig:teaser}) when combined with off-the-shelf text-to-image models, image-to-video models can be used to obtain a full text-(to-image)-to-video pipeline.

\textbf{Multi-View Generation}
Motivated by their success in 2D image generation, diffusion models have also been used for multi-view generation. Early promising diffusion-based results~\citep{watson2022novel,zhou2023sparsefusion,anciukevivcius2023renderdiffusion,nichol2022pointe,jun2023shape,deng2023nerdi} have mainly been restricted by lacking availability of useful real-world multi-view training data.     
To address this, more recent works such as Zero-123~\cite{liu2023zero1to3}, MVDream~\cite{shi2023mvdream}, and SyncDreamer~\cite{liu2023syncdreamer} propose techniques to adapt and finetune pretrained image generation models such as Stable Diffusion (SD) for multi-view generation, thereby leveraging image priors from SD.
One issue with Zero-123~\cite{liu2023zero1to3} is that the generated multi-views can be inconsistent with respect to each other as they are generated independently with pose-conditioning. 
Some follow-up works try to address this view-consistency problem by jointly synthesizing the multi-view images. MVDream~\cite{shi2023mvdream} proposes to jointly generate four views of an object using a shared attention module across images. SyncDreamer~\cite{liu2023syncdreamer} proposes to estimate a 3D voxel structure in parallel to the multi-view image diffusion process to maintain consistency across the generated views.

Despite rapid progress in multi-view generation research, these approaches rely on single image generation models such as SD. We believe that our video generative model is a better candidate for the multi-view generation as multi-view images form a specific form of video where the camera is moving around an object. As a result, it is much easier to adapt a video-generative model for multi-view generation compared to adapting an image-generative model. In addition, the temporal attention layers in our video model naturally assist in the generation of consistent multi-views of an object without needing any explicit 3D structures like in~\cite{liu2023syncdreamer}.

\section{\textbf{Data Processing}}
\label{supsec:data_proc}
In this section, we provide more details about our processing pipeline including their outputs on a few public video examples for demonstration purposes. 

\paragraph{Motivation}
We start from a large collection of raw video data which is not useful for generative text-video (pre)training~\citep{ramesh2022hierarchical,wang2023internvid} because of the following adverse properties: First, in contrast to discriminative approaches to video modeling, generative video models are sensitive to motion inconsistencies such as cuts of which usually many are contained in raw and unprocessed video data, \cf \Cref{fig:cuts_and_motion}, left. Moreover, our initial data collection is biased towards still videos as indicated by the peak at zero motion in \Cref{fig:cuts_and_motion}, right. Since generative models trained on this data would obviously learn to generate videos containing cuts and still scenes, this emphasizes the need for cut detection and motion annotations to ensure temporal quality. Another critical ingredient for training generative text-video models are captions - ideally more than one per video~\citep{somepalli2023understanding} - which are well-aligned with the video content. The last essential component for generative video training which we are considering here is the high visual quality of the training examples.

The design of our processing pipeline addresses the above points. Thus, to ensure temporal quality, we detect cuts with a cascaded approach directly after download, clip the videos accordingly, and estimate optical flow for each resulting video clip. After that, we apply three synthetic captioners to every clip and further extract frame-level CLIP similarities to all of these text prompts to be able to filter out outliers. Finally, visual quality at the frame level is assessed by using a CLIP-embeddings-based aesthetics score~\citep{schuhmann2022laion}. We describe each step in more detail in what follows.

\cutsfadesvis

\paragraph{Cascaded Cut Detection.}
Similar to previous work~\citep{wang2023internvid}, we use PySceneDetect~\footnote{\url{https://github.com/Breakthrough/PySceneDetect}} to detect cuts in our base video clips. However, as qualitatively shown in \Cref{fig:cuts_fades} we observe many fade-ins and fade-outs between consecutive scenes, which are not detected when running the cut detector at a unique threshold and only native fps. Thus, in contrast to previous work, we apply a cascade of 3 cut detectors which are operating at different frame rates and different thresholds to detect both sudden changes and slow ones such as fades.

\paragraph{Keyframe-Aware Clipping.} We clip the videos using FFMPEG~\citep{tomar2006converting} directly after cut detection by extracting the timestamps of the keyframes in the source videos and snapping detected cuts onto the closest keyframe timestamp, which does not cross the detected cut. This allows us to quickly extract clips without cuts via seeking and isn't prohibitively slow at scale like inserting new keyframes in each video.

\staticvis
\paragraph{Optical Flow.} 
As motivated in \Cref{subsec:data_proc} and \Cref{fig:cuts_and_motion} it is crucial to provide means for filtering out static scenes. To enable this, we extract dense optical flow maps at 2fps using the OpenCV~\citep{itseez2015opencv} implementation of the Farneb\"ack algorithm~\citep{farneback2003flow}. To further keep storage size tractable we spatially downscale the flow maps such that the shortest side is at 16px resolution. By averaging these maps over time and spatial coordinates, we further obtain a global motion score for each clip, which we use to filter out static scenes by using a threshold for the minimum required motion, which is chosen as detailed on \Cref{supsubsec:filtering_ablations}. Since this only yields rough approximate, for the final Stage III finetuning, we compute more accurate dense optical flow maps using RAFT~\cite{teed2020raft} at $800 \times 450$ resolution. The motion scores are then computed similarly. Since the high-quality finetuning data is relatively much smaller than the pretraining dataset, this makes the RAFT-based flow computation tractable.

\captionexamples
\paragraph{Synthetic Captioning.} At a million-sample scale, it is not feasible to hand-annotate data points with prompts. Hence we resort to synthetic captioning to extract captions. However in light of recent insights on the importance of caption diversity~\citep{somepalli2023understanding} and taking potential failure cases of these synthetic captioning models into consideration, we extract \emph{three} captions per clip by using i) the image-only captioning model CoCa~\citep{pucetti2023training}, which describes spatial aspects well, ii) - to also capture temporal aspects - the video-captioner VideoBLIP~\citep{videoblip} and iii) to combine these two captions and like that, overcome potential flaws in each of them, a lightweight LLM. Examples of the resulting captions are shown in \Cref{fig:captionextable}.

\paragraph{Caption similarities and Aesthetics.} Extracting CLIP~\citep{radford2021learning} image and text representations have proven to be very helpful for data curation in the image domain since computing the cosine similarity between the two allows for assessment of text-image alignment for a given example~\citep{schuhmann2022laion} and thus to filter out examples with erroneous captions. Moreover, it is possible to extract scores for visual aesthetics~\citep{schuhmann2022laion}. Although CLIP is only able to process images, and this consequently is only possible on a single frame level we opt to extract both CLIP-based i) text-image similarities and ii) aesthetics scores of the first, center, and last frames of each video clip. As shown in \Cref{subsec:data_curation,supsubsec:filtering_ablations}, using training text-video models on data curated by using these scores improves i) text following abilities and ii) visual quality of the generated samples compared to models trained on unfiltered data.

\paragraph{Text Detection.} 
In early experiments, we noticed that models trained on earlier versions of \datasetfiltered obtained a tendency to generate videos with excessive amounts of written text depicted which is arguably not a desired feat for a text-to-video model. To this end, we applied the off-the-shelf text-detector CRAFT~\citep{baek2019character} to annotate the start, middle, and end frames of each clip in our dataset with bounding box information on all written text. Using this information, we filtered out all clips with a total area of detected bounding boxes larger than 7\% to construct the final \datasetfiltered. 
\ocrvis
\section{Model and Implementation Details} \label{supsec:model_and_implementation_details}
\subsection{Diffusion Models} \label{subsec:diffusion_models}
In this section, we give a concise summary of DMs. We make use of the continuous-time DM framework~\citep{song2020score,karras2022elucidating}. 
Let $\pdata(\rvx_0)$ denote the data distribution and let $p(\rvx; \sigma)$ be the distribution obtained by adding i.i.d. $\sigma^2$-variance Gaussian noise to the data. Note that or sufficiently large $\sigma_{\mathrm{max}}$, $p(\rvx; \sigma_{\mathrm{max}^2}) \approx \gN\left(\bm{0}, \sigma_{\mathrm{max}^2}\right)$. DM uses this fact and, starting from high variance Gaussian noise $\rvx_M \sim\gN\left(\bm{0},\sigma_{\mathrm{max}^2}\right)$, sequentially denoise towards $\sigma_0=0$. In practice, this iterative refinement process can be implemented through the numerical simulation of the \emph{Probability Flow} ordinary differential equation (ODE)~\citep{song2020score}
\begin{align} \label{eq:probability_flow_ode}
    d\rvx = -\dot \sigma(t) \sigma(t) \nabla_\rvx \log p(\rvx; \sigma(t)) \, dt,
\end{align}
where $\nabla_\rvx \log p(\rvx; \sigma)$ is the \emph{score function}~\citep{hyvarinen2005estimation}. DM training reduces to learning a model $\vs_\vtheta(\rvx; \sigma)$ for the score function $\nabla_\rvx \log p(\rvx; \sigma)$. 
The model can, for example, be parameterized as $\nabla_\rvx \log p(\rvx; \sigma) \approx s_\vtheta(\rvx; \sigma) = (D_\vtheta(\rvx; \sigma) - \rvx)/ \sigma^2$~\citep{karras2022elucidating}, where $D_\vtheta$ is a learnable \emph{denoiser} that tries to predict the clean $\rvx_0$. 
The denoiser $D_\vtheta$ is trained via \emph{denoising score matching}~(DSM)
\begin{align} \label{eq:diffusion_objective}
    \E_{\substack{(\rvx_0, \rvc) \sim \pdata(\rvx_0, \rvc), (\sigma, \rvn) \sim p(\sigma, \rvn)}} \left[\lambda_\sigma \|D_\vtheta(\rvx_0 + \rvn; \sigma, \rvc) - \rvx_0 \|_2^2 \right],
\end{align}
where $p(\sigma, \rvn) = p(\sigma)\,\gN\left(\rvn; \bm{0}, \sigma^2\right)$, 
$p(\sigma)$ can be a probability distribution or density over noise levels $\sigma$. 
It is both possible to use a discrete set or a continuous range of noise levels. In this work, we use both options, which we further specify in \Cref{subsec:base_model_training_and_architecture}.   

$\lambda_\sigma \colon \R_+ \to \R_+$ is a weighting function, and $\rvc$ is an arbitrary conditioning signal. In this work, we follow the EDM-preconditioning framework~\citep{karras2022elucidating}, parameterizing the learnable denoiser $D_\vtheta$ as
\begin{align} \label{eq:edm_preconditioning}
    D_\vtheta(\rvx; \sigma) = \cskip(\sigma) \rvx + \cout(\sigma) F_\vtheta(\cin(\sigma) \rvx; \cnoise(\sigma)),
\end{align}
where $F_\vtheta$ is the network to be trained.

\textbf{Classifier-free guidance.} Classifier-free guidance~\citep{ho2022classifier} is a method used to guide the iterative refinement process of a DM towards a conditioning signal $\rvc$. The main idea is to mix the predictions of a conditional and an unconditional model
\begin{align} \label{eq:guidance}
    D^w(\rvx; \sigma, \rvc) = w D(\rvx; \sigma,\rvc) - (w - 1) D(\rvx; \sigma),
\end{align}
where $w \geq 0$ is the \emph{guidance strength}. The unconditional model can be trained jointly alongside the conditional model in a single network by randomly replacing the conditional signal $\rvc$ with a null embedding in~\Cref{eq:diffusion_objective}, e.g., 10\% of the time~\citep{ho2022classifier}. In this work, we use classifier-free guidance, for example, to guide video generation toward text conditioning.

\subsection{Base Model Training and Architecture} \label{subsec:base_model_training_and_architecture}
As discussed in~\label{sec:base-model}, we start the publicly available \emph{Stable Diffusion} 2.1~\citep{rombach2021high} (SD 2.1) model. In the EDM-framework~\citep{karras2022elucidating}, SD 2.1 has the following preconditioning functions:
\begin{align}
    \cskip^\mathrm{SD 2.1}(\sigma) &= 1, \\
    \cout^\mathrm{SD 2.1}(\sigma) &= -\sigma\,, \\
    \cin^\mathrm{SD 2.1}(\sigma) &= \frac{1}{\sqrt{\sigma^2 + 1}}\, ,  \\
    \cnoise^\mathrm{SD 2.1}(\sigma) &= \argmin_{j \in [1000]} (\sigma - \sigma_j)\,, \\
\end{align}
where $\sigma_{j+1} > \sigma_j$. The distribution over noise levels $p(\sigma)$ used for the original SD 2.1. training is a uniform distribution over the 1000 \emph{discrete} noise levels $\{\sigma_j\}_{j \in [1000]}$. One issue with the training of SD 2.1 (and in particular its noise distribution $p(\sigma)$) is that even for the maximum discrete noise level $\sigma_{1000}$ the \emph{signal-to-noise ratio}~\citep{kingma2021variational} is still relatively high which results in issues when, for example, generating very dark images~\citep{lin2023common,guttenberg2023diffusion}. \citet{guttenberg2023diffusion} proposed \emph{offset noise}, a modification of the training objective in~\Cref{eq:diffusion_objective} by making $p(\rvn \mid \sigma)$ non-isotropic Gaussian. In this work, we instead opt to modify the preconditioning functions and distribution over training noise levels altogether. 

\textbf{Image model finetuning.} We replace the above preconditioning functions with 
\begin{align}
    \cskip(\sigma) &= \left( \sigma^2 + 1 \right)^{-1}\,, \\
    \cout(\sigma) &=  \frac{-\sigma}{\sqrt{\sigma^2 + 1}}\,, \\
    \cin(\sigma) &= \frac{1}{\sqrt{\sigma^2 + 1}}\, , \\
    \cnoise(\sigma) &= 0.25 \log \sigma, \\
\end{align}
which can be recovered in the EDM framework~\citep{karras2022elucidating} by setting $\sigma_\mathrm{data} = 1$); the preconditioning functions were originally proposed in~\citep{salimans2022progressive}. We also use the noise distribution and weighting function proposed in~\citet{karras2022elucidating}, namely $\log \sigma \sim \gN(P_\mathrm{mean}, P_\mathrm{std}^2)$ and $\lambda(\sigma) = (1 + \sigma^2) \sigma^{-2}$, with $P_\mathrm{mean}=-1.2$ and $P_\mathrm{std}=1$. We then finetune the neural network backbone $F_\vtheta$ of SD2.1 for 31k iterations using this setup. For the first 1k iterations, we freeze all parameters of $F_\vtheta$ except for the time-embedding layer and train on SD2.1's original training resolution of $512 \times 512$. This allows the model to adapt to the new preconditioning functions without unnecessarily modifying the internal representations of $F_\vtheta$ too much. Afterward, we train all layers of $F_\vtheta$ for another 30k iterations on images of size $256 \times 384$, which is the resolution used in the initial stage of video pretraining. 

\textbf{Video pretraining.} We use the resulting model as the image backbone of our video model. We then insert temporal convolution and attention layers. In particular, we follow the exact setup from~\citep{blattmann2023align}, inserting a total of 656M new parameters into the UNet bumping its total size (spatial and temporal layers) to 1521M parameters. We then train the resulting UNet on 14 frames on resolution $256\times384$ for 150k iters using AdamW~\citep{loshchilov2017decoupled} with learning rate $10^{-4}$ and a batch size of 1536. We train the model for classifier-free guidance~\citep{ho2021classifierfree} and drop out the text-conditioning 15\% of the time. Afterward, we increase the spatial resolution to $320 \times 576$ and train for an additional 100k iterations, using the same settings as for the lower-resolution training except for a reduced batch size of 768 and a shift of the noise distribution towards more noise, in particular, we increase $P_\mathrm{mean} = 0$. During training, the base model and the high-resolution Text/Image-to-Video models are all conditioned on the input video's frame rate and motion score. This allows us to vary the amount of motion in a generated video at inference time.
\subsection{High-Resolution Text-to-Video Model} \label{subsec:high_resolution_text_to_video_model}
We finetune our base model on a high-quality dataset of $\sim$ 1M samples at resolution $576 \times 1024$. We train for $50k$ iterations at a batch size of 768, learning rate $3 \times 10^{-5}$, and set $P_\mathrm{mean} = 0.5$ and $P_\mathrm{std} = 1.4$. Additionally, we track an exponential moving average of the weights at a decay rate of 0.9999. The final checkpoint is chosen using a combination of visual inspection and human evaluation.
\subsection{High-Resolution Image-to-Video Model} \label{subsec:high_resolution_image_to_video_model}
We can finetune our base text-to-video model for the image-to-video task. In particular, during training, we use one additional frame on which the model is conditioned. We do not use text-conditioning but rather replace text embeddings fed into the base model with the CLIP image embedding of the conditioning frame. Additionally, we concatenate a noise-augmented~\citep{ho2021cascaded} version of the conditioning frame channel-wise to the input of the UNet~\cite{ronneberger2015u}. In particular, we add a small amount of noise of strength $\log \sigma \sim \gN(-3.0, 0.5^2)$ to the conditioning frame and then feed it through the standard SD 2.1 encoder. The mean of the encoder distribution is then concatenated to the input of the UNet (copied across the time axis). Initially, we finetune our base model for the image-to-video task on the base resolution ($320 \times 576$) for 50k iterations using a batch size of 768 and learning rate $3 \times 10^{-5}$. Since the conditioning signal is very strong, we again shift the noise distribution towards more noise, i.e., $P_\mathrm{mean} = 0.7$ and $P_\mathrm{std} = 1.6$. Afterwards, we fintune the base image-to-video model on a high-quality dataset of $\sim$ 1M samples at $576 \times 1024$ resolution. We train two versions: one to generate 14 frames and one to generate 25 frames. We train both models for $50k$ iterations at a batch size of 768, learning rate $3 \times 10^{-5}$, and set $P_\mathrm{mean} = 1.0$ and $P_\mathrm{std} = 1.6$. Additionally, we track an exponential moving average of the weights at a decay rate of 0.9999. The final checkpoints are chosen using a combination of visual inspection and human evaluation.
\subsubsection{Linearly Increasing Guidance} We occasionally found that standard vanilla classifier-free guidance~\citep{ho2021classifierfree} (see~\Cref{eq:guidance}) can lead to artifacts: too little guidance may result in inconsistency with the conditioning frame while too much guidance can result in oversaturation. Instead of using a constant guidance scale, we found it helpful to linearly increase the guidance scale across the frame axis (from small to high). A PyTorch implementation of this novel technique can be found in~\Cref{fig:linear_pred_guidance}.
\linearpredictioncode
\subsubsection{Camera Motion LoRA}
To facilitate controlled camera motion in image-to-video generation, we train a variety of \emph{camera motion LoRAs} within the temporal attention blocks of our model~\citep{guo2023animatediff}. In particular, we train low-rank matrices of rank 16 for 5k iterations. Additional samples can be found in~\Cref{fig:additional_motion_lora}.

\subsection{Interpolation Model Details} 
\label{supsubsec:interpolation_model_details} 
Similar to the text-to-video and image-to-video models, we finetune our interpolation model starting from the base text-to-video model, \cf \Cref{subsec:base_model_training_and_architecture}. To enable interpolation, we reduce the number of output frames from 14 to 5, of which we use the first and last as conditioning frames, which we feed to the UNet~\citep{ronneberger2015u} backbone of our model via the concat-conditioning-mechanism~\citep{rombach2021high}. To this end, we embed these frames into the latent space of our autoencoder, resulting in two image encodings $z_s,\, z_e \in \mathbb{R}^{c \times h \times w}$, where $c=4, \,h=52, \,w=128$. To form a latent frame sequence that is of the same shape as the noise input of the UNet, \ie $\mathbb{R}^{5 \times c \times h \times w}$, we use a learned mask embedding $z_m \in\mathbb{R}^{c \times h \times w}$ and form a latent sequence $ \boldsymbol{z} = \{z_s, z_m, z_m, z_m, z_e\}\in \mathbb{R}^{5 \times c \times h \times w}$. We concatenate this sequence channel-wise with the noise input and additionally with a binary mask where 1 indicates the presence of a conditioning frame and 0 that of a mask embedding. The final input for the UNet is thus of shape $\left( 5, 9, 52, 128 \right)$. In line with previous work~\citep{ho2022imagenvideo,singer2022make,blattmann2023align}, we use noise augmentation for the two conditioning frames, which we apply in the latent space. Moreover, we replace the CLIP text representation for the crossattention conditioning with the corresponding CLIP image representation of the start frame and end frame, which we concatenate to form a conditioning sequence of length 2.  

We train the model on our high-quality dataset at spatial resolution $576 \times 1024$ using AdamW~\citep{loshchilov2017decoupled} with a learning rate of $10^{-4}$ in combination with exponential moving averaging at decay rate 0.9999 and use a shifted noise schedule with  $P_\mathrm{mean}=1$ and $P_\mathrm{std}=1.2$. Surprisingly, we find this model, which we train with a comparably small batch size of 256, to converge extremely fast and to yield consistent and smooth outputs after only 10k iterations. We take this as another evidence of the usefulness of the learned motion representation our base text-to-video model has learned.   

\subsection{Multi-view generation}

We finetuned the high-resolution image-to-video model on our specific rendering of the Objaverse dataset. We render 21 frames per orbit of an object in the dataset at $576 \times 576$ resolution and finetune the 25-frame Image-to-Video model to generate these 21 frames. We feed one view of the object as the image condition. In addition, we feed the elevation of the camera as conditioning to the model. We first pass the elevation through a timestep embedding layer that embeds the sine and cosine of the elevation angle at various frequencies and concatenates them into a vector. This vector is finally concatenated to the overall vector condition of the UNet.

We trained for 12$k$ iterations with a total batch size of 16 across 8 A100 GPUs of 80GB VRAM at a learning rate of $1 \times 10^{-5}$.

\section{Experiment Details} \label{supsec:experiment_details}

\subsection{Details on Human Preference Assessment}
\label{supsubsec:human_eval}

For most of the evaluation conducted in this paper, we employ human evaluation as we observed it to contain the most reliable signal. For text-to-video tasks and all ablations conducted for the base model, we generate video samples from a list of 64 test prompts. We then employ human annotators to collect preference data on two axes: i) visual quality and ii) prompt following. More details on how the study was conducted~\Cref{supsubsubsec:human_eval_setup} and the rankings computed~\Cref{supsubsubsec:elo_score} are listed below.

\subsubsection{Experimental Setup}
\label{supsubsubsec:human_eval_setup}

Given all models in one ablation axis (\eg four models of varying aesthetic or motion scores), we compare each prompt for each pair of models (1v1). For every such comparison, we collect on average three votes per task from different annotators, i.e., \ three each for visual quality and prompt following, respectively. Performing a complete assessment between all pair-wise comparisons gives us robust and reliable signals on model performance trends and the effect of varying thresholds.
Sample interfaces that the annotators interact with are shown in~
\Cref{fig:humanevalserver}. The order of prompts and the order between models are fully randomized. Frequent attention checks are in place to ensure data quality.

\humanevalserver

\subsubsection{Elo Score Calculation}
\label{supsubsubsec:elo_score}
To calculate rankings when comparing more than two models based on 1v1 comparisons as outlined in \Cref{supsubsubsec:human_eval_setup}, we use Elo Scores (higher-is-better)~\citep{elo1978rating}, which were originally proposed as a scoring method for chess players but have more recently also been applied to compare instruction-tuned generative LLMs~\cite{bai2022training,askell2021general}. For a set of competing players with initial ratings $R_{\text{init}}$ participating in a series of zero-sum games, the Elo rating system updates the ratings of the two players involved in a particular game based on the expected and actual outcome of that game. Before the game with two players with ratings $R_1$ and $R_2$, the expected outcome for the two players is calculated as
\begin{align}
\label{eq:expected_elo}
E_1 =  \frac{1}{1 + 10^{\frac{R_2 - R_1}{400}}} \, , \\
E_2 =  \frac{1}{1 + 10^{\frac{R_1 - R_2}{400}}} \, . 
\end{align}
After observing the result of the game, the ratings $R_i$ are updated via the rule
\begin{align}
\label{eq:ranking_update}
R^{'}_{i} =  R_i + K \cdot \left(S_i - E_i \right), \quad i \in \{1,2\}
\end{align}
where $S_i$ indicates the outcome of the match for player $i$. In our case, we have $S_i=1$ if player $i$ wins and $S_i = 0$ if player $i$ loses. The constant $K$ can be seen as weight emphasizing more recent games. We choose $K=1$ and bootstrap the final Elo ranking for a given series of comparisons based on 1000 individual Elo ranking calculations in a randomly shuffled order. Before comparing the models, we choose the start rating for every model as $R_{\text{init}} = 1000$.

\subsection{Details on Experiments from \Cref{sec:approach}} 
 
\subsubsection{Architectural Details}
Architecturally, all models trained for the presented analysis in \Cref{sec:approach} are identical. To insert create a temporal UNet~\citep{ronneberger2015u} based on an existing spatial model, we follow \citet{blattmann2023align} and add temporal convolution and (cross-)attention layers after each corresponding spatial layer. As a base 2D-UNet, we use the architecture from \emph{Stable Diffusion 2.1}, whose weights we further use to initialize the spatial layers for all runs except the second one presented in \Cref{fig:imageonly_comp}, where we intentionally skip this initialization to create a baseline for demonstrating the effect of image-pretraining. 
Unlike \citet{blattmann2023align}, we train all layers, including the spatial ones, and do not freeze the spatial layers after initialization. All models are trained with the AdamW~\citep{loshchilov2017decoupled} optimizer with a learning rate of $1.e-4$ and a batch size of $256$. Moreover, in contrast to our models from \Cref{sec:sota}, we do not translate the noise process to continuous time but use the standard linear schedule used in \emph{Stable Diffusion 2.1}, including offset noise~\citep{guttenberg2023diffusion}, in combination with the v-parameterization~\citep{ho2022classifier}. We omit the text-conditioning in 10\% of the cases to enable classifier-free guidance~\citep{ho2022classifier} during inference. 
To generate samples for the evaluations, we use 50 steps of the deterministic DDIM sampler~\citep{song2020improved} with a classifier guidance scale of 12 for all models. 

\subsubsection{Calibrating Filtering Thresholds}
\label{supsubsec:filtering_ablations}
\smallscaleablations
Here, we present the outcomes of our study on filtering thresholds presented in \Cref{subsec:data_curation}. As stated there, we conduct experiments for the optimal filtering threshold for each type of annotation while not filtering for any other types. The only difference here is our assessment of the most suitable captioning method, where we simply compare all used captioning methods. We train each model on videos consisting of 8 frames at resolution $256 \times 256$ for exactly 40k steps with a batch size of 256, roughly corresponding to 10M training examples seen during training. For evaluation, we create samples based on 64 pre-selected prompts for each model and conduct a human preference study as detailed in \Cref{supsubsec:human_eval}. \Cref{fig:filtering_ablations} shows the ranking results of these human preference studies for each annotation axis for spatiotemporal sample quality and prompt following. Additionally, we show an averaged `aggregated' score. 

For \emph{captioning}, we see that - surprisingly - the captions generated by the simple clip-based image captioning method CoCa of \citet{yu2022coca} clearly have the most beneficial influence on the model. However, since recent research recommends using more than one caption per training example, we sample one of the three distinct captions during training. We nonetheless reflect the outcome of this experiment by shifting the captioning sampling distribution towards CoCa captions by using $p_{\text{CoCa}} = 0.5; \, p_{\text{V-BLIP}} = 0.25; \,p_{\text{LLM}} = 0.25; \,$.

For \emph{motion filtering}, we choose to filter out 25\% of the most static examples. However, the aggregated preference score of the model trained with this filtering method does not rank as high in human preference as the non-filtered score. The rationale behind this is that non-filtered ranks best primarily because it ranks best in the category `prompt following' which is less important than the `quality' category when assessing the effect of motion filtering. Thus, we choose the 25\% threshold, as mentioned above, since it achieves both competitive performances in `prompt following' and `quality'. 

For \emph{aesthetics filtering}, where, as for motion thresholding, the `quality' category is more important than the `prompt following'-category, we choose to filter out the 25 \% with the lowest aesthetics score, while for \emph{CLIP-score thresholding} we omit even 50\% since the model trained with the corresponding threshold is performing best. Finally, we filter out the 25\% of samples with the largest text area covering the videos since it ranks highest both in the `quality' category and on average.

Using these filtering methods, we reduce the size of \dataset by more than a factor of 3, \cf \Cref{tab:subset_stats}, but obtain a much cleaner dataset as shown in \Cref{sec:approach}. For the remaining experiments in \Cref{subsec:data_curation}, we use the identical architecture and hyperparameters as stated above. We only vary the dataset as detailed in \Cref{subsec:data_curation}.

\subsubsection{Finetuning Experiments}
\label{supsubsec:finetune_exps}
For the finetuning experiments shown in \Cref{subsec:stage3}, we again follow the architecture, training hyperparameters, and sampling procedure stated at the beginning of this section. The only notable differences are the exchange of the dataset and the increase in resolution from the pretraining resolution $256 \times 256$ to $512 \times 512$ while still generating videos consisting of 8 frames. We train all models presented in this section for 50k steps.

\subsection{Human Eval vs SOTA}

For comparison of our image-to-video model with state-of-the-art models like Gen-2~\cite{gen2} and Pika~\cite{pika}, we randomly choose 64 conditioning images generated from a $1024 \times 576$ finetune of SDXL~\cite{podell2023sdxl}. We employ the same framework as in~\Cref{supsubsubsec:human_eval_setup} to evaluate and compare the visual quality generated samples with other models. 

For Gen-2, we sample the image-to-video model from the web UI. We fixed the same seed of 23, used the default motion value of 5 (on a scale of 10), and turned on the ``Interpolate" and ``Remove watermark" features. This results in 4-second samples at $1408 \times 768$. We then resize the shorter side to yield $1056 \times 576$ and perform a center-crop to match our resolution of $1024 \times 576$. For our model, we sample our 25-frame image-to-video finetune to give 28 frames and also interpolate using our interpolation model to yield samples of 3.89 seconds at 28 FPS. We crop the Gen-2 samples to 3.89 seconds to avoid biasing the annotators.

For Pika, we sample the image-to-video model from the Discord bot. We fixed the same seed of 23, used the motion value of 2 (on a scale of 0-4), and specified a 16:9 aspect ratio. This results in 3-second samples at $1024 \times 576$, which matches our resolution. For our model, we sample our 25-frame image-to-video finetune to give 28 frames and also interpolate using our interpolation model to yield samples of 3.89 seconds at 28 FPS. We crop our samples to 3 seconds to match Pika and avoid biasing the annotators. Since Pika samples have a small ``Pika Labs" watermark in the bottom right, we pad that region with black pixels for both Pika and our samples to also avoid bias.

\subsection{UCF101 FVD} \label{supsec:ucf101_fvd}
This section describes the zero-shot UCF101 FVD computation of our base text-to-video model. The UCF101 dataset~\citep{soomro2012ucf101} consists of 13,320 video clips, which are classified into 101 action categories. All videos are of frame rate 25 FPS and resolution $240\times320$. To compute FVD, we generate 13,320 videos (16 frames at 25 FPS, classifier-free guidance with scale $w=7$) using the same distribution of action categories, that is, for example, 140 videos of ``TableTennisShot'', 105 videos of ``PlayingPiano'', etc. We condition the model directly on the action category (``TableTennisShot'', ``PlayingPiano'', etc.) and do not use any text modification. Our samples are generated at our model's native resolution $320 \times 576$ (16 frames), and we downsample to $240 \times 432$ using bilinear interpolation with antialiasing, followed by a center crop to $240 \times 320$. We extract features using a pretrained I3D action classification model~\citep{carreira2017quo}, in particular we are using a torchscript\footnote{\url{https://www.dropbox.com/s/ge9e5ujwgetktms/i3d_torchscript.pt} with keyword arguments \texttt{rescale=True, resize=True, return\_features=True}.} provided by~\citet{brooks2022generating}.

\subsection{Additional Samples}
Here, we show additional samples for the models introduced in \Cref{sec:base-model,sec:txt2vid,sec:img2vid,sec:multiview}.

\subsubsection{Additional Text-to-Video Samples}
\additionaltexttwovideo
In \Cref{fig:additional_txt2vid}, we show additional samples from our text-to-video model introduced in~\Cref{sec:txt2vid}.
\subsubsection{Additional Image-to-Video Samples}
\additionalimagetwovideo
In \Cref{fig:additional_img2vid}, we show additional samples from our image-to-video model introduced in \Cref{sec:img2vid}.
\subsubsection{Additional Camera Motion LoRA Samples}
\additionalmotionlora
In \Cref{fig:additional_motion_lora}, we show additional samples for our motion LoRA's tuned for camera control as presented in \Cref{subsec:motion_lora}.
\subsubsection{Temporal Prompting via Temporal Cross-Attention Layers}
\additionaltemporalattention
Our architecture follows \citet{blattmann2023align}, who introduced dedicated temporal cross-attention layers, which are used interleaved with the spatial cross-attention layers of the standard 2D-UNet~\citep{dhariwal2021diffusion,ho2020ddpm}. During probing our Text-to-Video model from \Cref{sec:txt2vid}, we noticed that it is possible to independently prompt the model spatially and temporally by using different text-prompts as inputs for the spatial and temporal cross-attention conditionings, see \Cref{fig:additional_temporal_attention}. To achieve this, we use a dedicated spatial prompt to describe the general content of the scene to be depicted while the motion of that scene is fed to the model via a separate temporal prompt, which is the input to the temporal cross-attention layers. We provide an example of these first experiments indicating this implicit disentanglement of motion and content in \Cref{fig:additional_temporal_attention}, where we show that varying the temporal prompt while fixing random seed and spatial prompt leads to spatially similar scenes that obtain global motion properties following the temporal prompt.       

\subsubsection{Additional Samples on Multi-View Synthesis}
In \Cref{fig:additional_mv1,fig:additional_mv2,fig:additional_mv3,fig:additional_mv4}, we show additional visual examples for SVD-MV, trained on our renderings of Objaverse and MVImageNet datasets as described in \Cref{sec:multiview}.
\MVIone
\MVItwo
\MVIthree
\MVIfour

\end{document}